\documentclass[sigconf]{acmart}

\renewcommand\footnotetextcopyrightpermission[1]{}

\AtBeginDocument{%
  }

\usepackage{array}           
\usepackage{booktabs}        
\usepackage{colortbl}        
\usepackage{multirow}        
\usepackage{hhline}          
\usepackage{caption}         
\usepackage{tabularx}        
\usepackage{booktabs}
\usepackage{makecell}
\usepackage{enumitem} 
\usepackage{tcolorbox}
\usepackage{booktabs}  
\usepackage{adjustbox} 
\usepackage{amsmath}
\usepackage{arydshln}   
\usepackage{pifont}
\usepackage{xcolor}

\begin{document}

\title{How Far Are Large Multimodal Models from Human-Level Spatial Action? A Benchmark for Goal-Oriented Embodied Navigation in Urban Airspace}

\author{Baining Zhao}
\authornote{All three authors contributed equally to this research. Ziyou Wang conducted this work during their internship at Tsinghua University.}
\affiliation{%
	\institution{Shenzhen International Graduate School, Tsinghua University}
	\city{Shenzhen}
	\country{China}}

\author{Ziyou Wang}
\authornotemark[1]
\affiliation{%
	\institution{Northeastern University}
	\city{Qinhuangdao}
	\country{China}}

\author{Jianjie Fang}
\authornotemark[1]
\affiliation{%
	\institution{Shenzhen International Graduate School, Tsinghua University}
	\city{Shenzhen}
	\country{China}}

\author{Zile Zhou}
\affiliation{%
	\institution{Shenzhen International Graduate School, Tsinghua University}
	\city{Shenzhen}
	\country{China}}
	
\author{Yanggang Xu}
\affiliation{%
	\institution{Shenzhen International Graduate School, Tsinghua University}
	\city{Shenzhen}
	\country{China}}

\author{Yatai Ji}
\affiliation{%
	\institution{National University of Defense Technology}
	\city{Changsha}
	\country{China}}
	
\author{Jiacheng Xu}
\affiliation{%
	\institution{Shandong University}
	\city{Jinan}
	\country{China}}
	
\author{Qian Zhang}
\affiliation{%
	\institution{Shenzhen International Graduate School, Tsinghua University}
	\city{Shenzhen}
	\country{China}}
	
\author{Weichen Zhang}
\affiliation{%
	\institution{Shenzhen International Graduate School, Tsinghua University}
	\city{Shenzhen}
	\country{China}}

\author{Chen Gao}
\authornote{Corresponding authors: Chen Gao (chgao96@gmail.com), Xinlei Chen (chen.xinlei@sz.tsinghua.edu.cn)}
\affiliation{%
	\institution{BNRist, Tsinghua University}
	\city{Beijing}
	\country{China}}

\author{Xinlei Chen}
\authornotemark[2]
\affiliation{%
	\institution{Shenzhen International Graduate School, Tsinghua University}
	\city{Shenzhen}
	\country{China}}



\settopmatter{authorsperrow=4}

\pagestyle{fancy}
\fancyhf{}
\fancyhead[L]{} 
\fancyhead[R]{}
\fancyfoot[C]{} 

\begin{abstract}
  Large multimodal models (LMMs) show strong visual-linguistic reasoning but their capacity for spatial decision-making and action remains unclear. In this work, we investigate whether LMMs can achieve embodied spatial action like human through a challenging scenario: goal-oriented navigation in urban 3D spaces. We first spend over 500 hours constructing a dataset comprising 5,037 high-quality goal-oriented navigation samples, with an emphasis on 3D vertical actions and rich urban semantic information. Then, we comprehensively assess 17 representative models, including non-reasoning LMMs, reasoning LMMs, agent-based methods, and vision-language-action models. Experiments show that current LMMs exhibit emerging action capabilities, yet remain far from human-level performance. Furthermore, we reveal an intriguing phenomenon: navigation errors do not accumulate linearly but instead diverge rapidly from the destination after a critical decision bifurcation. The limitations of LMMs are investigated by analyzing their behavior at these critical decision bifurcations. Finally, we experimentally explore four promising directions for improvement: geometric perception, cross-view understanding, spatial imagination, and long-term memory. The project is available at: \textcolor{blue}{\url{https://github.com/serenditipy-AC/Embodied-Navigation-Bench}}.
\end{abstract}




\maketitle

\section{Introduction}

Large multimodal models (LMMs) have demonstrated remarkable performance in tasks such as image analysis and video understanding~\cite{OpenAI_API,Gemini_API,Qwen_Website}. Recently, LMMs have been further explored as embodied brains, with the potential to control robots for movement and manipulation in 3D environments~\cite{zhang-kordjamshidi-2023-vln,fan-etal-2023-aerial}. This trend raises a fundamental question: Do LMMs possess human-level spatial intelligence? Specifically, can they develop an understanding of 3D environments and act accordingly, based on a continuous stream of visual observations~\cite{driess2023palm, gao2025openfly}? It is crucial for understanding the current boundaries of physical AI.


\begin{figure*}[!t]
	\centering
	\includegraphics[width=0.9\linewidth]{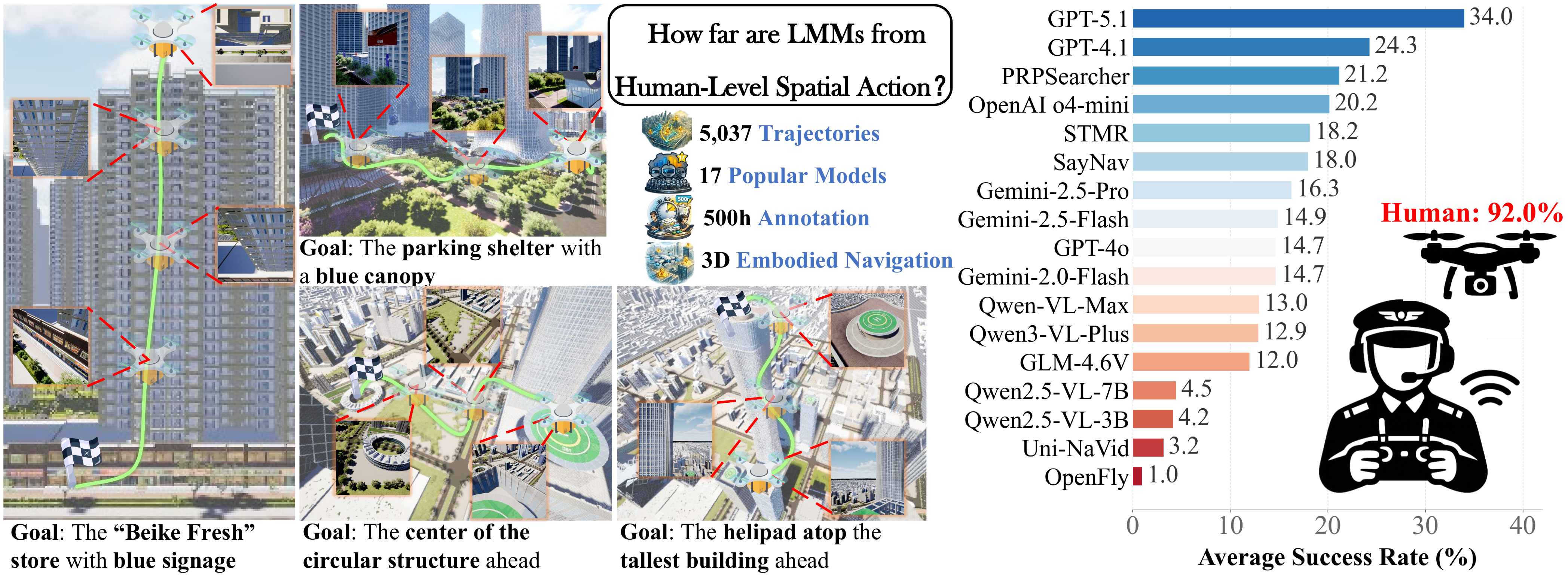}
	\caption{\textbf{Overview of the proposed benchmark.} Goal-oriented embodied navigation in urban airspace is defined as: given linguistic instructions, the agent takes progressive actions based on continuous embodied observations to approach the target location. Current LMMs still exhibit a substantial gap compared to humans in spatial action.}
	\label{data_example}
\end{figure*}

Goal-oriented embodied navigation is one of the core tasks for embodied spatial intelligence~\cite{zhu2021soon,wu2024vision,yin2025unigoal}, providing an effective means to evaluate the spatial action capabilities of LMMs. This task requires LMMs to control robots in a goal-driven manner, advancing beyond traditional manual programming or low-level instruction following. The target location is provided solely through natural language instructions, without the need for manually written low-level control codes or predefined routes~\cite{o2024open,kim2024openvla}. This represents a higher level of human-robot interaction and robotic intelligence~\cite{driess2023palm,ren2024embodied}. Specifically, LMMs must comprehend natural language instructions, dynamically interpret visual observations of the surrounding environment, continuously plan and execute actions to reach a specified position. Unlike offline, single-step question-answering tasks~\cite{yang2025thinking,zhao2025urbanvideo}, goal-oriented embodied navigation enables an online, continuous, and comprehensive evaluation of LMMs' spatial understanding, decision-making, and action capabilities.

Urban aerial environments present a challenging and suitable domain for evaluating the spatial intelligence of LMMs~\cite{xu2023urban,li2024vln}. Compared to indoor settings, urban spaces feature multi-dimensional semantics that are richer, more ambiguous, and open-ended, demanding advanced visual-semantic perception from LMMs. In contrast to ground-level scenarios, the aerial perspective encompasses a larger scale and a higher density of visual elements, necessitating scene understanding through the integration of continuous visual observations~\cite{yao2024aeroverse,lee2024citynav, gao2024embodied}. Furthermore, the expansive aerial action space requires superior planning and control capabilities built upon this robust perception and understanding~\cite{doole2020estimation, salama2023ai}. Therefore, this more complex scenario serves as a more effective benchmark for assessing the spatial abilities of LMMs. Additionally, the goal-oriented formulation is crucial for advancing real-world drone applications such as emergency response and logistics for future cities~\cite{mangiaracina2019innovative,ma2021hierarchical,roberts2023current,liu2023drones}.

Existing research in embodied navigation has predominantly focused on indoor or ground-based agents~\cite{liang-etal-2024-cornav,majumdar2022zson}. The few works~\cite{zhang-kordjamshidi-2023-vln,fan-etal-2023-aerial} that address vision-language navigation (VLN) in urban aerial settings are largely route-oriented, relying on highly detailed, step-by-step instructions (e.g., ``Fly... then descend...'')~\cite{shah2023lm, schumann2024velma, zhou2023navgpt}. 
Comparatively, we focus on the less-explored and more challenging goal-oriented embodied navigation (e.g., ``Approach a red tree near the building''). 

\begin{table*}[!t]
    \centering
    \caption{Benchmarks (simulators and datasets) for aerial embodied navigation in cities.}
    \label{tab:benchmark}
    \setlength{\tabcolsep}{12pt} 
    \resizebox{0.8\linewidth}{!}
{
    \begin{tabular}{cccccc}
        \hline
        Benchmark & Simulator &  Dataset Task Type  & \textbf{Action Distribution} & \textbf{LMMs} \\ \hline
        \makecell{AVDN \cite{fan2022aerial}} & Satellite Images  & Dialog Navigation & Pure Horizontal  & \ding{55} \\ 
        \makecell{AerialVLN \cite{liu2023aerialvln}} & UE + Airsim  & Route-Oriented Navigation & Horizontal-Dominant  & \ding{55} \\ 
        \makecell{AeroVerse  \cite{yao2024aeroverse}} & UE + Airsim  & Route-Oriented Navigation & Horizontal-Dominant  & \ding{51}  \\ 
        \makecell{EmbodiedCity \cite{gao2024embodied}} & UE + Airsim & Route-Oriented Navigation & Horizontal-Dominant & \ding{51} \\ 
        \makecell{CityNav \cite{lee2024citynav}} & WebGL-Based  & Goal-Oriented Navigation & Horizontal-Dominant  & \ding{55} \\ 
        \makecell{TRAVEL \cite{wang2024towards}} & UE + Airsim  & Goal-Oriented Navigation & Horizontal-Dominant  & \ding{55} \\ 
        \makecell{UAV-On \cite{xiao2025uav}} & UE + Airsim & Goal-Oriented navigation & Horizontal-Dominant  & \ding{55} \\ \hline
        \textbf{Ours} & UE + Airsim & Goal-Oriented navigation & Balanced Horizontal–Vertical  & \ding{51} \\ \hline
    \end{tabular}
    }
\end{table*}

This benchmark distinguishes itself from existing work in three key areas: scenario, task setup, action distribution, and model analysis.
\textbf{Scenario:} Existing research in embodied navigation predominantly focuses on indoor or ground-based agents~\cite{liang-etal-2024-cornav,majumdar2022zson}, with limited attention to urban aerial scenarios.
\textbf{Task Setup:} Current benchmarks~\cite{zhang-kordjamshidi-2023-vln,fan-etal-2023-aerial} addressing vision-language navigation (VLN) in urban aerial settings are largely route-oriented, relying on highly detailed, step-by-step instructions (e.g., ``Fly... then descend...'')~\cite{shah2023lm, schumann2024velma, zhou2023navgpt}. In contrast, we focus on the less-explored and more challenging goal-oriented embodied navigation (e.g., ``Approach a red tree near the building'').
\textbf{Action Distribution:} Existing drone navigation benchmarks predominantly focus on horizontal movements in their task design~\cite{liu2023aerialvln,xiao2025uav}. In contrast, we design navigation tasks with richer vertical attributes, making them more representative of real-world drone applications.
\textbf{Model Analysis:} Both the aforementioned route-oriented navigation benchmarks and the few goal-oriented navigation benchmarks~\cite{wang2024towards,xiao2025uav} have only provided performance evaluations of a limited number of small-scale neural networks or agent-based methods. This paper aims to provide comprehensive insights into the ability of large multimodal models (LMMs) to directly perform human-like spatial actions.

Specifically, we propose a benchmark to assess LMMs' ability to perform human-level actions in 3D spaces through goal-oriented embodied navigation in urban airspace, as presented in Figure~\ref{data_example}. To ensure reproducible experiments, we collected 5,037 high-quality navigation trajectories with goal-oriented instructions through over 500 hours of human-controlled data collection in a realistic simulator. We then evaluate 17 models, including non-reasoning LMMs, reasoning LMMs, existing agent-based approaches, and vision-language-action (VLA) models. By analyzing navigation completion progress curves, we observe a critical decision bifurcation (CDB) phenomenon, where LMM navigation errors do not accumulate linearly but instead diverge from the destination after a wrong decision. From the CDB perspective, we further dissect the reasoning processes of LMMs and identify four main shortcomings. Through further experiments, potential directions for improvement are explored. The contributions of this work are as follows:

\begin{itemize}[leftmargin=*]
	\item We build a high-quality dataset of 5,037 goal-oriented embodied navigation trajectories in urban airspace, with balanced horizontal-vertical action distribution. 
	\item We perform a comprehensive evaluation of spatial action capabilities on 17 popular models, including non-reasoning LMMs, reasoning LMMs, existing agent-based approaches, and VLA models.
	\item We introduce the critical decision bifurcation (CDB) phenomenon to characterize failures in LMM spatial action. Based on this perspective, we provide insights into current limitations and outline future optimization directions, including geometric perception, cross-view understanding, spatial imagination, and sparse memory.
	
\end{itemize}

\section{Related Work}

We review related research from three perspectives, with a progressively narrowing scope: embodied spatial intelligence, vision-language navigation, and benchmarks for urban aerial navigation.

\textbf{Embodied Spatial Intelligence.} 
Current research focuses on leveraging LMMs to handle tasks like navigation \cite{lin2023development,song2025towards}, manipulation \cite{huang2023voxposer}, and task planning \cite{wang2023describe,zhao2024large}. 
Some foundational models are designed specifically for embodied AI, such as Visual Cortex VC-1~\cite{majumdar2023we} and ViNT~\cite{shah2023vint}. These works motivate us to further investigate the capability boundaries of LMMs.
While spatial mobility is a fundamental capability for any robot, existing benchmarks have primarily evaluated the spatial reasoning of LMMs through offline question-answering~\cite{feng2025survey}. For instance, VSI-Bench~\cite{yang2025thinking} and UrbanVideo-Bench~\cite{zhao-etal-2025-urbanvideo} assess spatial cognition in indoor and outdoor environments, respectively. However, for a system like a robot that interacts with its environment in real-time, an ``embodied brain'' must demonstrate continuous, multi-step capabilities for spatial understanding and decision-making. It remains underexplored whether LMMs, when operating under these dynamic conditions, can act like humans to achieve long-horizon mobility in 3D space.

\textbf{Vision-Language Navigation.} 
There are various types of tasks in VLN, each with distinct challenges and solutions~\cite{wu2024vision,dorbala2022clip,liang2024cornav}.
Unlike route-oriented tasks~\cite{shah2023lm, schumann2024velma, zhang-etal-2025-citynavagent} that require following specific paths or vision-and-dialog tasks~\cite{fan2022aerial} that involve human interaction, goal-oriented VLN~\cite{gadre2023cows,majumdar2022zson} demands agents to independently plan their routes. For example, NoMaD~\cite{sridhar2024nomad} introduce a unified diffusion policy that handles both goal-directed navigation and goal-agnostic exploration. Additionally, it contrasts with tasks like geographic navigation~\cite{lee2024citynav} that rely heavily on geographic cues and embodied planning that focuses on obstacle avoidance~\cite{yao2024aeroverse}. These differences in objectives and methodologies make direct transfer of models across different VLN tasks challenging.
Besides, existing goal-oriented VLN studies primarily focus on indoor environments for ground robots~\cite{gadre2023cows,majumdar2022zson,zhou2023esc,rajvanshi2023saynav}.

\textbf{Benchmarks on Urban Aerial Navigation.} 
Research on aerial embodied navigation in urban spaces is gradually gaining traction~\cite{liu2024navagent,gao2024aerial,zhang-etal-2025-citynavagent}. Due to stringent regulations on real drone flights in cities, the validation of these methods heavily relies on numerous benchmarks~\cite{liu2023aerialvln,fan2022aerial,lee2024citynav,yao2024aeroverse,wang2024towards,gao2024embodied,xiao2025uav}. As shown in Table \ref{tab:benchmark}, most benchmarks do not directly support goal-oriented navigation in the urban 3D spaces proposed in this paper. Among the remaining ones, TRAVEL~\cite{wang2024towards} and UAV-On~\cite{xiao2025uav} place greater emphasis on searching along a roughly specified horizontal direction based on scene descriptions. In contrast, the scenario in this work is closer to real-world applications of aerial agents, where the agent is already near the target location and navigates within the surrounding 3D space with a stronger focus on vertical actions. Moreover, these benchmarks primarily focus on task definition, lacking insights from the perspective of LMM analysis.

\section{Problem Formulation}

The objective of the goal-oriented embodied navigation problem is to reach the specified goal-location $p_L$, given through a natural-language goal instruction $I$. 
The process involves following an algorithm ${\rm{\pi }}$ to guide the agent through a sequence of observations and actions to reach the target location $p_L$.
At each time step $t$, the agent obtains an RGB observation $o_t$ with camera gimbal angle $\alpha $. 
The agent takes action ${a_t}$ based on ${\rm{\pi }}$: 
\begin{equation}
    {a_t} = {\rm{\pi }}\left( {{o_t},{\alpha}_t,I} \right).
\end{equation}
The action $a_t$ can be formed by the arbitrary combination of discrete control commands: turn-left, turn-right, move-forth, move-left, move-right, move-back, move-up, move-down, adjust-camera-gimbal-upwards, and adjust-camera-gimbal-downwards. 

In the urban space, the real position $p_t$ of the agent, which can not be observed by itself, is changed based on physical dynamics rules:
\begin{equation}
    {p_t} = f\left( {{a_t},{p_{t - 1}}} \right).
\end{equation}
After $T$ steps, the navigation is successful if the agent stops within a Euclidean distance of $\varepsilon $ meters from the target location $p_L$:
\begin{equation}
    \left\| {{p_T} - {p_L}} \right\| \le \varepsilon.
\end{equation}

We aim for the agent to reach the target location in diverse scenarios. Assuming there are scenarios $i=1,2,...,N$, the objective can be formally stated as follows:
\begin{equation}
    \mathop {\max }\limits_{\rm{\pi }} \frac{1}{N}\sum\limits_{i = 1}^N {1\left( {\left\| {p_T^{(i)} - p_L^{(i)}} \right\| \le \varepsilon } \right)},
\end{equation}
where $1\left(  \cdot  \right)$ is the indicator function, which is 1 if the condition inside is true and 0 otherwise.

\section{Dataset Construction}

\begin{figure*}[!t]
	\centering
	\includegraphics[width=0.85\linewidth]{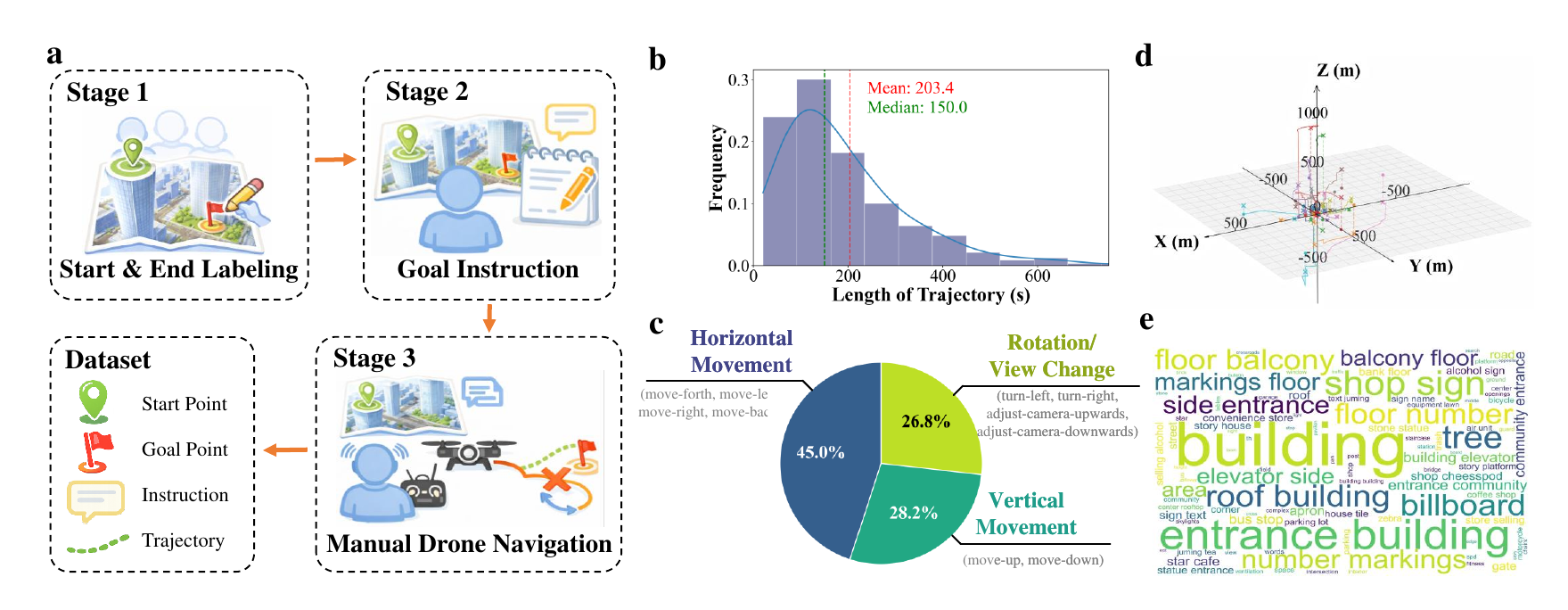}
    \vspace{-5pt}
	\caption{\textbf{a.} Dataset Construction Pipeline. \textbf{b.} The length distribution of navigation trajectories. \textbf{c.} Proportion of various types of actions. \textbf{d.} The relative position of trajectories to the origin.  \textbf{e.} Word cloud of goal instructions.}
	\label{fig:statistics}
\end{figure*}

We introduce the urban aerial goal-oriented navigation dataset from the following aspects: first, the choice of simulator; second, the construction and annotation of navigation cases; third, the statistical characteristics of the dataset to highlight its 3D properties; and finally, the three main metrics used for navigation evaluation.

\subsection{Simulator Selection}
We first select a simulator to establish the dataset. 
After evaluating simulators in Table \ref{tab:benchmark}, we excluded simulators that are 2D (AVDN), low-quality (CityNav), and those with restricted editing permissions (AerialVLN, AeroVerse, TRAVEL).
EmbodiedCity~\cite{gao2024embodied} consists of an open-source simulator and a route-oriented VLN dataset. The route-oriented VLN dataset differs from the focus of this work, but the simulator is highly adaptable for further development. It is built on a real city and integrates various urban elements.

\subsection{Data Collection and Annotation}
For each goal-oriented navigation trajectory, we collect the start coordinates, the ground-truth end coordinates, a textual instruction specifying the goal location, and the corresponding ground-truth trajectory. The trajectories are required to (i) exhibit full 3D aerial characteristics consistent with real-world UAV scenarios, (ii) use semantically clear instructions that uniquely identify the target location, and (iii) closely reflect human goal-directed navigation behavior. These requirements make existing automated waypoint selection and annotation methods inadequate for producing high-quality goal-oriented navigation data. As shown in Figure~\ref{fig:statistics}a, we therefore adopt a three-stage pipeline: First, start and end locations are selected in urban environments to form semantically meaningful navigation tasks guided by city-level cues. Second, a goal-oriented textual instruction is generated that specifies only the target location, avoiding explicit action commands typical of route-oriented instructions while remaining sufficient for commonsense, step-by-step planning. Third, professional UAV pilots with over 100 hours of real-world flight experience execute the task by navigating from the start point to the destination according to the instruction. To approximate a zero-shot setting, task design and ground-truth trajectory execution are conducted by separate groups. Overall, the process involved 10 volunteers and required more than 500 person-hours of manual effort.

\begin{table*}[t]
	\centering
	\caption{Goal-oriented embodied navigation results of the popular LMMs and agent-based approaches. The short, middle, and long groups correspond to ground truth trajectories of less than 118.2 meters, between 118.2 meters and 223.6 meters, and greater than 223.6 meters, respectively.}
	\label{comparison_results}

		\begin{minipage}[t]{0.75\textwidth} 
        \raisebox{1.25\height}{
\centering
\setlength{\tabcolsep}{1.3pt} 
\resizebox{\linewidth}{!}{
		\begin{tabular}{ccccccccccccccc}
			\toprule
			\multirow{2}*{Method}  & \multirow{2}*{RGB} & \multirow{2}*{Depth} &  \multicolumn{3}{c}{Short} & \multicolumn{3}{c}{Middle} & \multicolumn{3}{c}{Long}  & \multicolumn{3}{c}{Average}             \\
			\cmidrule(r){4-6}
			\cmidrule(r){7-9}
			\cmidrule(r){10-12}
			\cmidrule(r){13-15}
			~     & ~       & ~   & SR $\uparrow$      & SPL $\uparrow$  & DTG $\downarrow$ & SR $\uparrow$      & SPL $\uparrow$  & DTG $\downarrow$ & SR $\uparrow$      & SPL $\uparrow$  & DTG $\downarrow$ & SR $\uparrow$      & SPL $\uparrow$  & DTG $\downarrow$  \\ 
			\midrule
			\multicolumn{15}{c}{\cellcolor[HTML]{ECF4FF}\textit{\textbf{Basic Baselines}}}\\
			Human&  $\checkmark$ & & 95.2 & 93.8 & 15.1 & 94.6 & 92.9 & 31.0 & 86.3 & 84.4 & 57.1 & 92.0 & 90.4 & 34.4 \\ 
			Random&  & & 7.3 & 6.0 & 63.9 & 0 & 0 & 125.3 & 0 & 0 & 257.7 & 2.4 & 2.0 & 149.0\\ 
			Action Sampling&   & & 7.4 & 5.3 & 84.1 & 0 & 0 & 160.8 & 0 & 0 & 271.2 & 2.5 & 1.8 & 172.0 \\ 
			\multicolumn{15}{c}{\cellcolor[HTML]{ECF4FF}\textit{\textbf{Action-as-Language: Non-Reasoning LMMs}}}\\
			Qwen2.5-VL-3B & $\checkmark$ & & 9.7& 9.2& 148.6 & 3.0 & 2.8 & 215.0 & 0 & 0 & 467.5 & 4.2 & 4.0 & 277.0 \\
			Qwen2.5-VL-7B & $\checkmark$ & & 9.9& 9.5& 86.2 & 3.5 & 3.2 & 109.9 & 0 & 0 & 300.7 & 4.5 & 4.2 & 165.6 \\
			Qwen-VL-Max & $\checkmark$ & & 27.3 & 27.0 & 78.9 & 6.4 & 6.2 & 148.2 & 5.3 & 4.7 & 344.7 & 13.0 & 12.6 & 190.6 \\
            GLM-4.6V  & $\checkmark$ & & 29.3 & 29.1 & 62.3 & 6.2 & 6.1 & 115.5 & 0 & 0 & 278.2 & 12.0 & 11.9 & 151.1  \\
			Gemini-2.0-Flash & $\checkmark$ & & 31.8& 29.3 & 64.4 & 11.3 & 10.6 & 130.8 & 0 & 0 & 335.7 & 14.7 & 13.3 & 177.0 \\
			GPT-4o & $\checkmark$ & & 34.2 & 30.5 & 70.7 & 9.8 & 7.9 & 90.9 & 0 & 0 & 200.2 & 14.7 & 12.8 & 120.6 \\
			GPT-4.1 & $\checkmark$ & & 51.5 & 45.7 & 56.8 & 15.2 & 15.0 & 93.0 & 6.3 & 5.9 & 189.3 & 24.3 & 22.2 & 113.0 \\ \multicolumn{15}{c}{\cellcolor[HTML]{ECF4FF}\textit{\textbf{Action-as-Language: Reasoning LMMs}}}\\
			Qwen3-VL-Plus & $\checkmark$ & & 30.1 & 29.7 & 83.7 & 8.7 & 8.5 & 122.9 & 0 & 0 & 301.2 & 12.9 & 12.7 & 169.3\\
            Gemini-2.5-Flash & $\checkmark$ & & 35.5& 33.6 & 71.4 & 9.1 & 9.1 & 95.2 & 0 & 0 & 223.3 & 14.9 & 14.2 & 129.4\\
			Gemini-2.5-Pro & $\checkmark$ & & 27.7& 24.8& 64.0 & 11.5 & 10.1 & 105.5 & 9.8 & 8.3 & 388.2 & 16.3  & 14.4 & 185.9 \\
			OpenAI o4-mini & $\checkmark$ & & 35.6 & 33.1 & 51.7 & 22.4 & 19.8 & 91.4 & 2.7 & 2.6 & 224.3 & 20.2 & 18.5 & 122.5 \\
            GPT-5.1 & $\checkmark$ & & \textbf{52.9} & \textbf{48.6} & 60.9 & \textbf{30.4} & \textbf{29.0} & 75.3 & \textbf{18.2} & \textbf{18.0} & 186.1 & \textbf{34.0} & \textbf{32.0} & 107.0 \\
			\multicolumn{15}{c}{\cellcolor[HTML]{ECF4FF}\textit{\textbf{Action-as-Planning: Agent-Based Models}}}\\
			SayNav & $\checkmark$ & $\checkmark$ & 25.6 & 24.3 & 124.0 & 21.7 & 19.0 & 132.6 & 6.8 & 5.5 & 225.0 & 18.0 & 16.3 & 160.5 \\
			STMR & $\checkmark$ & $\checkmark$ & 35.9 & 23.8 & 71.5 & 15.0 & 10.6 & 87.7 & 3.6 & 3.1 & 183.8 & 18.2 & 12.5  & 114.3 \\
			PRPSearcher   & $\checkmark$ & $\checkmark$ & 42.4 & 24.4 & \textbf{27.0} & 18.2 & 12.6 & \textbf{52.0} & 2.9 & 1.9 & \textbf{170.6} &21.2 & 13.0 &\textbf{83.2}\\
            \multicolumn{15}{c}{\cellcolor[HTML]{ECF4FF}\textit{\textbf{Action-as-Token: VLA Models}}}\\
			OpenFly & $\checkmark$ & & 3.0 & 3.0 & 63.3 & 0 & 0 & 114.1 & 0 & 0 & 258.7 & 1.0 & 1.0 & 146.5  \\
			Uni-NaVid & $\checkmark$ & & 6.1 & 5.2 & 80.9 & 3.4 & 3.0 & 118.5 & 0 & 0 & 262.0 & 3.2 & 2.7 & 154.9 \\
			\bottomrule
		\end{tabular}
        }}
\end{minipage}
\hfill
		\begin{minipage}[t]{0.215\textwidth}
        \raisebox{0.16\height}{
 \centering \includegraphics[width=1\linewidth]{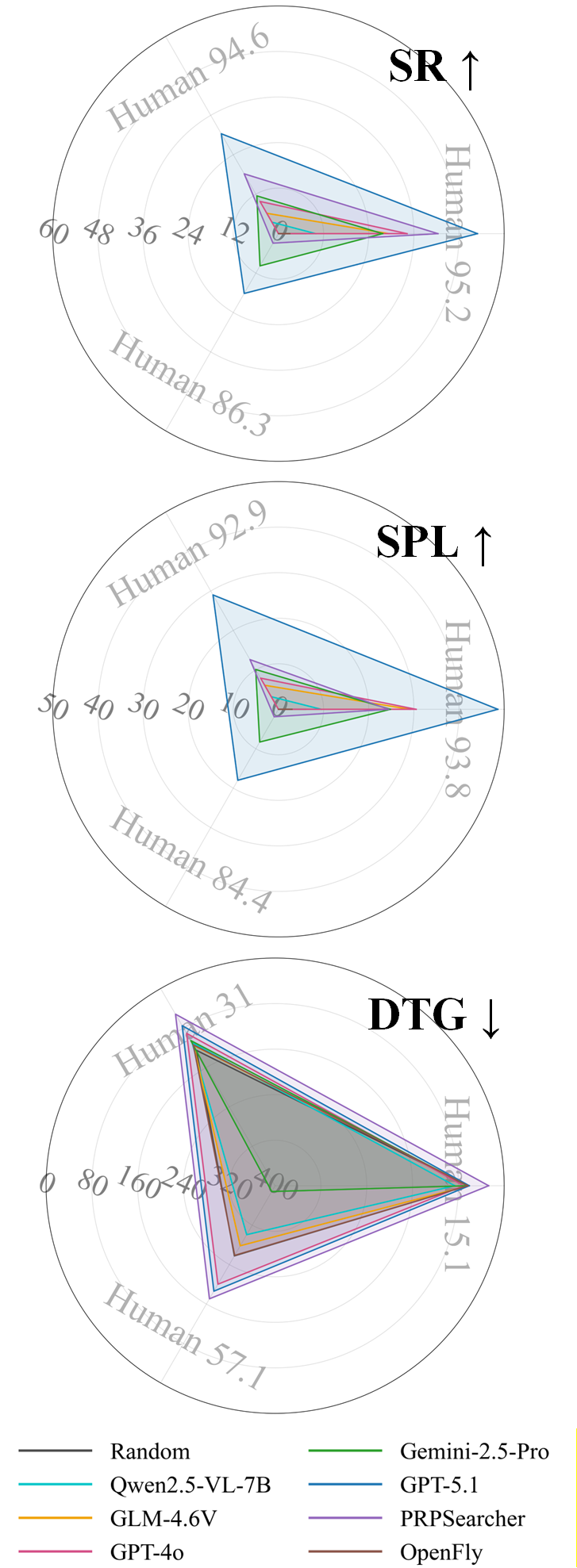}} 
\end{minipage}
\vspace{-40pt}
\end{table*}

\subsection{Dataset Statistics}
We have totally collected and annotated $N=5,037$ flight trajectories. The statistical characteristics of the dataset are presented in the Figure \ref{fig:statistics}b-e. Figure \ref{fig:statistics}b shows the distribution of ground-truth trajectory lengths, with an average of approximately 203.4 meters. Figure \ref{fig:statistics}c shows the average proportions of horizontal movement, vertical movement, and rotation actions in a complete navigation task, accounting for 45.0\%, 28.2\%, and 26.8\%, respectively. The proportion of vertical actions highlights the dataset's emphasis on low-altitude urban 3D navigation, distinguishing it from existing datasets. Although rotation and gimbal adjustment actions constitute the smallest share, they are critical for maintaining proper orientation in perception and action. Figure \ref{fig:statistics}d randomly samples several trajectories and aligns their starting points at the origin within the same coordinate system. It reveals that trajectories move uniformly in all horizontal directions, while downward movements occur more frequently than upward ones in the vertical direction. This aligns with practical applications, as drones typically navigate downward toward their destinations. Figure \ref{fig:statistics}e presents a word cloud of goal instructions, highlighting the urban elements included.

\subsection{Evaluation Metrics}
We utilize three standard metrics to evaluate the navigation performance: Success Rate (SR), Success Weighted by Path Length (SPL), and Distance to Goal (DTG) \cite{anderson2018evaluation, yu2023l3mvn, zhou2023navgpt}. SR indicates the proportion of delivery episodes where the agent successfully reaches the target location within a specific margin of error. It is calculated using ${\rm{SR}} = \frac{1}{N}\sum\nolimits_{i = 1}^N {{s_i}} $, where $N$ is the number of delivery episodes and $s_i$ represents the success of the $i$-th delivery, where it takes a value of 1 for success and 0 for failure. As a metric that considers both navigation precision and efficiency, SPL comprehensively takes into account the SR and the corresponding ratio of the optimal path length ${{l_i}}$ to the actual delivery path length ${{g_i}}$. The calculation formula is represented as ${\rm{SPL}} = \frac{1}{N}\sum\nolimits_{i = 1}^N {{s_i}\frac{{{l_i}}}{{\max ({l_i},{g_i})}}} $. DTG is computed by ${\rm{DTL}} = \frac{1}{N}\sum\nolimits_{i = 1}^N {{d_i}} $, where ${{d_i}}$ denotes average distance from the agent's final location to the destination.

\section{Experiment}

In experiments, we first quantitatively evaluate the performance of LMMs and related models on the navigation task. Subsequently, we conduct a qualitative analysis of the LMMs' action logic through case studies, focusing on their decision-making processes during navigation. Based on this analysis, we identify and summarize four primary error categories that impede the successful application of LMMs for end-to-end action in urban airspace. Correspondingly, we conducted four experiments to validate the aforementioned drawbacks and furnish insights that can guide future research on improving LMMs.

 \begin{figure*}[!t]
     \centering
     \includegraphics[width=0.9\linewidth]{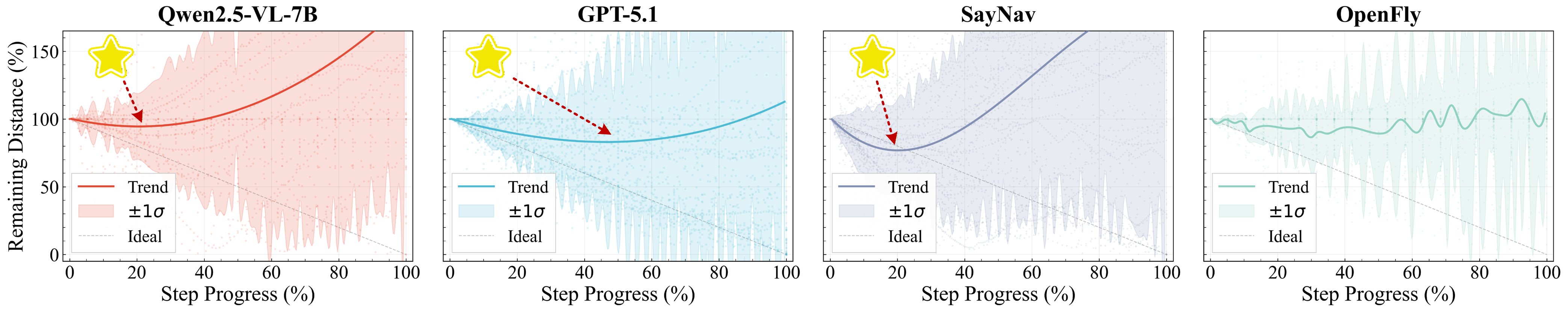}
     \caption{The change in navigation completion progress (\%) as a function of navigation steps. Navigation completion progress is defined as the ratio between the distance from the current step to the goal and the distance from the initial position to the goal. The star-marked point is defined as the Critical Decision Bifurcation (CDB), which characterizes the spiral of divergence in LMM's embodied navigation.}
     \label{fig:cdb}
 \end{figure*}

\subsection{Experimental Setup}

\subsubsection{Implementation: }
To evaluate the spatial action capabilities of LMMs, we adopt three paradigms.
First, \textbf{action-as-language}, where the model directly outputs language-based action commands; discrete spatial actions are expressed in natural language. At each timestep, the input consists of the current RGB observation and a memory of past moments, including previous observations, actions, and rationales, and the output comprises an action command with its rationale.
Second, \textbf{action-as-reasoning}, an agent-based paradigm in which the LMM primarily performs high-level reasoning and planning.
Third, \textbf{action-as-token}, following the vision-language-action (VLA) paradigm, where actions are represented as tokens and predicted end-to-end for direct decoding. 
See Appendix~\ref{Implementation_Details} for details.

\subsubsection{Baselines: }
Accordingly, the representative baselines can be categorized into the following types: 

\begin{itemize}[leftmargin=*,partopsep=0pt,topsep=0pt]
	\setlength{\itemsep}{0pt}
	\setlength{\parsep}{0pt}
	\setlength{\parskip}{0pt}
	\item \textbf{Basic Navigation Baselines:} This includes methods such as \textbf{Human Operation}, \textbf{Random}, and \textbf{Action Sampling}~\cite{vemprala2023chatgpt}, which serve as fundamental approaches for navigation tasks.
	\item \textbf{Non-Reasoning LMMs (Action-as-Language):} This category encompasses both open-source and commercial models, including \textbf{Qwen2.5-VL series}~\cite{bai2025qwen2}, \textbf{Qwen-VL-Max}~\cite{Qwen_Website}, \textbf{GLM-4.6V}~\cite{glm2024chatglm}, \textbf{Gemini-2.0-Flash}~\cite{Gemini_API}, and \textbf{GPT multimodal series}~\cite{OpenAI_API}.
	\item \textbf{Reasoning LMMs (Action-as-Language):} Inspired by the success of slow thinking in solving complex logical problems such as mathematics, reasoning models may exhibit transferable capabilities for embodied tasks. Here, we select four popular multimodal reasoning models: \textbf{Qwen3-VL-Plus}~\cite{Qwen_Website}, \textbf{Gemini-2.5 series}~\cite{Gemini_API}, and \textbf{OpenAI o4-mini, GPT-5.1}~\cite{OpenAI_API}.
	\item \textbf{Agent-Based Approaches (Action-as-Reasoning):} Direct solutions for the goal-oriented aerial embodied navigation task in urban spaces are currently unavailable. Therefore, we adapt an indoor object navigation agent and two existing aerial VLN methods to suit our scenario: \textbf{SayNav} \cite{rajvanshi2023saynav}, \textbf{STMR}~\cite{gao2024aerial}, and \textbf{PRPSearcher}~\cite{ji2025towards}. For the agent-based approaches, GPT-4o was selected as the backbone model owing to considerations of computational cost and inference time.
    \item \textbf{Vision-Language-Action Model (Action-as-Token):} Recently, VLA models specifically designed for action have gained significant attention~\cite{kim2024openvla}. These models align physical action tokens with language tokens through training. We select the publicly available mobility-focused VLA models, \textbf{OpenFly}~\cite{gao2025openfly} and \textbf{Uni-NaVid}~\cite{zhang2024uni}, for evaluation. We match the inputs and outputs of the VLA models with our goal-oriented navigation task while striving to preserve model performance.
\end{itemize}

\subsection{Overall Performance}

All navigation cases are divided into three groups based on the trisection points of trajectory length: short, middle, and long. Generally, the longer the navigation distance of a case, the more complex and challenging the decision-making process becomes. The results, presented in Table \ref{comparison_results}, lead us to the following observations.

\begin{itemize}[leftmargin=*,partopsep=0pt,topsep=0pt]
\setlength{\itemsep}{0pt}
\setlength{\parsep}{0pt}
\setlength{\parskip}{0pt}
    \item \textbf{The proposed benchmark is suited for evaluating the spatial action capabilities of LMMs.} Both the random and action sample methods exhibit SR and SPL scores close to 0 in middle-distance and long-distance groups. This indicates that the task encompasses a vast action space. The agent fails to reach or even approach the destination without understanding the instructions, visual perceptions, and their alignment.
    \item \textbf{LMMs still exhibit a considerable gap compared to human spatial action capabilities, but they show a trend toward generalizing this ability.} The accuracy of LMMs ranges from 4.2\% to 34.0\%, significantly lagging behind human-level performance.
    Despite this overall deficit, the results demonstrate a clear trajectory of improvement in LMMs' action capabilities. A cross-model comparison reveals that close-source models possess significantly better generalization capabilities than their open-source counterparts. Furthermore, newer models consistently outperform their predecessors within the same model family. As one of the SOTA LMMs, GPT-5 reaches an accuracy of 52.9\% in the short group. This suggests that the spatial intelligence of LMMs benefits from increased model scale and advanced training methodologies, indicating potential for stronger performance in the future. In previous navigation research, LMMs were typically not used as a baseline because they were assumed to have only high-level planning capabilities without the ability to output low-level actions. However, this situation has significantly improved, especially given their performance in zero-shot scenarios.
    \item \textbf{Longer navigation trajectories necessitate strong reasoning and planning.} We observe that reasoning-enhanced LMMs outperform other models in the middle and long groups. Their superior performance is attributable to their proficiency in planning, memory, and error correction—key requirements for long-horizon tasks and core strengths of reasoning models.
    \item \textbf{Two types of LMM-based navigation models, agent-based and VLA, exhibit contrasting performances.} Agent-based methods remain the efficient and stable solution. Incorporating depth information to assess position and distance is the most straightforward way to enhance spatial navigation capabilities. This approach, often designed based on human priors, consistently improves performance, albeit with limited generalization. Surprisingly, VLA methods, which align action semantics specifically for navigation data, perform worse than random. This aligns with current consensus: they overfit to in-sample data, with generalization abilities even inferior to the original backbone VLM.
\end{itemize}

\subsection{Critical Decision Bifurcation (CDB)}

We further investigate the relationship between the navigation completion progress and steps, as shown in Figure~\ref{fig:cdb}. An intriguing phenomenon emerges: LLMs exhibit a \textbf{Critical Decision Bifurcation (CDB)} during navigation: Models often make an incorrect decision at a specific node, after which the distance to the goal increases monotonically. This indicates that navigation errors do not accumulate linearly; instead, irreversible path bifurcations are triggered at particular spatial-semantic junctions due to insufficient spatial competence.  
From the curves of four representative models, GPT‑5.1, which possesses stronger reasoning abilities, exhibits a smaller post‑CDB slope, suggesting that reflective reasoning mitigates error propagation. In contrast, Qwen2.5‑VL‑7B and SayNav diverge rapidly, while OpenFly frequently oscillates locally.  

These observations indicate that key decision nodes in each case are particularly informative. Accordingly, we analyze both successful and failed cases at the CDB to infer how far LMMs remain from human-level spatial action.

\subsection{Successful Case Analysis: Reasoning Process}

\begin{figure}[t]
	\centering
	\includegraphics[width=0.9\linewidth]{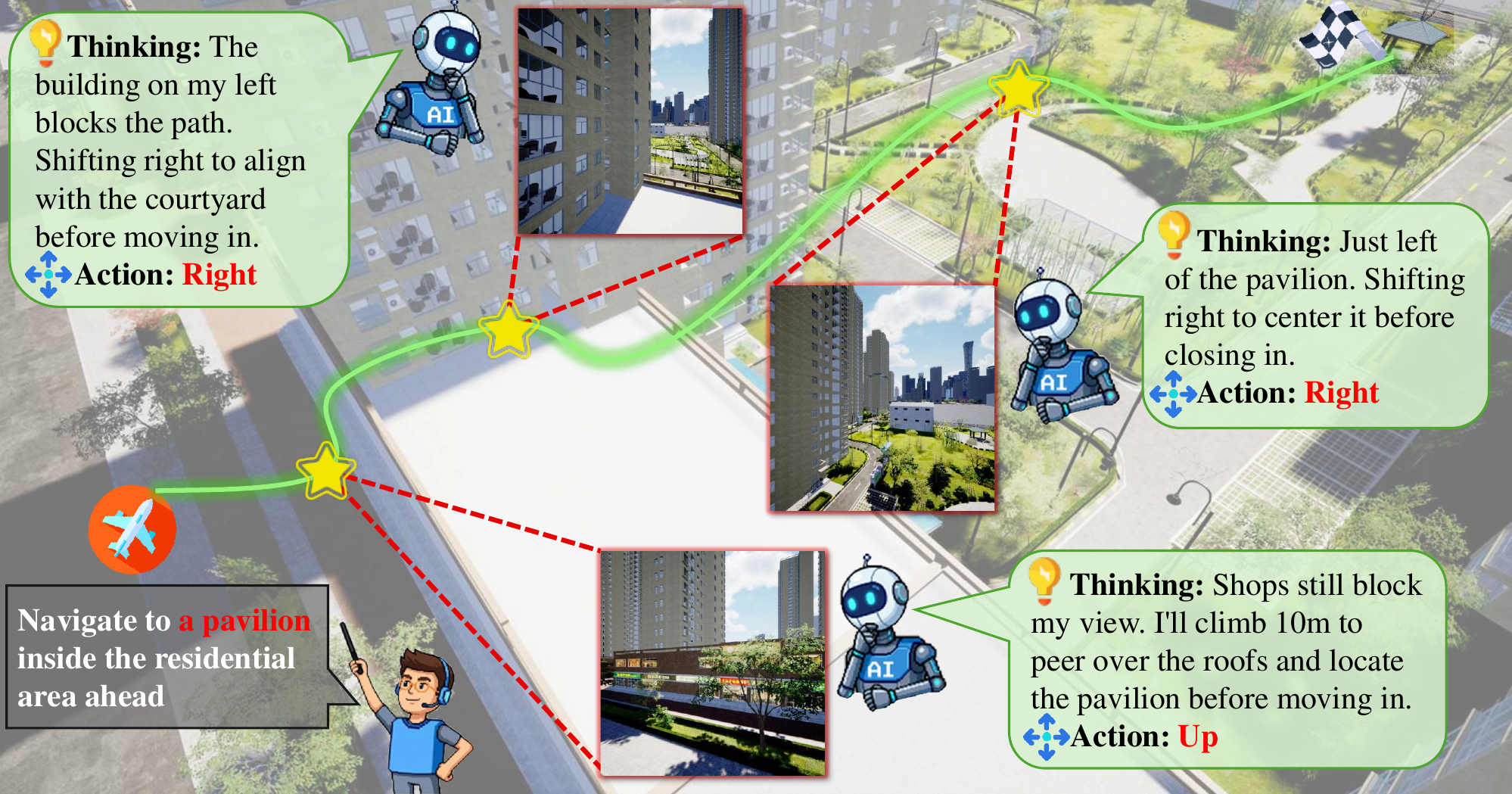}
	\caption{A goal-oriented embodied navigation case: GPT-5.1's trajectory, along with its embodied observations and thought processes at key moments, demonstrated human-like reasoning and action output.}
	\label{fig:goodcase}
\end{figure}

To examine whether LMMs exhibit human-like reasoning and action output, we analyzed the successful navigation cases of various models. A representative case is presented in Figure~\ref{fig:goodcase}, where the star marks CDBs at which other models frequently fail (See Appendix~\ref{appendix:nav_process} for details of other models). In contrast, GPT‑5.1 demonstrates:

\begin{itemize}[leftmargin=*,partopsep=0pt,topsep=0pt]
\setlength{\itemsep}{0pt}
\setlength{\parsep}{0pt}
\setlength{\parskip}{0pt}
    \item Understanding the relationship between semantic instructions and urban elements, such as recognizing that the target is roughly behind the shop ahead at the first node.
    \item Actively adjusting actions to improve understanding. Successful navigation hinges on the interaction between actions and visual observations, rather than merely mapping observations to actions. 
    \item Assessing task progress with historical observations to understand the current state of navigation. 
\end{itemize}

\subsection{Failure Case Analysis: Gaps Between LMMs and Humans}

\begin{figure*}[t]
	\centering
	\includegraphics[width=0.85\linewidth]{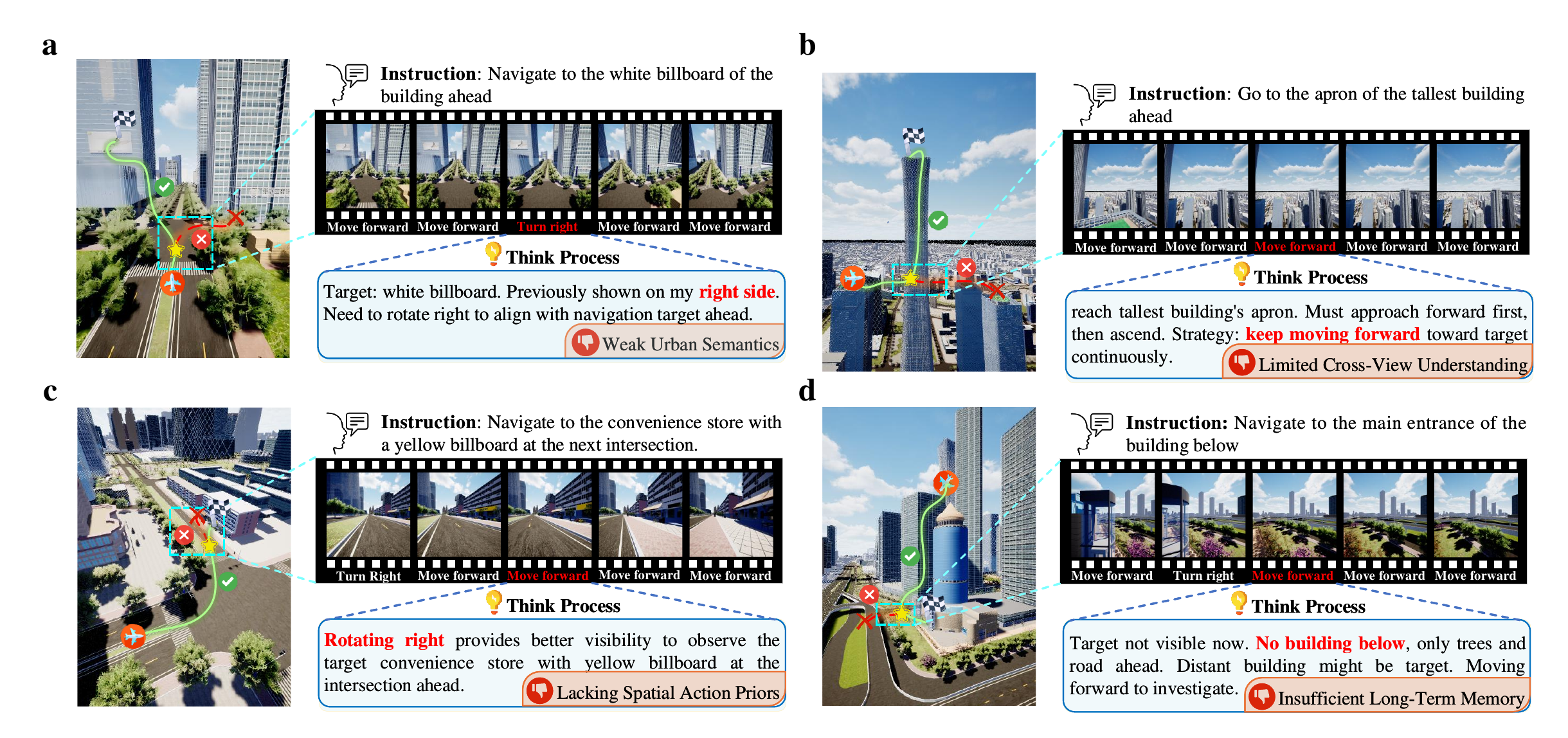}
    \vspace{-5pt}
	\caption{The gaps between LMMs and humans in spatial actions can be summarized into four aspects: \textbf{a.} Insufficient ability in urban semantic perception. \textbf{b.} Limited cross-view scene understanding. \textbf{c.} Lack of common sense about the consequences of actions. \textbf{d.} Deficiency in long-term memory and planning. A third-person perspective view comparing the ground truth and the LMM-navigated trajectories. The green line represents the ground truth path, while the red line depicts the actual route taken by the LMM. Stars denote critical decision bifurcations (CDBs). }
	\label{fig:gap}
\end{figure*}

Although the previous section presents successful navigation cases of LMMs, they fail in the majority of scenarios. Accordingly, we focus on identifying the causes of CDBs to uncover the underlying limitations of LMMs. Through a step-by-step analysis of multiple cases, we summarize four key deficiencies:

\begin{itemize}[leftmargin=*,partopsep=0pt,topsep=0pt]
\setlength{\itemsep}{0pt}
\setlength{\parsep}{0pt}
\setlength{\parskip}{0pt}
    \item \textbf{Insufficient ability in urban geometric perception.} A primary failure mode of LMMs is their erroneous or unstable recognition of urban elements. As shown in Figure~\ref{fig:gap}a, the navigation target in the first-person view is the white billboard, which is visible in the upper-left of the agent's field of view. However, the LMM incorrectly determines that the billboard is on its right, thus issuing an erroneous ``turn right'' command. This action causes the agent to lose sight of the white billboard entirely, triggering a cascade of subsequent reasoning and action errors. 
    \item \textbf{Limited cross-view scene understanding.} The vast scale of open urban environments makes it difficult to achieve comprehensive scene awareness from a single viewpoint. To overcome this, LMMs must integrate continuous visual observations, establishing spatiotemporal consistency across overlapping frames to build a coherent understanding of the scene. Figure~\ref{fig:gap}b illustrates a failure in this process. Initially, the first-person view correctly shows that the building on the left is the tallest in sight. However, after two forward movements, this target building is no longer visible. At this point, the LMM fails to integrate its historical observations and instead erroneously identifies a different building in its current view as the target. 
    \item \textbf{Lack of common sense about the consequences of actions.} A third critical limitation of LMMs is their lack of intuitive world knowledge regarding the physics of motion, specifically how actions alter perception. This deficit is evident in Figure~\ref{fig:gap}c. At the moment marked by a star, the target convenience store with a yellow sign is visible on the right side of the first-person view, but at a suboptimal angle. The LMM correctly identifies the need to adjust its position to center the target in its view. However, instead of executing the logical ``turn right'' action, it outputs ``move forward.'' This demonstrates that LMMs lack an intuitive understanding of how actions will transform egocentric perspective.
    \item \textbf{Deficiency in long-term memory and planning.} A fourth major failure is that deficiencies in long-term memory and planning cause LMMs to gradually forget their initial objective. As illustrated in Figure~\ref{fig:gap}d, the target building is no longer in the agent's field of view due to its movement. Relying solely on its recent memory and the initial instruction, the LMM incorrectly concludes, ``No building below.'' In reality, the building that was initially below the agent is now behind it. This case demonstrates that without an effective memory mechanism, the agent's spatial understanding becomes skewed, leading to goal abandonment during long-horizon actions. 

\end{itemize}

\section{Further Exploration}
\begin{figure*}[t]
    \centering
    \includegraphics[width=0.85\linewidth]{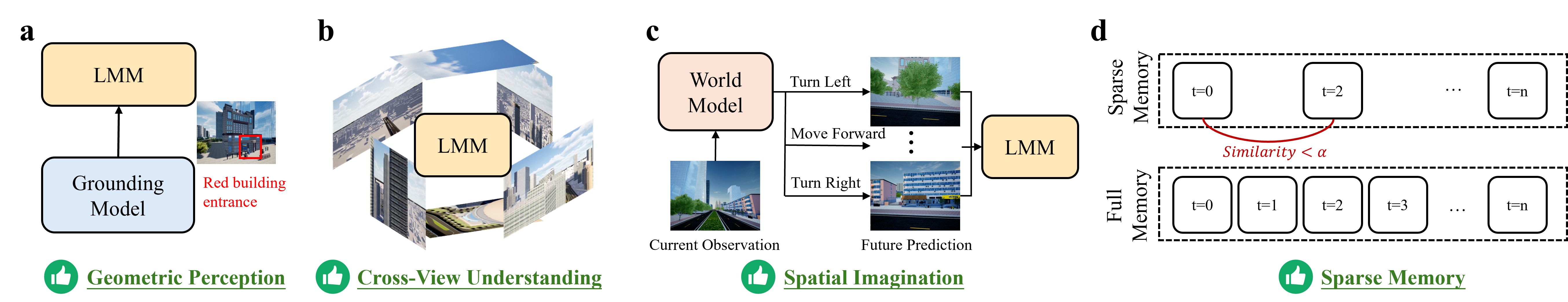}
    \vspace{-5pt}
    \caption{Experimental design to enhance spatial action capability of LMMs: \textbf{a.} Geometric Perception Enhancement. \textbf{b.} Cross-View Understanding Enhancement. \textbf{c.} Spatial Imagination. \textbf{d.} Sparse Memory.}
    \label{fig:four_experiments}
\end{figure*}

\begin{table}[t]
	\caption{Further exploration for LMM improvement.}
	\label{fig:improve}
	\centering
	\setlength{\tabcolsep}{2pt}
	\resizebox{0.85\linewidth}{!}{
		\begin{tabular}{lccc}
			\toprule
			\multirow{2}*{Method}  & \multicolumn{3}{c}{Average}             \\
			\cmidrule(r){2-4}
			~     & SR $\uparrow$      & SPL $\uparrow$  & DTG $\downarrow$ \\ 
			\midrule
Backbone &  14.7 & 12.8 & 120.6 \\
			+ Geometric Perception & $24.2^{\color{green!70!black}\uparrow 9.5}$ & $22.9^{\color{green!70!black}\uparrow 10.1}$ & $138.9^{\color{red}\uparrow 18.3}$  \\
            + Cross-View Understanding & $25.0^{\color{green!70!black}\uparrow 10.3}$ & $16.7^{\color{green!70!black}\uparrow 3.9}$ & $118.6^{\color{green!70!black}\downarrow 2.0}$ \\
            + Spatial Imagination & $19.0^{\color{green!70!black}\uparrow 4.3}$  & $16.7^{\color{green!70!black}\uparrow 3.9}$  & $115.9^{\color{green!70!black}\downarrow 4.7}$  \\
            + Sparse Memory & $17.0^{\color{green!70!black}\uparrow 2.3}$  & $16.3^{\color{green!70!black}\uparrow 3.5}$  & $134.3^{\color{red}\uparrow 13.7}$  \\
			\bottomrule
		\end{tabular}
	}
\end{table}

To both \textbf{validate the identified shortcomings of LMMs and provide insights for the improvement of future LMM architectures and training strategies}, we conduct four targeted experiments, as illustrated in Figure~\ref{fig:four_experiments}. The spatial action capability of LMMs have improved significantly, as listed in Table~\ref{fig:improve}.

\begin{itemize}[leftmargin=*,partopsep=0pt,topsep=0pt]
\setlength{\itemsep}{0pt}
\setlength{\parsep}{0pt}
\setlength{\parskip}{0pt}
    \item \textbf{Geometric Perception Enhancement:} To address the limitations in LMMs' geometric perception capabilities, we employed grounding models to first mark the targets that LMMs focus on before outputting actions. This approach improved navigation success rates by 9.5\%. This indicates that LMMs may lack geometric encoding and alignment, suggesting future exploration of additional geometric encoders or strengthening through training loss enhancements.
    \item \textbf{Cross-View Understanding Enhancement:} To determine the impact of multi-view perspectives on LMM decision-making, we input panoramic views instead of single views during each decision. Explicitly providing cross-view information resulted in significant improvements, indicating that models do not fully comprehend 3D spaces. Future research could focus on enhancing LMMs through cross-view perception understanding.
    \item \textbf{Spatial Imagination:} To enable LMMs to understand the impact of each action on embodied observations, we simulated the principles of world models by explicitly inputting the result of action execution to the LMM. Subsequently, the LMM can decide whether to execute the action or replan its decision. This indicates integrated models that combine understanding and generation may hold more promise for achieving spatial action capabilities~\cite{zhang2025unified}.
    \item \textbf{Sparse Memory:} Due to continuous embodied observations, there is considerable redundancy, yet LMMs often suffer from forgetting issues. We applied a simple rule to store only frames where the field of view (FOV) changes beyond a certain threshold, which also enhanced model performance. 
    Currently, agent-based methods are more effective than LMMs at achieving efficient memory within architectural frameworks.

\end{itemize}
(See Appendix~\ref{appendix:further_exploration} for details.)

\section{Conclusion}
To investigate whether LMMs can act in 3D space like humans, we propose a goal-oriented urban aerial navigation dataset containing 5,037 trajectories with rich semantic annotations. We evaluate 17 representative models on this benchmark, achieving success rates of 4.2\%–34.0\%, far below the 92.0\% human level. By examining navigation completion progress curves, we identify a critical decision bifurcation (CDB) phenomenon, in which navigation errors of LMMs do not grow gradually but instead deviate after a pivotal decision point. From the CDB viewpoint, we derive that LMMs exhibit emerging reasoning and action capabilities but still suffer from four key limitations: insufficient geometric perception, limited cross-view understanding, lack of spatial imagination about action consequences, and weak long-term memory for planning. Further experiments lead to four corresponding directions for improvement. 
These insights outline a path toward bridging the gap between LMMs and spatial intelligence.

\bibliographystyle{ACM-Reference-Format}
\bibliography{ref}

\appendix

\section{Details of Dataset}


The following Figure \ref{fig:data_example} presents the first-person view images of the goal-riented embodied navigation, showing several examples and the navigation tasks. 

The tasks are situated in semantically rich urban environments, where goals are specified using high-level spatial and semantic descriptions (e.g., ``the entrance of the red building on the left front'' or ``the apron of the building on the right side ahead''), rather than explicit coordinates. Successfully completing these tasks therefore requires reasoning over urban semantic elements, including buildings, roads, intersections, vegetation, and functional areas.

Moreover, the goal-oriented formulation of the benchmark introduces an explicit reasoning component, as agents must infer goal locations from relational and descriptive cues and maintain consistency between local observations and global intent throughout the trajectory.

Finally, the benchmark covers both horizontal and vertical motion, as illustrated by cases involving street-level navigation, intersection traversal, rooftop and apron approaches, and altitude changes in dense city blocks. This design encourages models to integrate spatial reasoning across different motion modes, reflecting the complexity of real-world urban navigation.

\begin{figure*}[t]
	\centering
	\includegraphics[width=\linewidth]{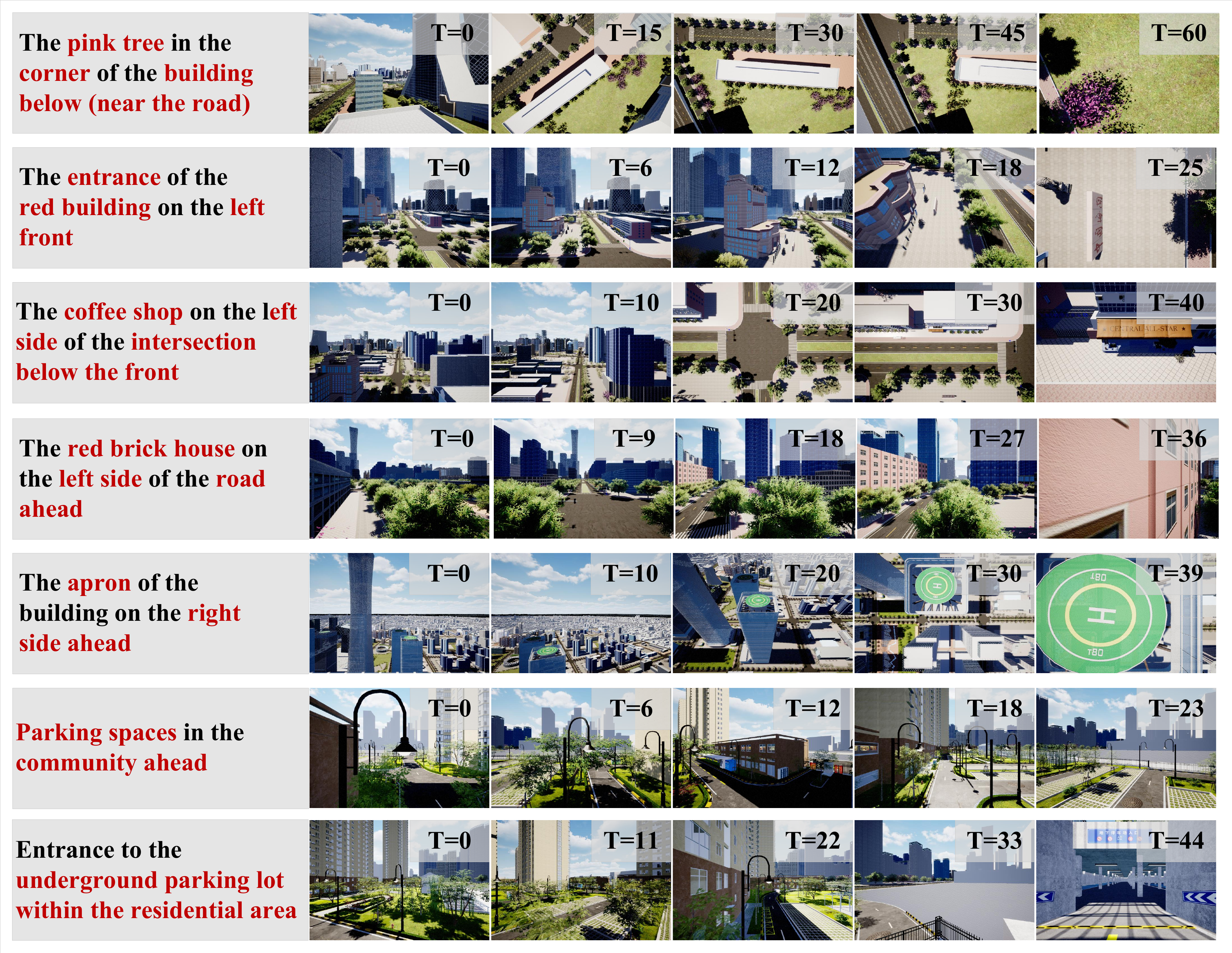}
	\caption{Navigation dataset examples.}
	\label{fig:data_example}
\end{figure*}

\section{Additional Experimental Settings}
\subsection{Brief Introduction on Baselines}
\label{sec:basline_app}

We introduce them one by one:

\textbf{Gemini-2.0-Flash}. Released on Februray 5, 2025, with an API service, Gemini-2.0-Flash is a lightweight model in the Gemini series. It supports 1,048,576 input tokens, an output token limit of 8,192, and a maximum image input of 3,000. 

\textbf{Gemini-2.5-Flash}. Released on June 17, 2024, with an API service, Gemini 2.5 Flash is the first Flash model with thinking capabilities in the Gemini series. It supports 1,048,576 input tokens, an output token limit of 65,535, and a maximum image input of 3,000. 

\textbf{Gemini-2.5-Pro}. Released on June 17, 2024, with an API service, Gemini-2.5-Pro is currently the model with the strongest reasoning ability in the Gemini series. It supports 1,048,576 input tokens, an output token limit of 65,535, and a maximum image input of 3,000. 

\textbf{GLM-4.6V}. Released on December 8, 2025, with an API service, GLM-4.6V is a 106B-parameter visual understanding model with a 128K context length. In this experiment, we set the resolution to 560 * 560 and adopted the model's default settings. 

\textbf{OpenAI o4-mini}. Released on April 16, 2025, OpenAI o4-mini is a compact version of GPT-4o, designed for faster inference with a 0.2M context length. It has an output token limit of 100,000 . In this experiment, we set the resolution to 560 * 560 and adopted the model's default settings.

\textbf{GPT-4o}. Released on May 13, 2024, with an API service, GPT-4o is a multimodal LMM from OpenAI, featuring a 128K context length. It has an output token limit of 16,384 . In this experiment, we set the resolution to 560 * 560 and adopted the model's default settings.

\textbf{GPT-4.1}. Released on April 14, 2025, with an API service, GPT-4.1 is a multimodal LMM from OpenAI, featuring a 1M context length. It has an output token limit of 32,768. In this experiment, we set the resolution to 560 * 560 and adopted the model's default settings.

\textbf{GPT-5.1}. Released on November 13, 2025, with an API service, GPT-4o is the latest multimodal LMM from OpenAI, featuring a 0.4M context length. It has an output token limit of 128,000. In this experiment, we set the resolution to 560 * 560 and adopted the model's default settings.

\textbf{Qwen-VL-Max-latest}. Released in April 2024, Qwen-VL-Max-latest is a advanced model in the Qwen-VL series, supporting multimodal tasks with a 128K context length. It has an input token limit of 128K, an output token limit of 8192. In this experiment, we set the resolution to 560 * 560 and adopted the model's default settings.

\textbf{Qwen2.5-VL-3B-Instruct}. Released on January 26, 2025, Qwen2.5-VL-3B-Instruct is a lightweight instruct-tuned model with 3B parameters and a 32,768 tokens context length. In this experiment, we set the resolution to 560 * 560 and adopted the model's default settings.

\textbf{Qwen2.5-VL-7B-Instruct}. Released on January 26, 2025, Qwen2.5-VL-7B-Instruct is a mid-sized instruct-tuned model with 7B parameters and a 32,768 tokens context length. In this experiment, we set the resolution to 560 * 560 and adopted the model's default settings.

\textbf{Qwen3-VL-Plus}. Released on December 19, 2025, with an API service, Qwen3-VL-Plus is the most advanced model with 258,048 tokens context length. In this experiment, we set the resolution to 560 * 560 and adopted the model's default settings.

\subsection{Implementation Details}\label{Implementation_Details}
To evaluate the direct-action capabilities of LMMs in spatial environments, we employ a straightforward input-output paradigm. For each timestep, the model's input consists of the current RGB observation combined with a historical memory of past moments. Each historical moment includes the embodied visual observation, the action taken, and a textual rationale for that action. The model's output is similarly structured: a direct action command and its corresponding rationale.
To manage memory under the token constraints and computational overhead of most LMMs, we utilize a fixed-size memory window of 30 moments. This memory is populated via uniform sampling of the history, with the crucial first and most recent frames always preserved. All RGB observations have a resolution of 560x560 and a 90-degree Field of View. Other settings for drones remain as the AirSim default~\cite{airsim2017fsr}. 

\subsection{Cost of Proprietary Models}
In Table \ref{tab:evaluation_cost}, we show the tokens and cost of our evaluation of proprietary models.

\begin{table}[h]
\centering
\caption{Evaluation Cost of Different Models}
    \setlength{\tabcolsep}{2pt} 
    \resizebox{\linewidth}{!}{
\begin{tabular}{lccc}
\hline
\textbf{Model} & \textbf{Input Cost (\$)} & \textbf{Output Cost (\$)} & \textbf{Total Cost (\$)} \\ \hline
Gemini-2.0-Flash & 157.70 & 41.03 & 198.73 \\ 
Gemini-2.5-Flash & 315.41 & 170.94 & 486.35 \\ 
Gemini-2.5-Pro & 1314.20 & 683.75 & 1997.95 \\ 
GLM-4.6V & 150.62 & 29.39 & 180.01 \\ 
OpenAI o4-mini & 1156.49 & 300.85 & 1457.35 \\ 
GPT-4o & 2628.40 & 683.75 & 3312.15 \\ 
GPT-4.1 & 2102.72 & 547.00 & 2649.72 \\ 
GPT-5.1 & 1314.20 & 683.75 & 1997.95 \\ 
Qwen-VL-Max-latest & 482.00 & 125.39 & 607.39 \\ 
Qwen2.5-VL-3B-Instruct & 180.75 & 35.27 & 216.01 \\ 
Qwen2.5-VL-7B-Instruct & 301.25 & 48.98 & 350.23 \\ 
Qwen3-VL-Plus & 301.25 & 195.92 & 497.17 \\ 
\textbf{Total} & \textbf{10404.98} & \textbf{3546.01} & \textbf{13950.99} \\ \hline
\end{tabular}
}
\label{tab:evaluation_cost}
\end{table}

\subsection{Prompt}

\begin{figure*}[t]
	\centering
	\includegraphics[width=\linewidth]{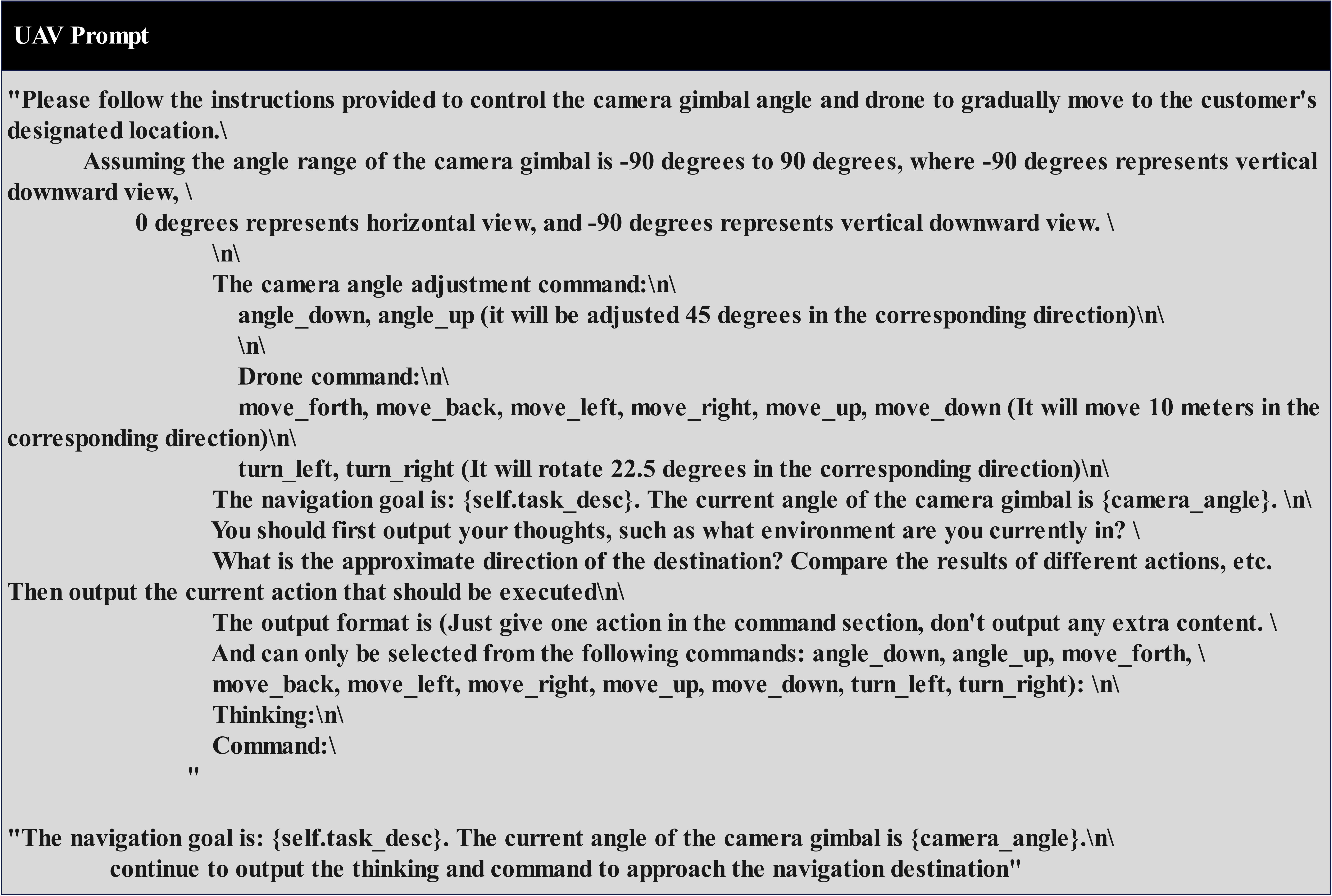}
	\caption{Navigation prompt details.}
	\label{fig:prompts}
\end{figure*}

The following Figure~\ref{fig:prompts} provides a detailed description of our prompt design. At each timestep, the large model is provided with a structured natural-language prompt that specifies (i) the navigation goal described in free-form language (e.g., building entrances, functional areas, or semantic landmarks), (ii) the current camera gimbal angle, and (iii) a fixed set of discrete control commands for both camera orientation and drone motion.

As illustrated in the prompt, the camera gimbal angle ranges from $-90^\circ$ (vertical downward view) to $0^\circ$ (horizontal view), and the model can explicitly issue \texttt{angle\_up} or \texttt{angle\_down} commands to adjust its visual perspective. Drone motion is discretized into translational actions (\texttt{move\_forth}, \texttt{move\_back}, \texttt{move\_left}, \texttt{move\_right}, \texttt{move\_up}, \texttt{move\_down}), each corresponding to a fixed displacement, as well as rotational actions (\texttt{turn\_left}, \texttt{turn\_right}) with a fixed rotation angle.

The prompt requires the model to analyze the surrounding environment, infer the relative direction of the target, and select exactly one executable command at each step. Since the navigation goal is typically not directly observable from the initial viewpoint, successful task completion relies on multi-step reasoning over both horizontal and vertical movements, as well as explicit viewpoint adjustment. This design ensures that benchmark performance reflects the model's ability to integrate language grounding, spatial reasoning, and sequential decision-making, rather than low-level control execution.



\section{Details of Case Analysis}
\label{appendix:nav_process}
To facilitate a direct comparison of different models on the same case, we additionally list the detailed action trajectories of five representative models for the case discussed in Section~5.4 below.

For the GPT‑5.1, it follows a predominantly forward‑driven strategy with minimal lateral corrections, relying on early altitude gain to overcome occlusions caused by trees and low‑rise shop buildings. Once elevated above the foreground structures, the model quickly establishes a stable global view of the residential courtyard and consistently advances toward the pavilion with little backtracking. The trajectory is largely monotonic, characterized by sustained forward motion and only a single late-stage heading correction when the target is slightly overshot. This behavior suggests stronger global spatial alignment and more efficient long-horizon planning, although fine-grained final positioning still requires minor lateral adjustment near the target.

For the o4-mini, it successfully reaches the target pavilion through a long-horizon sequence of viewpoint adjustment, lateral repositioning, and forward advancement. The model explicitly alternates between camera gimbal control and spatial movements to recover visibility when the target is initially occluded by foreground structures, such as rows of shops and utility buildings. Notably, o4-mini repeatedly relies on vertical ascent and lateral shifts to obtain a higher vantage point and resolve partial occlusions, before committing to forward motion into the courtyard. While the model eventually aligns with and reaches the target, the trajectory exhibits redundant corrective actions and oscillatory lateral movements near the goal, indicating inefficiencies in fine-grained spatial alignment despite successful semantic grounding.

Gemini-2.5-Flash is able to progressively approach the target pavilion through a sequence of forward movements, elevation adjustments, and viewpoint alignment. Once the pavilion becomes visible, the model successfully centers it in the field of view and navigates to a position directly above the target.
However, the failure emerges during the final approach phase. When descending toward the pavilion, the model repeatedly encounters dense foliage that occludes the view. Lacking an explicit understanding of vertical geometry and obstacle clearance, the model enters a repetitive loop of \texttt{move down} and \texttt{move up} actions. Although the pavilion remains visible from an overhead perspective, the model fails to anticipate that descending further will consistently result in visual obstruction.
This behavior highlights two core limitations. First, the model lacks fine-grained geometric perception, particularly regarding vertical clearance and object height relationships. Second, it does not possess spatial imagination—the ability to predict the outcome of an action before execution. Without anticipating that descending would lead to occlusion, the model cannot replan a safer or more lateral descent strategy. This failure case directly motivates the need for geometric grounding and world-model-based action anticipation, as explored in our targeted experiments.

Qwen2.5-VL-7B fails at an earlier stage of navigation. As shown in Table~Y, the model exhibits extensive exploratory behavior, including repeated forward movements, camera gimbal adjustments, and frequent turns. Despite prolonged exploration, the model never successfully localizes the pavilion.
A key issue lies in the model's inability to maintain a coherent spatial belief about the target once it is not directly visible. The navigation process is dominated by local visual cues, such as storefronts, corridors, and interior spaces, causing the model to drift away from the intended outdoor search space. Notably, the model enters indoor environments and even lands unintentionally, indicating a breakdown in high-level spatial constraints and task grounding.
Moreover, Qwen2.5-VL-7B lacks cross-view understanding and global orientation. The model treats each observation largely independently, resulting in repeated revisiting of similar viewpoints and ineffective exploration. The absence of a persistent spatial memory or map-like representation prevents the model from reasoning about unexplored regions or returning to promising directions.

OpenFly fails to reach the target pavilion. Throughout the episode, the agent primarily executes repeated forward movements at ground level, followed by local heading changes near the shopfronts, without attempting to gain altitude or traverse over the low-rise buildings that block the line of sight to the inner courtyard. As a result, OpenFly remains confined to the street-level space and eventually enters a dead-end corridor between buildings. This behavior indicates a strong bias toward ground-constrained navigation and a lack of proactive exploration strategies, such as vertical ascent or building overflight, which are necessary to resolve large-scale occlusions in this scenario.

\section{Details of Further Exploration}
\label{appendix:further_exploration}

We provide additional implementation details and methodological explanations for the four targeted experiments introduced in the main paper. These experiments are designed to further validate the identified shortcomings of LMMs in embodied navigation and to offer insights into potential architectural and training improvements.

\subsection{Geometric Perception Enhancement}
\label{app:geometric_perception}

LMMs often exhibit limited geometric perception and spatial alignment capabilities, leading to unstable or inaccurate low-level action predictions. To isolate high-level semantic reasoning from geometric execution, we adopt a two-stage navigation pipeline in which the LMM is only responsible for identifying the navigation target in the current observation.

Concretely, given the current egocentric view, the LMM outputs the name of the object or region it intends to navigate toward, rather than directly predicting a navigation action. This object name is then passed to a grounding algorithm based on GroundingDINO~\cite{liu2024grounding}, which returns the corresponding 2D location of the target in the image. Based on the grounded position, a deterministic controller first outputs an action that centers the target in the field of view, and then issues a forward movement action.

By removing the requirement for the LMM to generate fine-grained geometric actions, this approach reduces action noise and improves spatial alignment. The resulting performance gain suggests that current LMMs lack explicit geometric encoding, motivating future work on incorporating geometric representations or geometry-aware objectives into LMM architectures.

\subsection{Cross-View Understanding Enhancement}
\label{app:cross_view}

A common failure mode of LMM-based navigation arises from limited field-of-view observations. When the target leaves the current view, the model often loses its orientation and fails to recover the correct navigation direction, indicating a lack of awareness of objects outside the visible frame.

To alleviate this issue, we replace single-view inputs with a multi-view observation setting. At each decision step, the LMM is provided with six images corresponding to evenly spaced viewpoints, each with a $90^\circ$ field of view, jointly covering the full $360^\circ$ surroundings. By explicitly exposing cross-view information, the model is better able to maintain target awareness and spatial consistency.

Although this setup does not provide explicit 3D reconstruction, it significantly reduces navigation failures caused by target disappearance and orientation drift, highlighting the importance of cross-view perception for embodied decision-making.

\subsection{Spatial Imagination}
\label{app:spatial_imagination}

Human spatial cognition relies on the ability to anticipate the consequences of actions, a capability commonly attributed to world models. To emulate this behavior, we introduce a spatial imagination mechanism that allows the LMM to reason about the potential outcomes of candidate actions before execution.

We assume the existence of a world model that predicts the next observation given a state-action pair. Instead of training such a model, we simulate this functionality by directly querying the simulator to obtain the ground-truth post-action observation. For each candidate action proposed by the LMM, the resulting observation is fed back to the model, which then decides whether to execute the action or replan. This process iterates until a final action is selected.

This experiment demonstrates that integrating action outcome anticipation into the decision loop can substantially improve spatial reasoning and navigation performance, suggesting that unified models combining perception, imagination, and action generation are a promising direction for future research~\cite{zhang2025unified}.

\subsection{Sparse Memory}
\label{app:sparse_memory}

Embodied navigation produces continuous streams of highly redundant observations. However, LMMs typically process navigation by repeatedly ingesting the entire history of observations, reasoning steps, and actions, which leads to excessive token consumption and degraded performance due to memory overload.

To address this issue, we introduce a sparse memory mechanism based on field-of-view (FOV) novelty. Specifically, a new observation is added to the memory buffer only if its FOV overlap with recent stored observations is below a predefined threshold. In this way, only observations that contribute sufficiently new visual information are retained.

This strategy effectively reduces token usage while minimizing redundant memory content, leading to more stable decision-making and improved navigation performance. Nonetheless, agent-based systems with dedicated memory architectures remain more effective at long-term memory management, highlighting an important gap between current LMM-based approaches and modular embodied agents.

\begin{table*}[h]
\centering
\footnotesize
\caption{Action Process of GPT-5.1}

\begin{tabular}{p{0.8cm} p{1.6cm} p{12.0cm} p{1.5cm}}
\toprule
Step & Obs & Thinking & Action \\
\midrule

{\centering 0\par} &
{\centering \raisebox{-0.9\height}{\includegraphics[width=1.0cm]{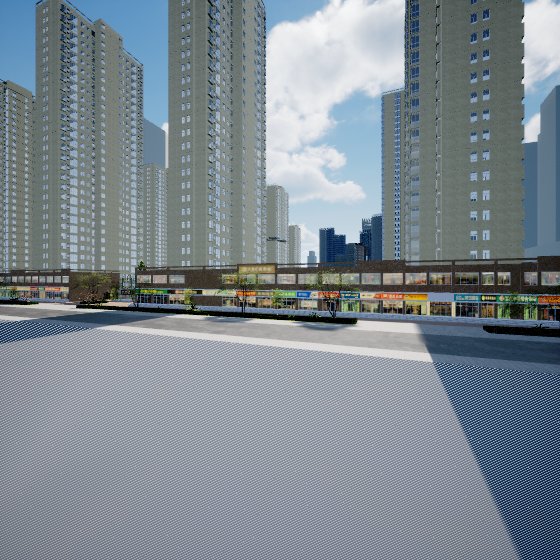}}\par} &
\parbox[t]{12.0cm}{\vspace{0pt}\scriptsize We appear to be in or near a residential/commercial complex with tall apartment buildings and a low-rise row of shops ahead. The pavilion is said to be ``inside the residential area ahead,'' so it's likely further forward among or behind the front row of buildings. Adjusting the gimbal down will give better situational awareness before advancing.
}
 &
angle down \\

{\centering 1\par} &
{\centering \raisebox{-0.9\height}{\includegraphics[width=1.0cm]{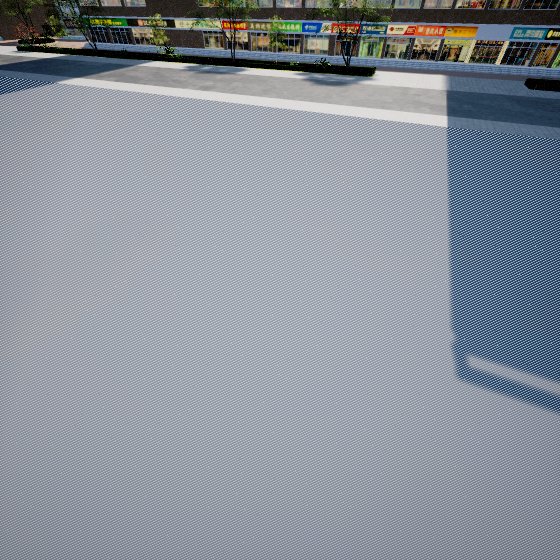}}\par} &
\parbox[t]{12.0cm}{\vspace{0pt}\scriptsize The drone is in a residential/commercial plaza area. The camera is tilted down at -45\textdegree, showing mostly the ground near the drone. The pavilion is said to be inside the residential area ahead. Advancing forward will bring us closer to the inner area where a pavilion is likely located.
}
 &
move forth \\

{\centering 2\par} &
{\centering \raisebox{-0.9\height}{\includegraphics[width=1.0cm]{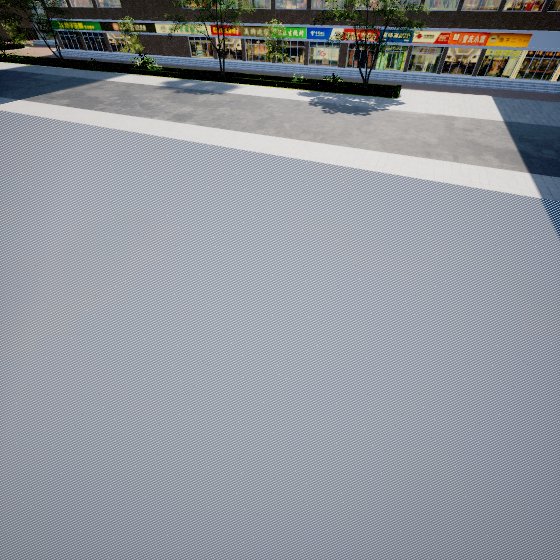}}\par} &
\parbox[t]{12.0cm}{\vspace{0pt}\scriptsize We are moving across an open paved area toward the row of shops and trees that mark the edge of the residential zone. The pavilion should be further inside this area. The camera at -45\textdegree{} gives a good view of the ground and the base of the buildings ahead. We should continue flying straight forward.
}
 &
move forth \\

{\centering 3\par} &
{\centering \raisebox{-0.9\height}{\includegraphics[width=1.0cm]{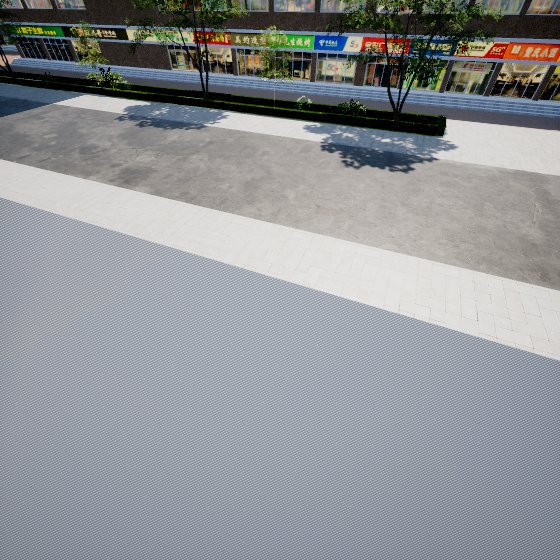}}\par} &
\parbox[t]{12.0cm}{\vspace{0pt}\scriptsize We are now close to the edge of the open plaza, with a walkway, planting strip, and trees ahead. The pavilion should be further inside the residential area, likely beyond this tree line. Continuing straight forward will bring us into the landscaped zone where a pavilion is more likely to appear.
}
 &
move forth \\

{\centering 4\par} &
{\centering \raisebox{-0.9\height}{\includegraphics[width=1.0cm]{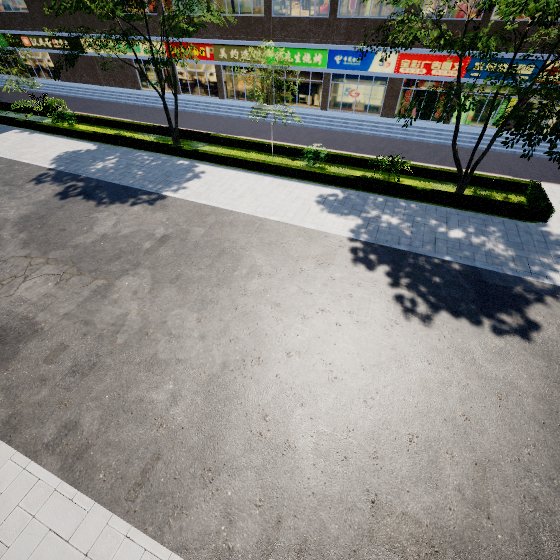}}\par} &
\parbox[t]{12.0cm}{\vspace{0pt}\scriptsize We are now just in front of the tree-lined planting strip and walkway near the shopfronts. The pavilion should be further into the residential interior. We should continue flying straight ahead over the walkway and closer to the buildings, maintaining the current gimbal angle for good ground visibility.
}
 &
move forth \\

{\centering 5\par} &
{\centering \raisebox{-0.9\height}{\includegraphics[width=1.0cm]{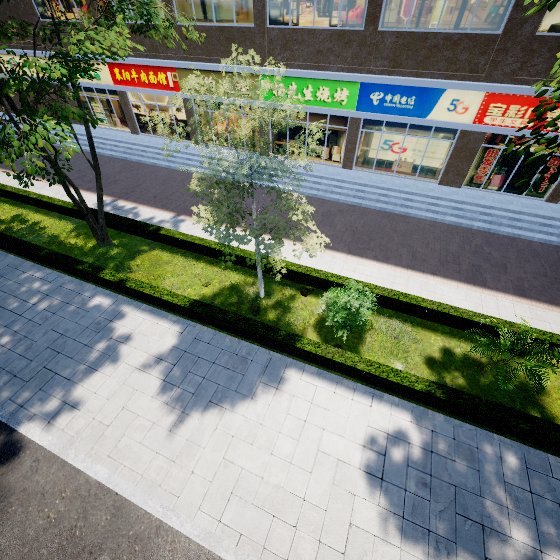}}\par} &
\parbox[t]{12.0cm}{\vspace{0pt}\scriptsize We have reached the sidewalk and green belt directly in front of the shops; trees are now close beneath us. To avoid getting too close to tree canopies while continuing to advance, it's safer to gain a bit more altitude first, then proceed forward afterward.
}
 &
move up \\

{\centering 6\par} &
{\centering \raisebox{-0.9\height}{\includegraphics[width=1.0cm]{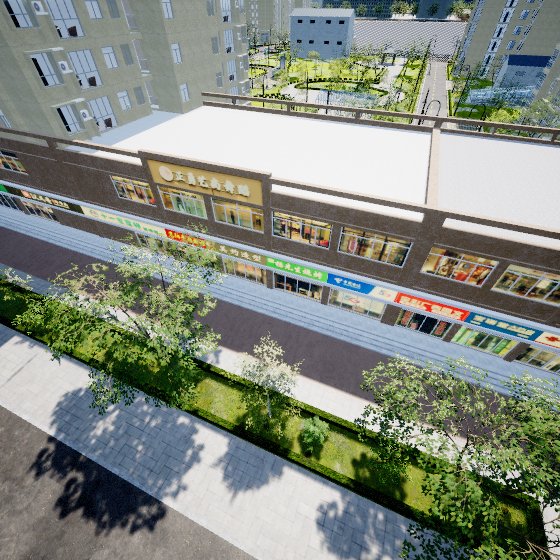}}\par} &
\parbox[t]{12.0cm}{\vspace{0pt}\scriptsize We are now above the trees and can see over the row of shops into the residential courtyards behind them. There is a landscaped area with paths, greenery, and small structures consistent with where a pavilion might be located. We should continue moving forward over the roof toward the inner courtyard.
}
 &
move forth \\

{\centering 7\par} &
{\centering \raisebox{-0.9\height}{\includegraphics[width=1.0cm]{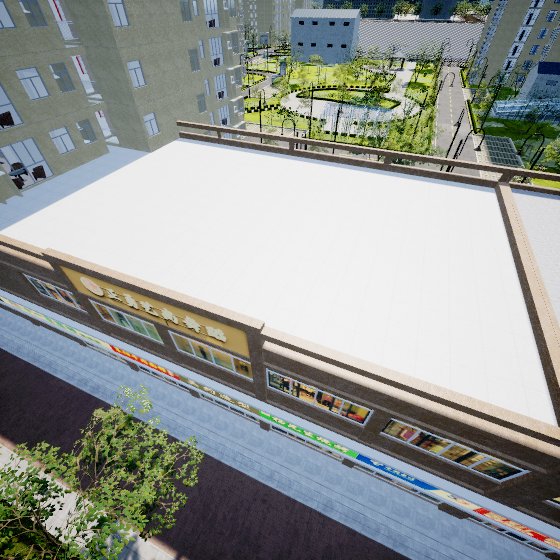}}\par} &
\parbox[t]{12.0cm}{\vspace{0pt}\scriptsize We are above the commercial building, looking slightly down. Behind the building, a landscaped residential courtyard with paths, trees, and a circular pond is visible. The pavilion is likely located within that courtyard near the water feature. We should continue flying straight forward over the roof toward the courtyard.
}
 &
move forth \\

{\centering 8\par} &
{\centering \raisebox{-0.9\height}{\includegraphics[width=1.0cm]{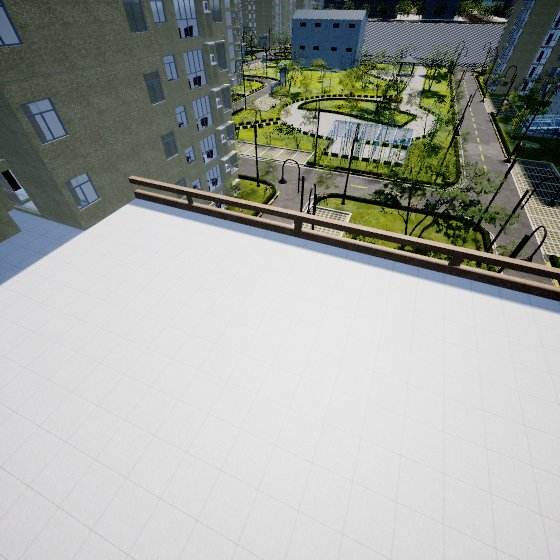}}\par} &
\parbox[t]{12.0cm}{\vspace{0pt}\scriptsize We are above the roof edge of the commercial building, looking down into the residential courtyard ahead. In the courtyard, there is a circular pond and a rectangular roofed structure that appears to be the pavilion. We should continue flying forward into the courtyard airspace.
}
 &
move forth \\

{\centering 9\par} &
{\centering \raisebox{-0.9\height}{\includegraphics[width=1.0cm]{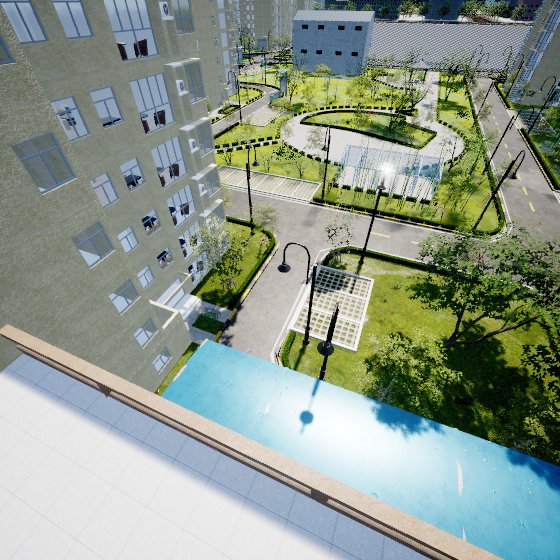}}\par} &
\parbox[t]{12.0cm}{\vspace{0pt}\scriptsize We are now beyond the commercial roof edge, overlooking the inner residential courtyard. Ahead, there is a landscaped area with a circular pond and a rectangular roofed structure that appears to be the pavilion. We should continue moving straight forward over the open space and paths.
}
 &
move forth \\

{\centering 10\par} &
{\centering \raisebox{-0.9\height}{\includegraphics[width=1.0cm]{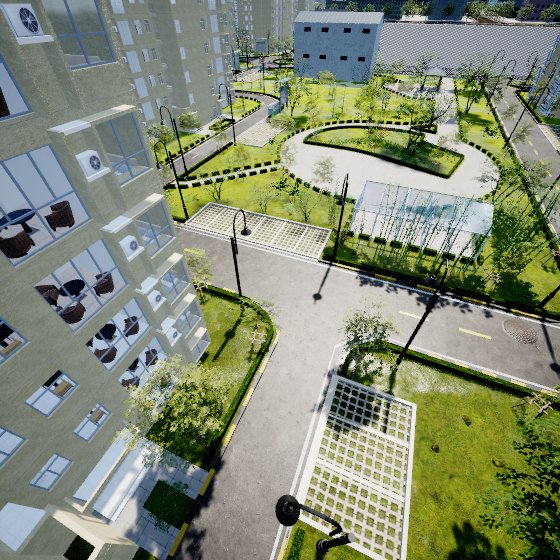}}\par} &
\parbox[t]{12.0cm}{\vspace{0pt}\scriptsize We are now over the internal road of the residential courtyard, with the landscaped park clearly ahead. The rectangular roofed structure just right of the circular path area appears to be the pavilion. Continuing straight forward will bring us closer into the park area and nearer to the pavilion.
}
 &
move forth \\

{\centering 11\par} &
{\centering \raisebox{-0.9\height}{\includegraphics[width=1.0cm]{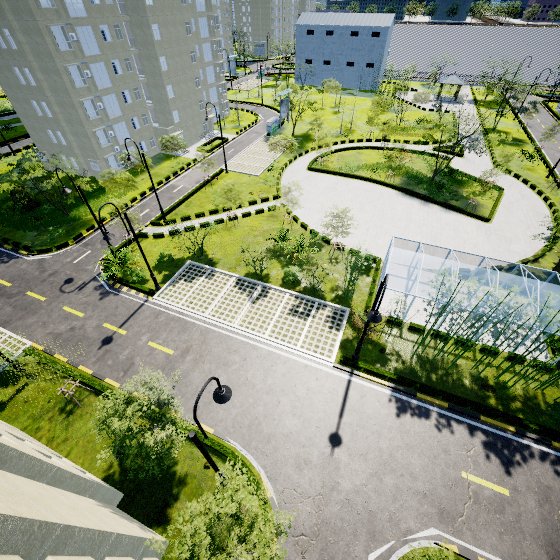}}\par} &
\parbox[t]{12.0cm}{\vspace{0pt}\scriptsize We are above an internal road, facing the central green space. The pavilion with a transparent/roofed structure is clearly visible ahead on the right side of the park. We should continue moving forward toward the park; from there we can make minor lateral adjustments if needed.
}
 &
move forth \\

{\centering 12\par} &
{\centering \raisebox{-0.9\height}{\includegraphics[width=1.0cm]{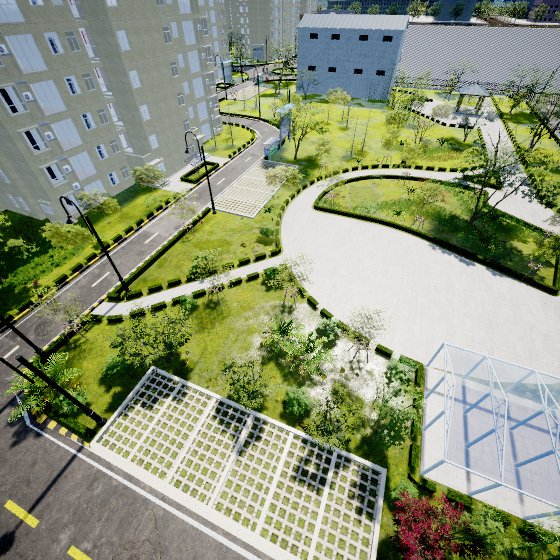}}\par} &
\parbox[t]{12.0cm}{\vspace{0pt}\scriptsize We are now at the edge of the central green area, with curved walking paths and lawns below. The glass-roofed pavilion is clearly visible in the lower-right area of the view. Moving straight forward into the plaza will bring us closer over the open paved area.
}
 &
move forth \\

{\centering 13\par} &
{\centering \raisebox{-0.9\height}{\includegraphics[width=1.0cm]{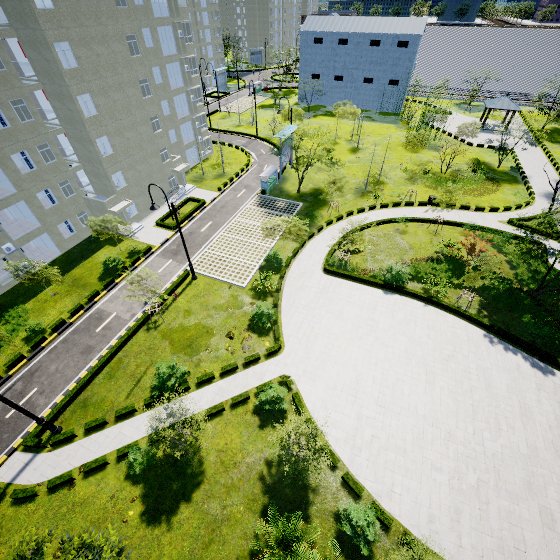}}\par} &
\parbox[t]{12.0cm}{\vspace{0pt}\scriptsize We've moved into the central plaza area; the paved open space is directly below, and the pavilion (glass-roofed structure) is now slightly behind us to the lower-right. To reacquire and center the pavilion, we should adjust our heading to the right.
}
 &
turn right \\

{\centering 14\par} &
{\centering \raisebox{-0.9\height}{\includegraphics[width=1.0cm]{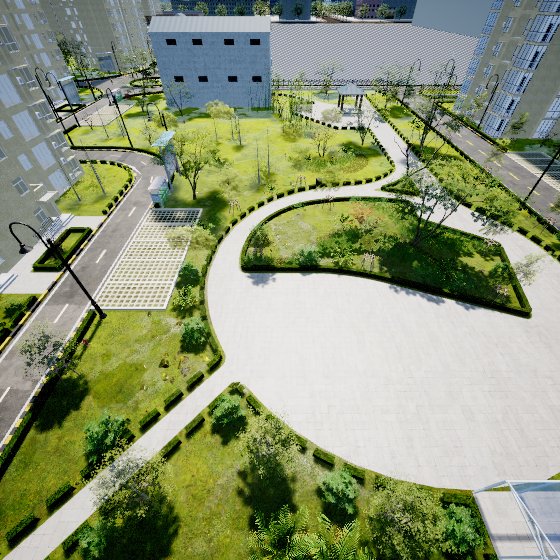}}\par} &
\parbox[t]{12.0cm}{\vspace{0pt}\scriptsize We are over the central plaza, looking toward the landscaped park. The pavilion is ahead-right: a roofed structure near the top-right area, just beyond the curved paths. There are no immediate obstacles between us and the pavilion. We should move straight forward toward it.
}
 &
move forth \\

{\centering 15\par} &
{\centering \raisebox{-0.9\height}{\includegraphics[width=1.0cm]{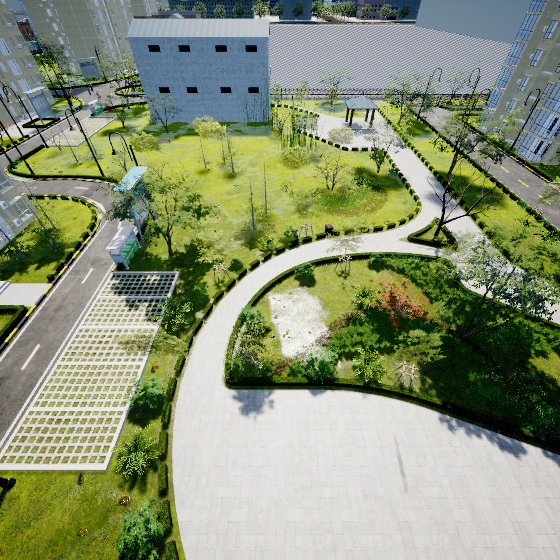}}\par} &
\parbox[t]{12.0cm}{\vspace{0pt}\scriptsize We are above the central plaza and moving toward the inner green park. The pavilion is ahead-right, visible as the roofed structure near the upper-right area close to the path. Continuing straight will bring us closer to the pavilion.
}
 &
move forth \\

{\centering 16\par} &
{\centering \raisebox{-0.9\height}{\includegraphics[width=1.0cm]{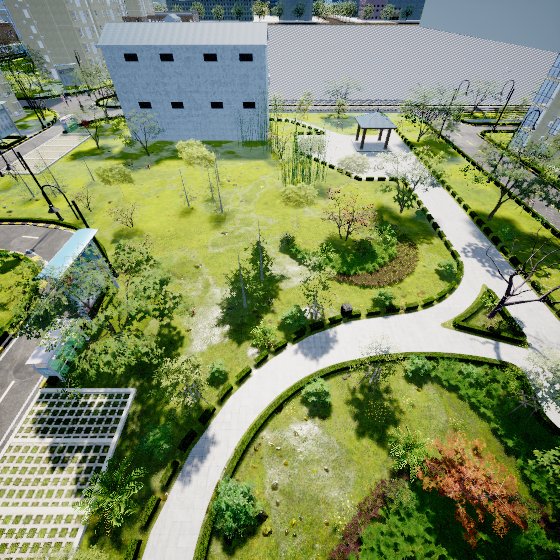}}\par} &
\parbox[t]{12.0cm}{\vspace{0pt}\scriptsize We are now directly above the main green area of the courtyard. The pavilion---an open, roofed structure with four pillars---is clearly visible ahead and slightly to the right. Our heading already points roughly toward it. We should continue flying straight forward.
}
 &
move forth \\

\bottomrule
\end{tabular}
\end{table*}

\begin{table*}[h]
\centering
\footnotesize
\caption{Action Process of GPT 5.1}

\begin{tabular}{p{0.8cm} p{1.6cm} p{12.0cm} p{1.5cm}}
\toprule
Step & Obs & Thinking & Action \\
\midrule

{\centering 17\par} &
{\centering \raisebox{-0.9\height}{\includegraphics[width=1.0cm]{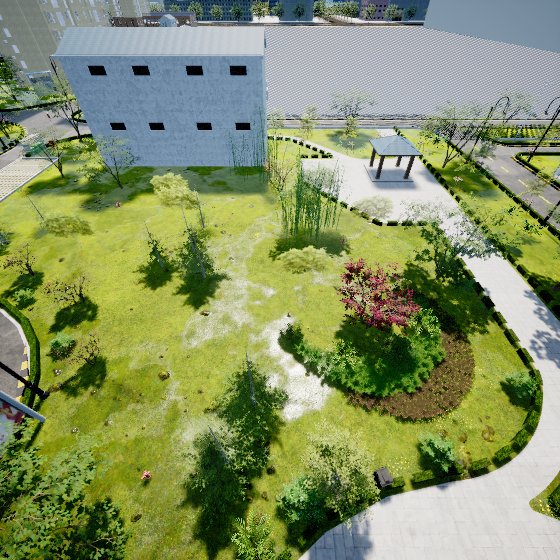}}\par} &
\parbox[t]{12.0cm}{\vspace{0pt}\scriptsize The pavilion is now clearly visible ahead and slightly to the right, standing on a paved area near the path. We are at a good altitude with no high obstacles between us and the pavilion. Continuing to fly straight forward will bring us directly closer to the pavilion.
}
 &
move forth \\

{\centering 18\par} &
{\centering \raisebox{-0.9\height}{\includegraphics[width=1.0cm]{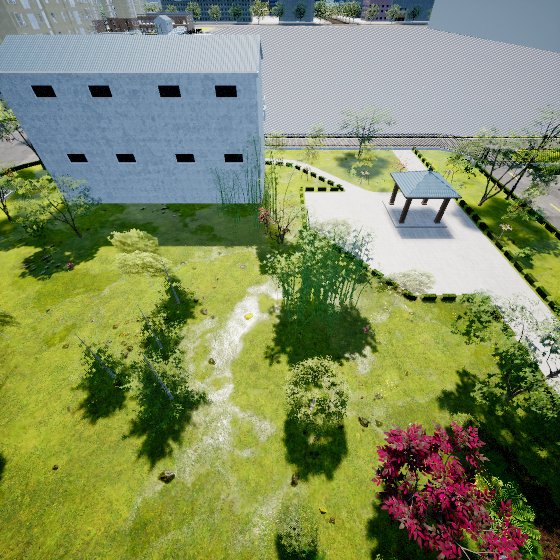}}\par} &
\parbox[t]{12.0cm}{\vspace{0pt}\scriptsize The pavilion is now slightly ahead and to the right, clearly visible on the paved area. We are nearly above the grassy area in front of it. Moving straight will bring us closer while still maintaining safe clearance. We should continue advancing forward.
}
 &
move forth \\

{\centering 19\par} &
{\centering \raisebox{-0.9\height}{\includegraphics[width=1.0cm]{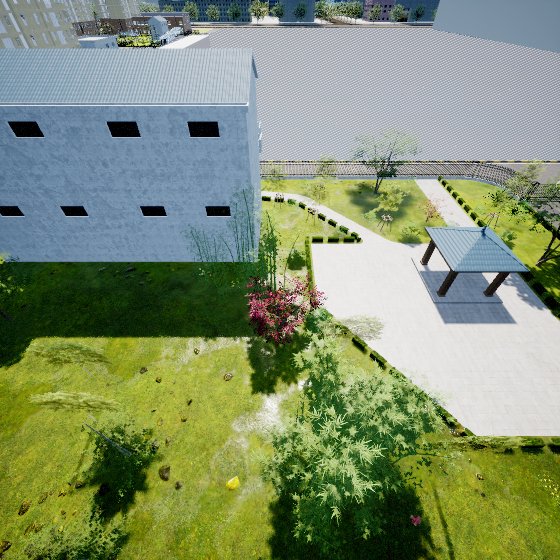}}\par} &
\parbox[t]{12.0cm}{\vspace{0pt}\scriptsize We are now very close to the pavilion, which is ahead-right on the paved area. To position the drone more directly over or beside the pavilion as the final navigation target, we should move slightly closer along our current heading.
}
 &
move forth \\

{\centering 20\par} &
{\centering \raisebox{-0.9\height}{\includegraphics[width=1.0cm]{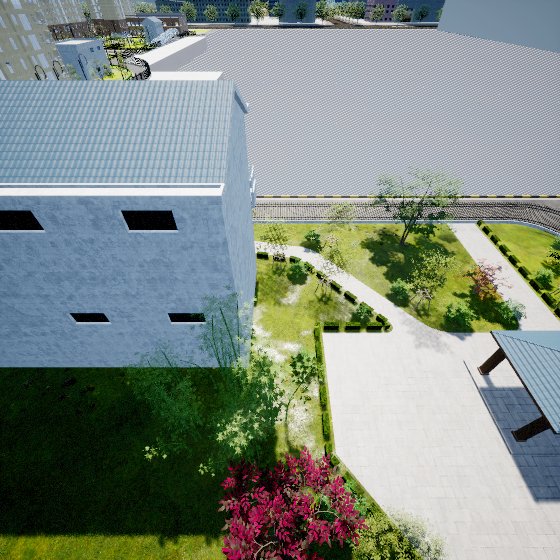}}\par} &
\parbox[t]{12.0cm}{\vspace{0pt}\scriptsize The pavilion is now partially visible at the bottom-right of the image; we have flown slightly past and to the left of it. To better align with and be above the pavilion, we should shift our position laterally to the right rather than continue forward.
}
 &
move right \\

\bottomrule
\end{tabular}
\end{table*}

\begin{table*}[h]
\centering
\footnotesize
\caption{Action Process of o4-mini}

\begin{tabular}{p{0.8cm} p{1.6cm} p{12.0cm} p{1.5cm}}
\toprule
Step & Obs & Thinking & Action \\
\midrule

{\centering 0\par} &
{\centering \raisebox{-0.9\height}{\includegraphics[width=1.0cm]{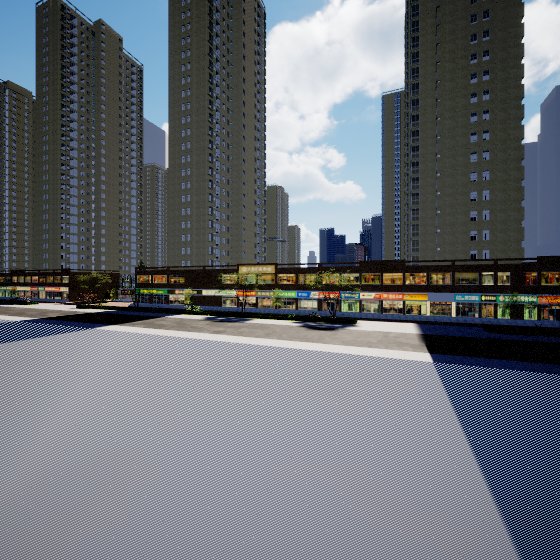}}\par}&
\parbox[t]{12.0cm}{\vspace{0pt}\scriptsize The drone is hovering over a large open plaza with a row of shops at the base of tall residential towers directly ahead. The customer's designated pavilion is likely located within the residential area beyond the shops, so it's currently out of view at the horizontal angle. Lowering the gimbal will reveal ground-level features and help identify pathways or the pavilion structure itself before advancing further.
}
 &
adjust camera gimbal downwards \\

{\centering 1\par} &
{\centering \raisebox{-0.9\height}{\includegraphics[width=1.0cm]{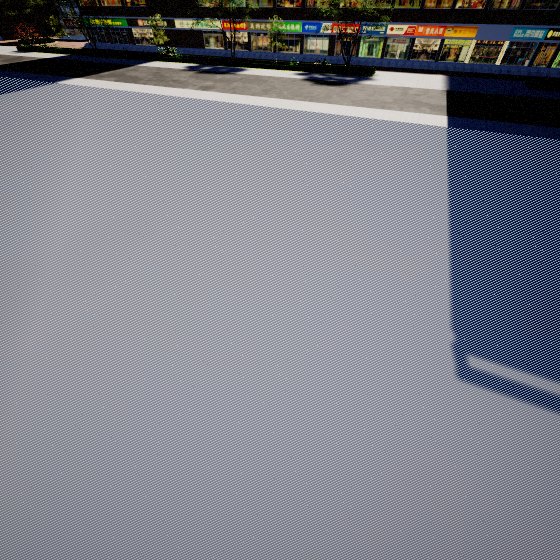}}\par}&
\parbox[t]{12.0cm}{\vspace{0pt}\scriptsize The camera is currently pitched at –45°, looking mostly at the ground and the row of shops. To identify and frame the pavilion farther ahead in the residential complex, I need to restore the camera to a horizontal view. Once leveled, I can then spot the pavilion and advance directly toward it.
}
 &
adjust camera gimbal upwards \\

{\centering 2\par} &
{\centering \raisebox{-0.9\height}{\includegraphics[width=1.0cm]{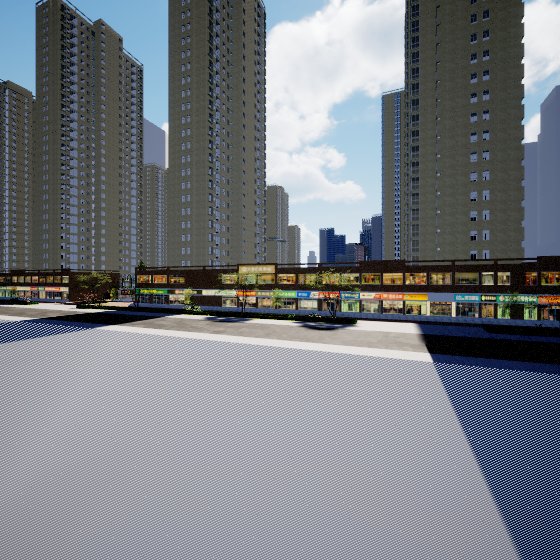}}\par} &
\parbox[t]{12.0cm}{\vspace{0pt}\scriptsize With the gimbal leveled at 0°, I can see the front row of shops at the base of the residential towers. The pavilion is further inside the complex and not yet visible, so I need to advance the drone toward the buildings to close the distance before I can spot and align with the pavilion.
}
 &
move forth \\

{\centering 3\par} &
{\centering \raisebox{-0.9\height}{\includegraphics[width=1.0cm]{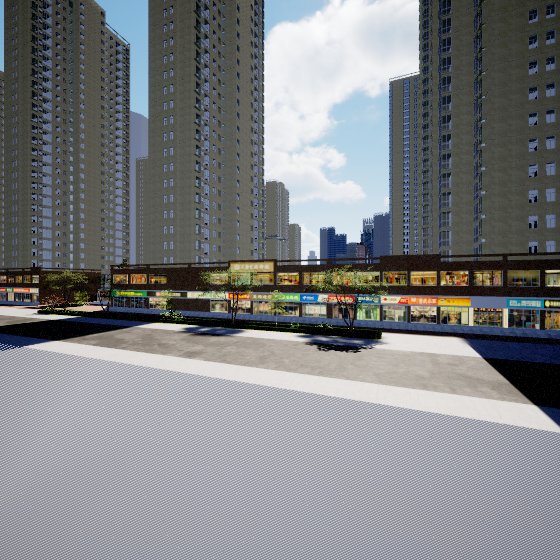}}\par} &
\parbox[t]{12.0cm}{\vspace{0pt}\scriptsize The drone is now closer to the front row of shops but the pavilion is still hidden behind. Continuing forward will bring the drone nearer to the residential complex entrance, allowing the pavilion to come into view.
}
 &
move forth \\

{\centering 4\par} &
{\centering \raisebox{-0.9\height}{\includegraphics[width=1.0cm]{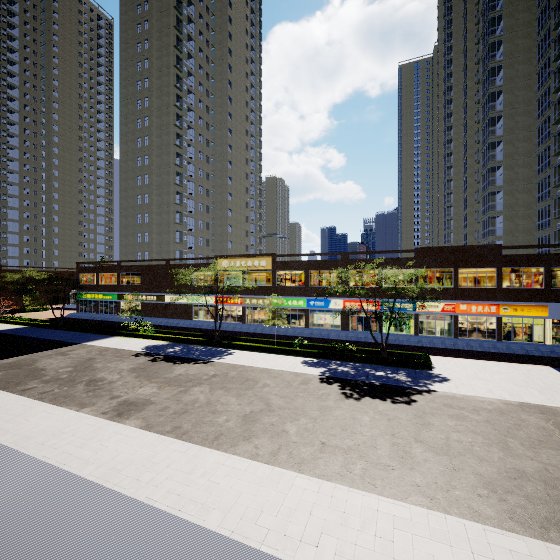}}\par}&
\parbox[t]{12.0cm}{\vspace{0pt}\scriptsize The row of shops still blocks the pavilion from view even after advancing twice. To bring the pavilion into view and reach the residential area entrance, I should continue moving forward.
}
 &
move forth \\

{\centering 5\par} &
{\centering \raisebox{-0.9\height}{\includegraphics[width=1.0cm]{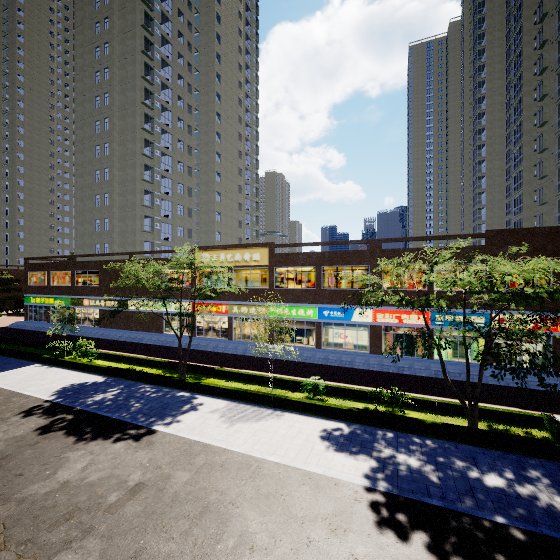}}\par} &
\parbox[t]{12.0cm}{\vspace{0pt}\scriptsize The camera is currently pitched at –45°, looking mostly at the ground and the row of shops. To identify and frame the pavilion farther ahead in the residential complex, I need to restore the camera to a horizontal view. Once leveled, I can then spot the pavilion and advance directly toward it.
}
 &
adjust camera gimbal upwards \\

{\centering 6\par} &
{\centering \raisebox{-0.9\height}{\includegraphics[width=1.0cm]{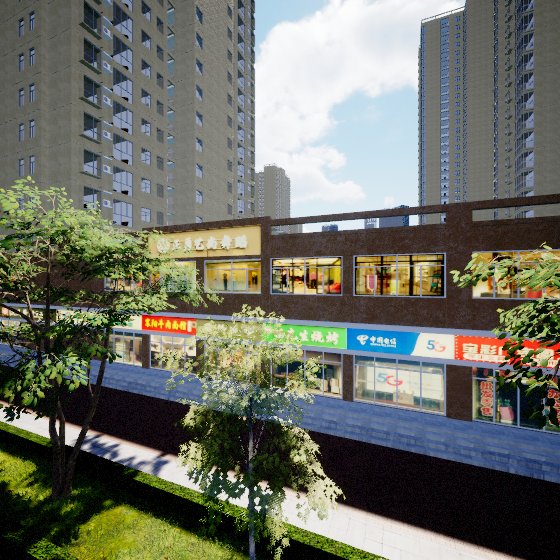}}\par} &
\parbox[t]{12.0cm}{\vspace{0pt}\scriptsize The drone is still directly in front of a row of shops, and the pavilion entrance lies between the tall towers off to the left. I need to reposition laterally to align with that gap before proceeding forward into the courtyard.
}
 &
move left \\

{\centering 7\par} &
{\centering \raisebox{-0.9\height}{\includegraphics[width=1.0cm]{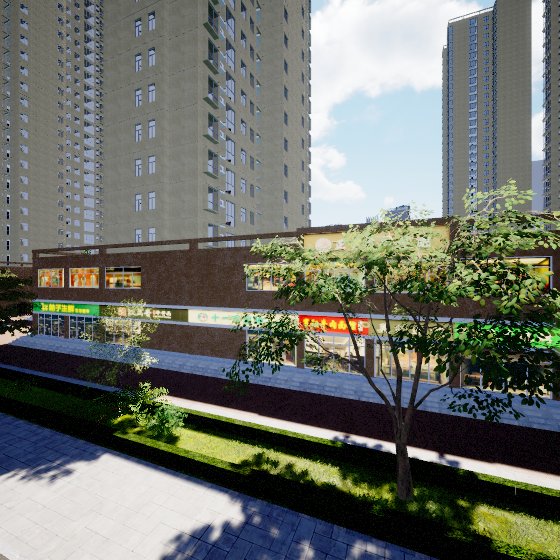}}\par} &
\parbox[t]{12.0cm}{\vspace{0pt}\scriptsize I’m now roughly aligned with the courtyard entrance but the single‐story shops and trees are still blocking my line of sight to the pavilion. By climbing 10 m, the drone will have a higher vantage point to peer over the shop roofs and locate the pavilion inside before moving forward into the complex.
}
 &
move up \\

{\centering 8\par} &
{\centering \raisebox{-0.9\height}{\includegraphics[width=1.0cm]{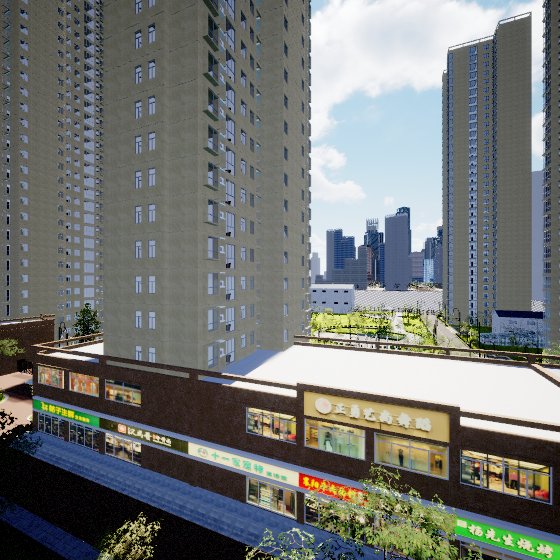}}\par} &
\parbox[t]{12.0cm}{\vspace{0pt}\scriptsize From this higher vantage I can now see the courtyard past the shops and pinpoint the small pavilion nested between the buildings. To close the distance and enter the courtyard, I should move forward toward it.
}
 &
move forth \\

{\centering 9\par} &
{\centering \raisebox{-0.9\height}{\includegraphics[width=1.0cm]{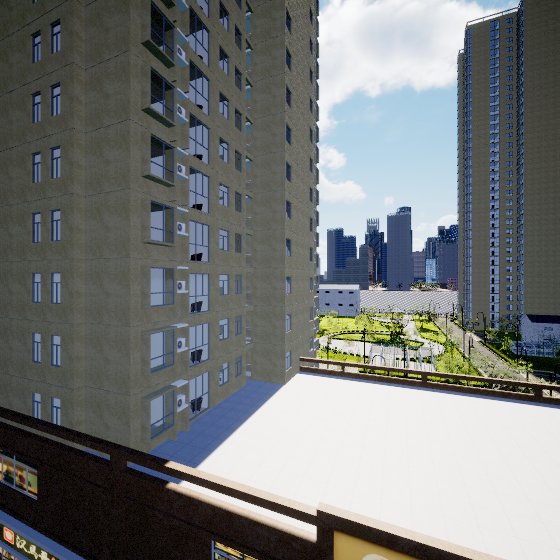}}\par} &
\parbox[t]{12.0cm}{\vspace{0pt}\scriptsize At this higher vantage I can clearly see the pavilion nestled in the courtyard beyond the shop roofs. I’m aligned with the entrance, so advancing will bring me directly into the courtyard and closer to the pavilion.
}
 &
move forth \\

{\centering 10\par} &
{\centering \raisebox{-0.9\height}{\includegraphics[width=1.0cm]{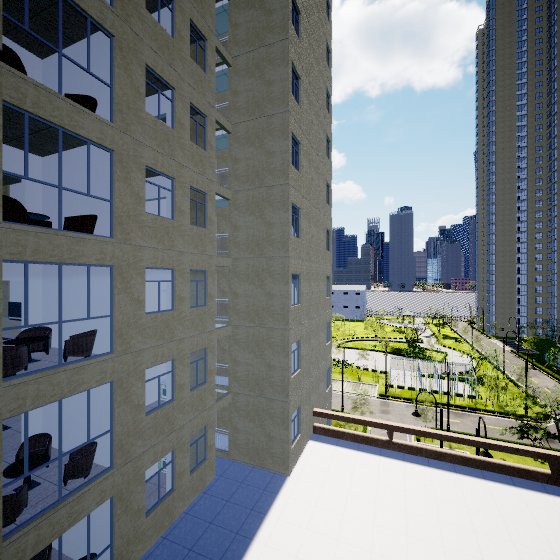}}\par} &
\parbox[t]{12.0cm}{\vspace{0pt}\scriptsize I’m perched at the edge of the shop roof and the tall building on the left is still blocking my path. The courtyard entrance and pavilion lie to my right. I need to shift laterally to the right to align over the courtyard before moving forward into the space.
}
 &
move right \\

{\centering 11\par} &
{\centering \raisebox{-0.9\height}{\includegraphics[width=1.0cm]{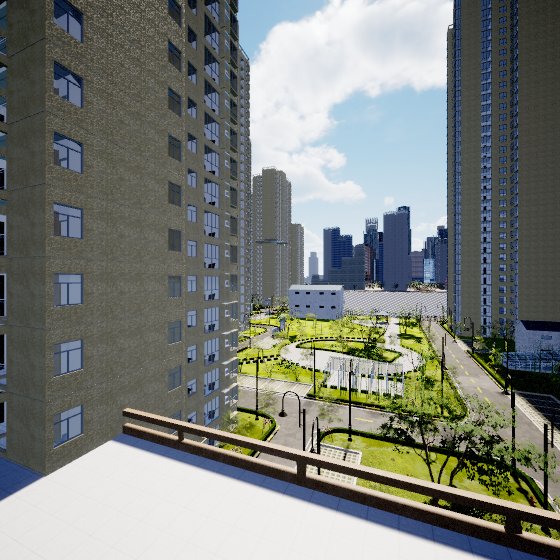}}\par} &
\parbox[t]{12.0cm}{\vspace{0pt}\scriptsize I’m now positioned above the courtyard entrance with a clear line of sight to the pavilion nestled between the buildings. To enter the courtyard and close the distance to the pavilion, I should advance directly forward.
}
 &
move forth \\

{\centering 12\par} &
{\centering \raisebox{-0.9\height}{\includegraphics[width=1.0cm]{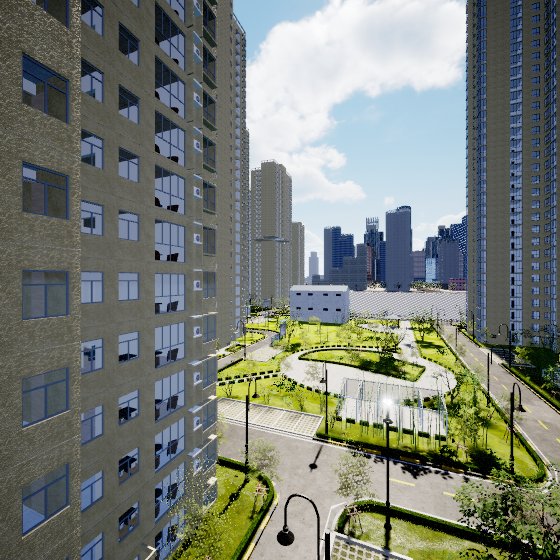}}\par}&
\parbox[t]{12.0cm}{\vspace{0pt}\scriptsize From this vantage point I have a clear view of the pavilion nestled in the courtyard ahead. I’m aligned with its entrance pathway, so advancing forward will bring the drone directly over the pavilion for a close‐up.
}
 &
move forth \\

{\centering 13\par} &
{\centering \raisebox{-0.9\height}{\includegraphics[width=1.0cm]{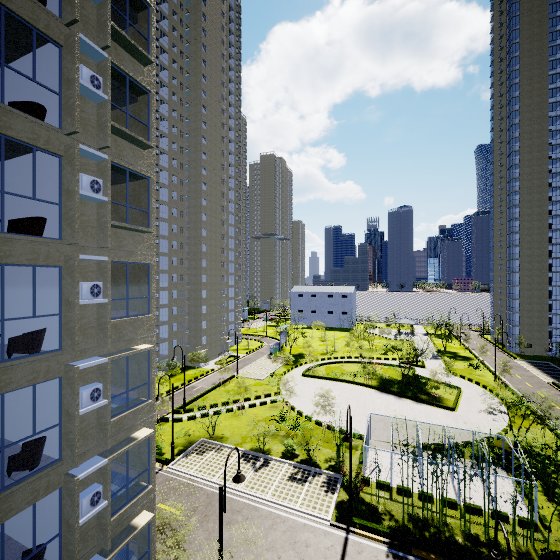}}\par} &
\parbox[t]{12.0cm}{\vspace{0pt}\scriptsize The pavilion is now centered in view and I’m aligned with its roof. Advancing will position the drone directly above it, bringing me closer for a detailed inspection.
}
 &
move forth \\

{\centering 14\par} &
{\centering \raisebox{-0.9\height}{\includegraphics[width=1.0cm]{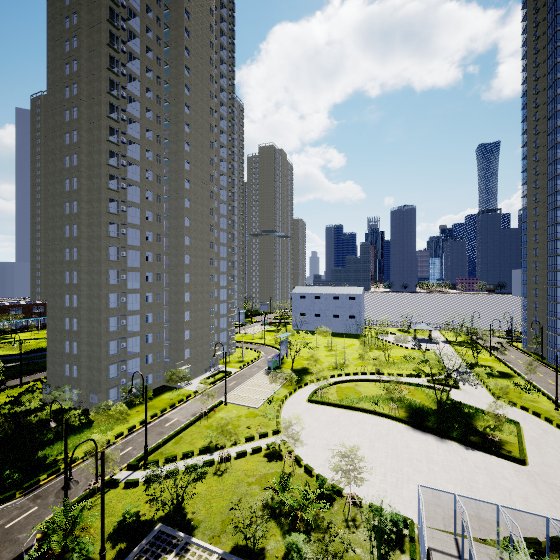}}\par} &
\parbox[t]{12.0cm}{\vspace{0pt}\scriptsize The pavilion is now clearly visible ahead in the courtyard and I’m roughly aligned with it. To position the drone directly above the pavilion for a close inspection, I should advance forward into the courtyard.
}
 &
move forth \\

{\centering 15\par} &
{\centering \raisebox{-0.9\height}{\includegraphics[width=1.0cm]{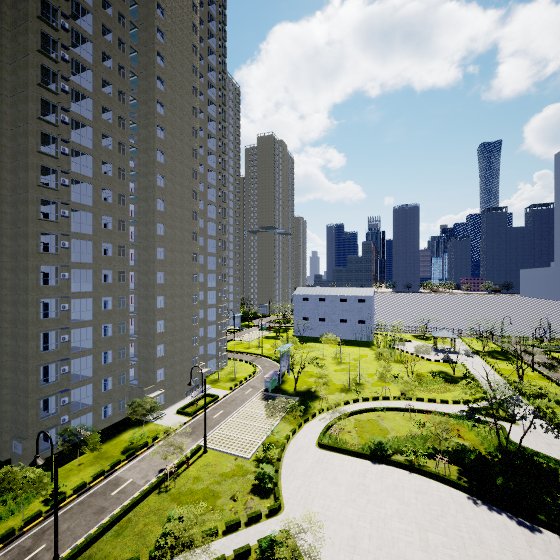}}\par}&
\parbox[t]{12.0cm}{\vspace{0pt}\scriptsize The pavilion is now clearly visible ahead within the courtyard. Continuing to advance will close the remaining distance and place the drone directly above the pavilion for a closer view.
}
 &
move forth \\

{\centering 16\par} &
{\centering \raisebox{-0.9\height}{\includegraphics[width=1.0cm]{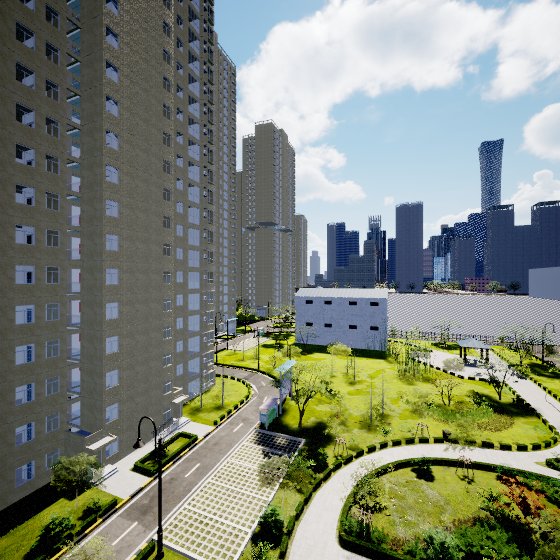}}\par} &
\parbox[t]{12.0cm}{\vspace{0pt}\scriptsize The small pavilion with the white roof is directly ahead in the courtyard and now well centered. Advancing one more time will bring the drone right above the pavilion for a close approach.

}
 &
move forth \\

\bottomrule
\end{tabular}
\end{table*}

\begin{table*}[h]
\centering
\footnotesize
\caption{Action Process of o4-mini}

\begin{tabular}{p{0.8cm} p{1.6cm} p{12.0cm} p{1.5cm}}
\toprule
Step & Obs & Thinking & Action \\
\midrule

{\centering 17\par} &
{\centering \raisebox{-0.9\height}{\includegraphics[width=1.0cm]{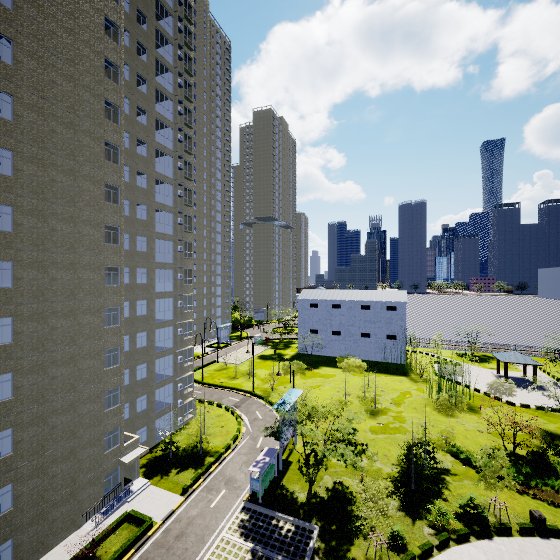}}\par} &
\parbox[t]{12.0cm}{\vspace{0pt}\scriptsize I’m now just left of the pavilion hidden behind the white box structure. Shifting to the right will bring the drone directly over the pavilion so I can center it before closing in further.
}
 &
move right \\

{\centering 18\par} &
{\centering \raisebox{-0.9\height}{\includegraphics[width=1.0cm]{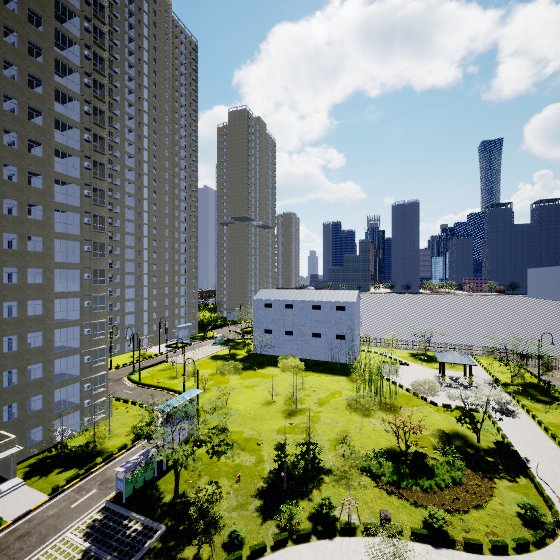}}\par} &
\parbox[t]{12.0cm}{\vspace{0pt}\scriptsize The pavilion is now visible to the right of the white utility building and I’m aligned to enter the courtyard. Advancing forward will close the distance and position the drone directly over the pavilion.
}
 &
move forth \\

{\centering 19\par} &
{\centering \raisebox{-0.9\height}{\includegraphics[width=1.0cm]{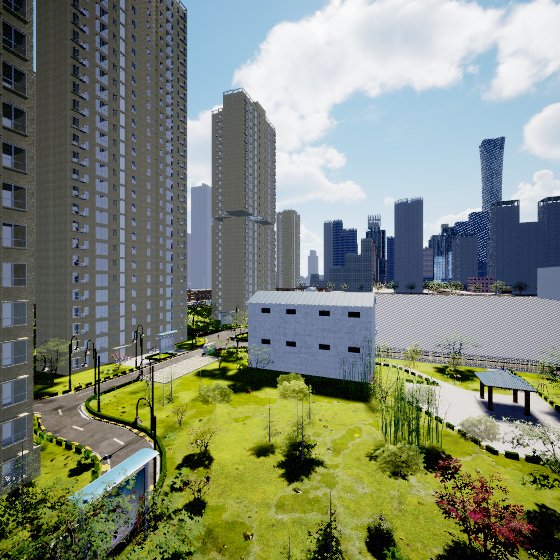}}\par} &
\parbox[t]{12.0cm}{\vspace{0pt}\scriptsize The pavilion is still partially obscured by the white utility building in the center. Shifting right will bring the drone directly over the pavilion area and center it in frame before advancing further.
}
 &
move right \\

{\centering 20\par} &
{\centering \raisebox{-0.9\height}{\includegraphics[width=1.0cm]{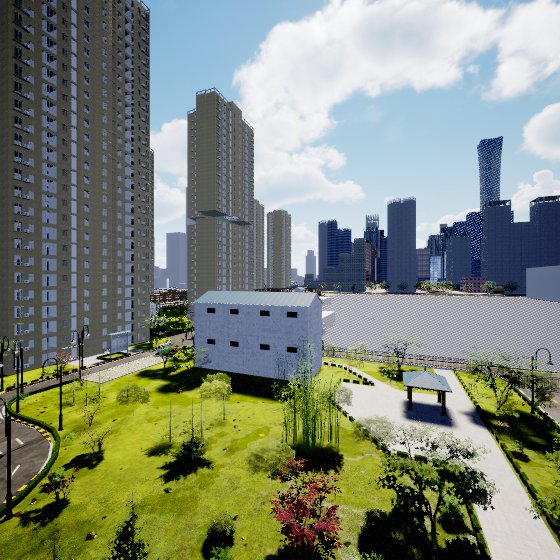}}\par} &
\parbox[t]{12.0cm}{\vspace{0pt}\scriptsize The small pavilion is still partially hidden behind the white utility building. Shifting the drone to the right will center the pavilion in view and clear the obstruction before the final forward approach.
}
 &
move right \\

{\centering 21\par} &
{\centering \raisebox{-0.9\height}{\includegraphics[width=1.0cm]{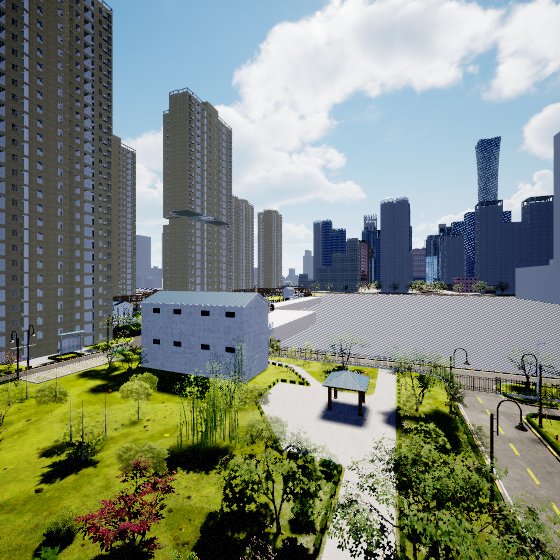}}\par}&
\parbox[t]{12.0cm}{\vspace{0pt}\scriptsize The pavilion is now fully visible to the right of the utility building and well within my view. Advancing forward will close the remaining distance and position the drone directly above the pavilion for a close inspection.
}
 &
move forth \\

{\centering 22\par} &
{\centering \raisebox{-0.9\height}{\includegraphics[width=1.0cm]{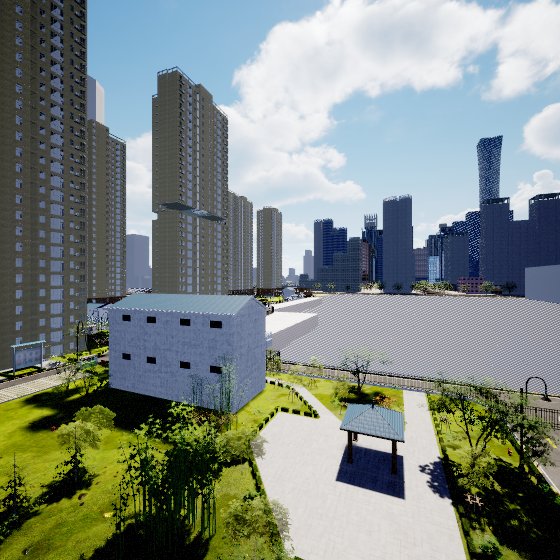}}\par} &
\parbox[t]{12.0cm}{\vspace{0pt}\scriptsize The pavilion is now directly ahead and clearly visible at ground level. Advancing forward one more time will place the drone directly above the pavilion for a close-up view.
}
 &
move forth \\

{\centering 23\par} &
{\centering \raisebox{-0.9\height}{\includegraphics[width=1.0cm]{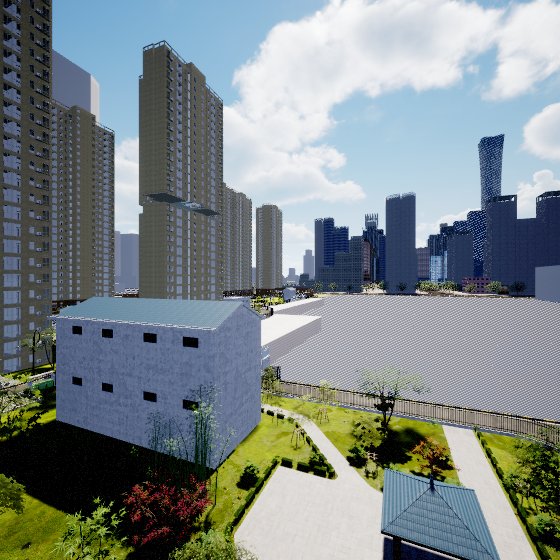}}\par} &
\parbox[t]{12.0cm}{\vspace{0pt}\scriptsize The pavilion is now just to the lower right of my view, still slightly offset behind the white building. Shifting the drone to the right will center the pavilion in frame before the final forward approach.
}
 &
move right \\

{\centering 24\par} &
{\centering \raisebox{-0.9\height}{\includegraphics[width=1.0cm]{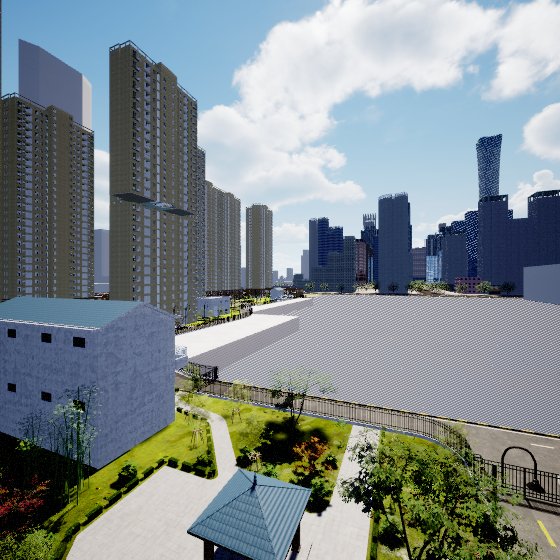}}\par} &
\parbox[t]{12.0cm}{\vspace{0pt}\scriptsize The pavilion is partially visible at the bottom center but my drone is too far to the right. I need to shift left to center the pavilion in view before making the final forward approach.
}
 &
move left \\

{\centering 25\par} &
{\centering \raisebox{-0.9\height}{\includegraphics[width=1.0cm]{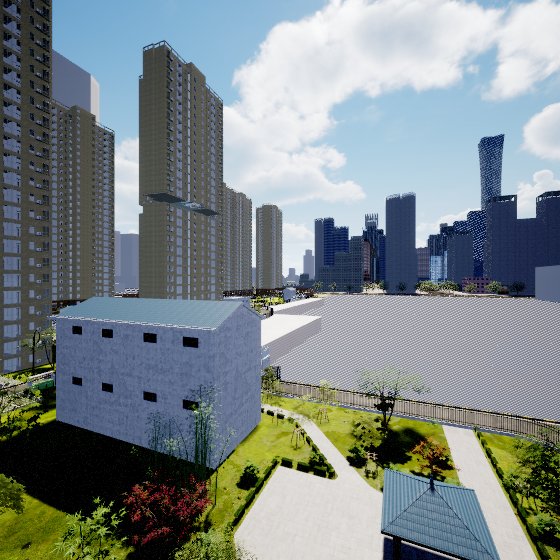}}\par}&
\parbox[t]{12.0cm}{\vspace{0pt}\scriptsize The pavilion is now centered in view at the bottom of the frame and unobstructed by the utility building. To close the remaining gap and hover directly above it for inspection, I should advance forward.
}
 &
move forth \\

{\centering 26\par} &
{\centering \raisebox{-0.9\height}{\includegraphics[width=1.0cm]{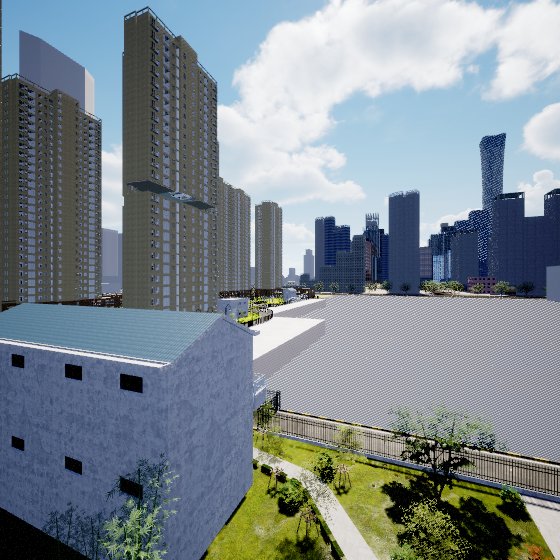}}\par}&
\parbox[t]{12.0cm}{\vspace{0pt}\scriptsize I’m much closer now but the pavilion remains off to the lower right and is still partly obscured by the utility building. Shifting the drone to the right will center the pavilion in frame before the final forward approach.
}
 &
move right \\

\bottomrule
\end{tabular}
\end{table*}

\begin{table*}[h]
\centering
\footnotesize
\caption{Action Process of Gemini-2.5-Flash}

\begin{tabular}{p{0.8cm} p{1.6cm} p{12.0cm} p{1.5cm}}
\toprule
Step & Obs & Thinking & Action \\
\midrule

{\centering 0\par} &
{\centering \raisebox{-0.9\height}{\includegraphics[width=1.0cm]{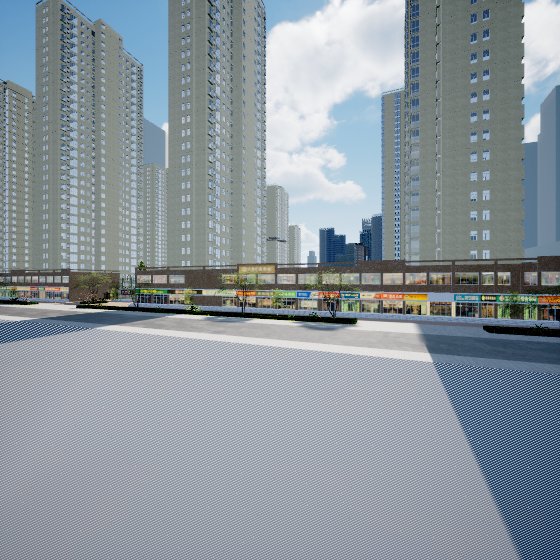}}\par} &
\parbox[t]{12.0cm}{\vspace{0pt}\scriptsize I am currently positioned in front of tall residential buildings with a commercial strip at their base and a wide empty paved area. The camera gimbal is horizontal (0.0\textdegree). I cannot see any pavilion yet. The most logical first step is to move forward to get closer to the residential area and potentially reveal the pavilion.
}
 &
move forth \\

{\centering 1\par} &
{\centering \raisebox{-0.9\height}{\includegraphics[width=1.0cm]{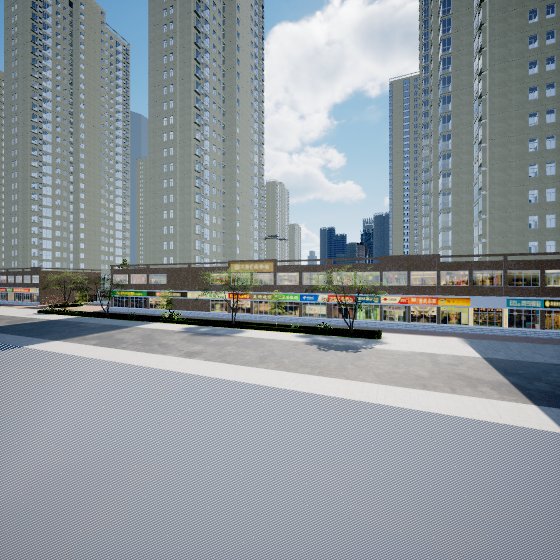}}\par} &
\parbox[t]{12.0cm}{\vspace{0pt}\scriptsize The current image is identical to the previous one. The goal is still ``a pavilion inside the residential area ahead'' and I still cannot see it. To find the pavilion, I need to move into the residential area. The most direct path is forward.
}
 &
move forth \\

{\centering 2\par} &
{\centering \raisebox{-0.9\height}{\includegraphics[width=1.0cm]{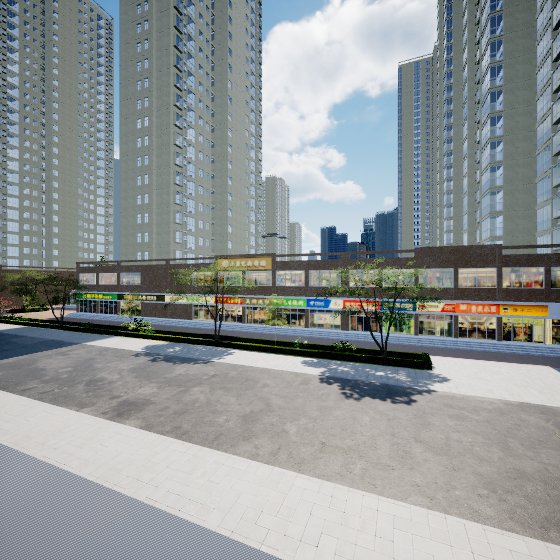}}\par} &
\parbox[t]{12.0cm}{\vspace{0pt}\scriptsize I have moved forward since the last image. The commercial strip now appears closer. However, I still cannot see the pavilion, which is expected to be ``inside the residential area ahead.'' To reach the residential area, I need to continue moving past the commercial establishments.
}
 &
move forth \\

{\centering 3\par} &
{\centering \raisebox{-0.9\height}{\includegraphics[width=1.0cm]{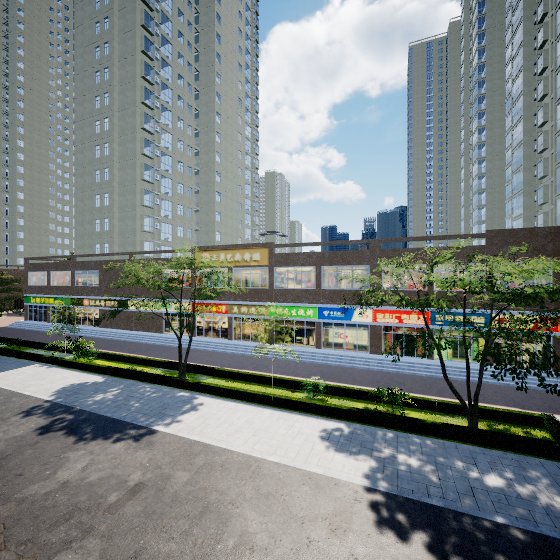}}\par} &
\parbox[t]{12.0cm}{\vspace{0pt}\scriptsize The drone has moved forward again. The commercial strip is now even closer. I still cannot see any pavilion. It's likely located past this commercial strip, within the residential complex. Moving forward is still the best option to continue approaching the residential area.
}
 &
move forth \\

{\centering 4\par} &
{\centering \raisebox{-0.9\height}{\includegraphics[width=1.0cm]{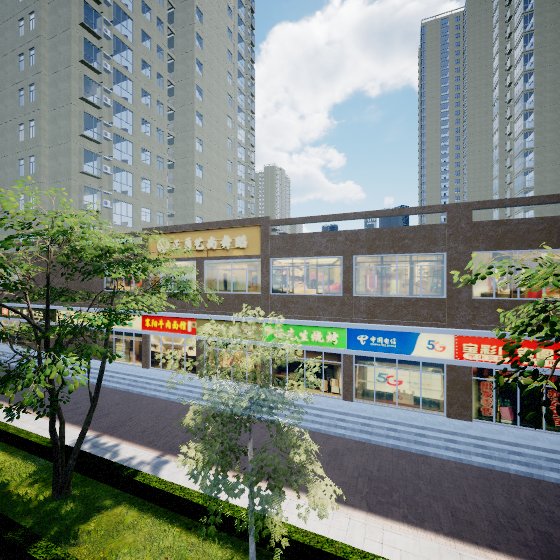}}\par} &
\parbox[t]{12.0cm}{\vspace{0pt}\scriptsize The drone has moved closer to the commercial strip. The street and the grassy median are now prominent. I am still at the external boundary of the residential area. Continuing to move forward is the most direct way to get past the shops and into the residential complex.
}
 &
move forth \\

{\centering 5\par} &
{\centering \raisebox{-0.9\height}{\includegraphics[width=1.0cm]{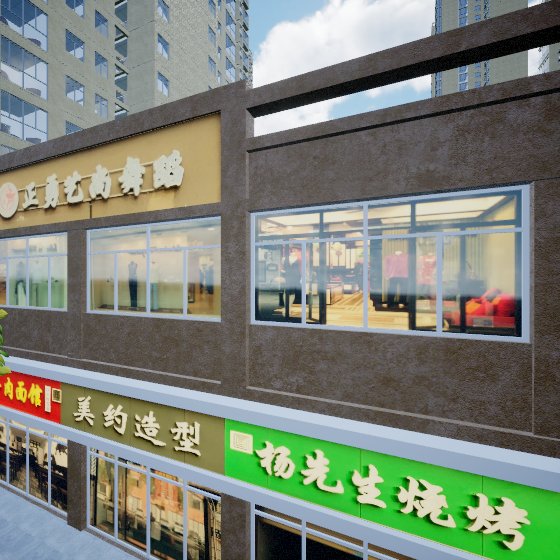}}\par} &
\parbox[t]{12.0cm}{\vspace{0pt}\scriptsize I have repeatedly moved forward and am now very close to the commercial strip. The previous strategy of simply moving forward is no longer viable as it would lead to collision. Gaining elevation (move\_up) appears to be the most promising action to survey the area and locate the pavilion from a higher vantage point.
}
 &
move up \\

{\centering 6\par} &
{\centering \raisebox{-0.9\height}{\includegraphics[width=1.0cm]{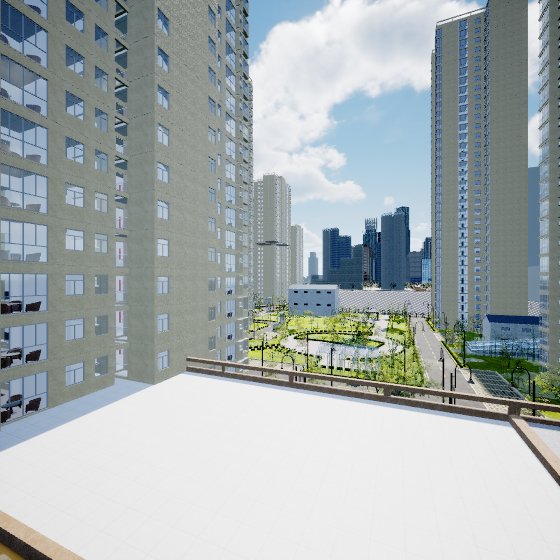}}\par} &
\parbox[t]{12.0cm}{\vspace{0pt}\scriptsize I have successfully moved up and now have an elevated view over the commercial strip, looking into a landscaped residential area. This area contains paths, greenery, and a small blue-roofed building on the right side---a strong candidate for the pavilion. I should adjust the camera angle downwards for a better look.
}
 &
angle down \\

{\centering 7\par} &
{\centering \raisebox{-0.9\height}{\includegraphics[width=1.0cm]{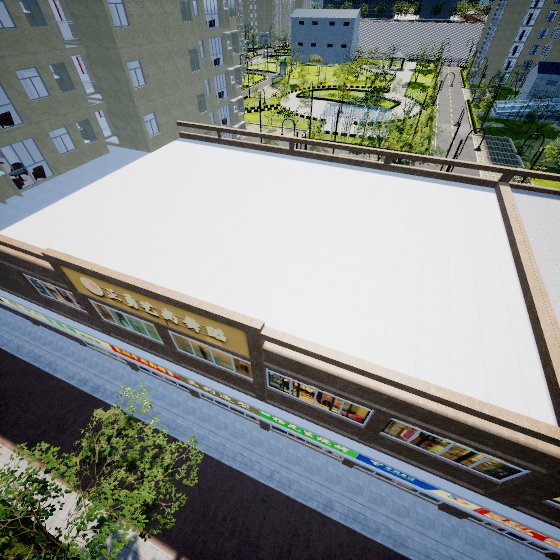}}\par} &
\parbox[t]{12.0cm}{\vspace{0pt}\scriptsize I have moved up and angled the camera down to -45\textdegree, which gives me a good overview of the residential area. I can clearly see a park-like area with a small blue-roofed building towards the right side. To reach the pavilion, I need to move horizontally into the residential area.
}
 &
move forth \\

{\centering 8\par} &
{\centering \raisebox{-0.9\height}{\includegraphics[width=1.0cm]{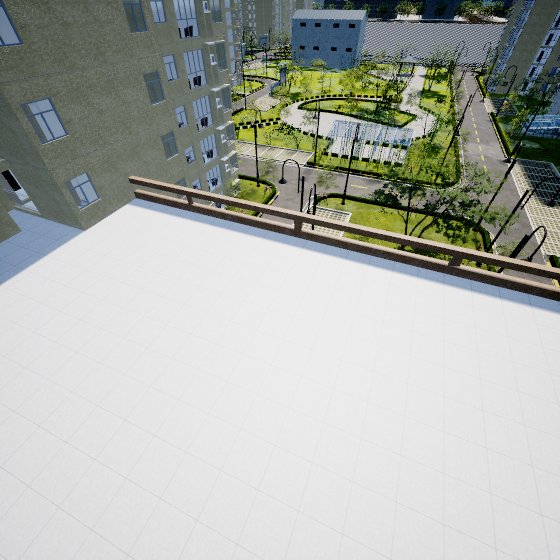}}\par} &
\parbox[t]{12.0cm}{\vspace{0pt}\scriptsize I am currently positioned above the commercial strip, looking into the residential area. I can clearly see a park-like area with the blue-roofed pavilion. To reach the pavilion, I need to continue moving forward into this area.
}
 &
move forth \\

{\centering 9\par} &
{\centering \raisebox{-0.9\height}{\includegraphics[width=1.0cm]{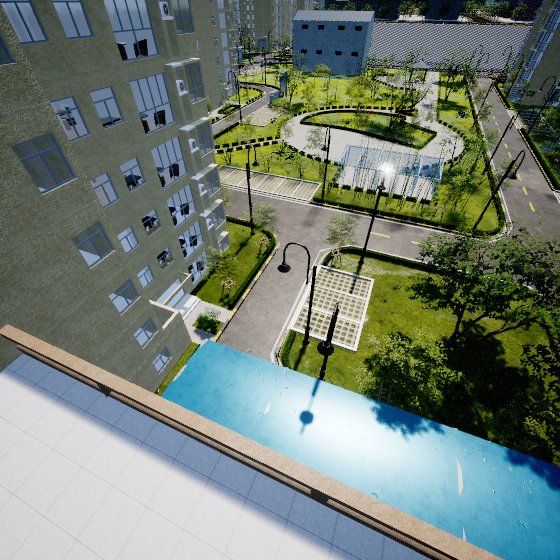}}\par} &
\parbox[t]{12.0cm}{\vspace{0pt}\scriptsize The pavilion is visible but is located towards the right side of my current field of view. To approach it more directly, I should turn the drone to align my forward direction with the pavilion. Turning right by 22.5\textdegree{} will center the pavilion in my view.
}
 &
turn right \\

{\centering 10\par} &
{\centering \raisebox{-0.9\height}{\includegraphics[width=1.0cm]{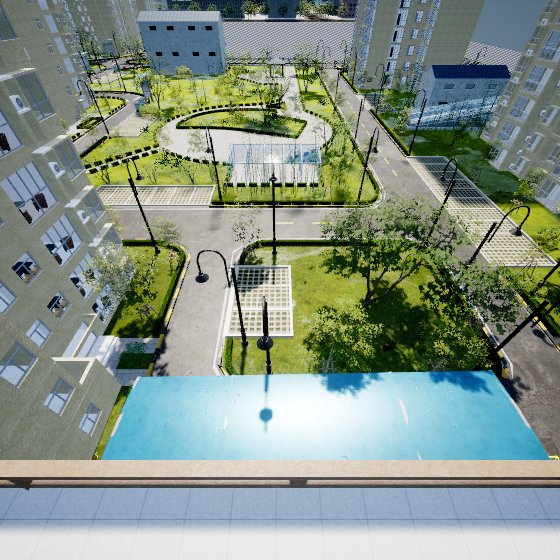}}\par} &
\parbox[t]{12.0cm}{\vspace{0pt}\scriptsize I have turned right, and the potential pavilion is now more centered in my view. My camera gimbal is still at -45.0\textdegree. To continue approaching the pavilion, I need to cover the horizontal distance. Moving forward will take me directly towards it.
}
 &
move forth \\

{\centering 11\par} &
{\centering \raisebox{-0.9\height}{\includegraphics[width=1.0cm]{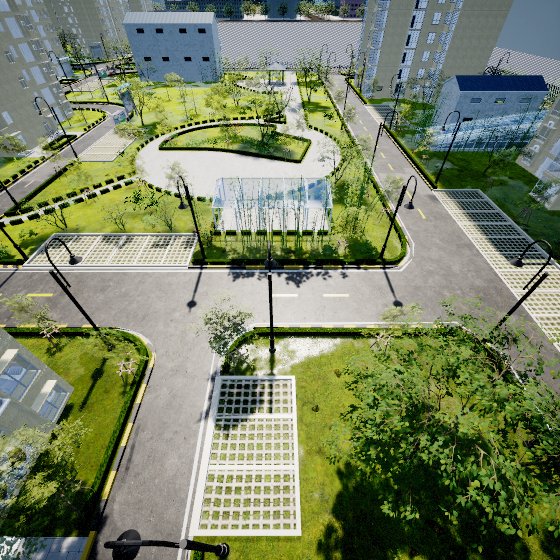}}\par} &
\parbox[t]{12.0cm}{\vspace{0pt}\scriptsize I have successfully moved further into the residential area. The pavilion is now closer and more clearly visible. Moving forward by another 10 meters will bring me significantly closer to the target.
}
 &
move forth \\

{\centering 12\par} &
{\centering \raisebox{-0.9\height}{\includegraphics[width=1.0cm]{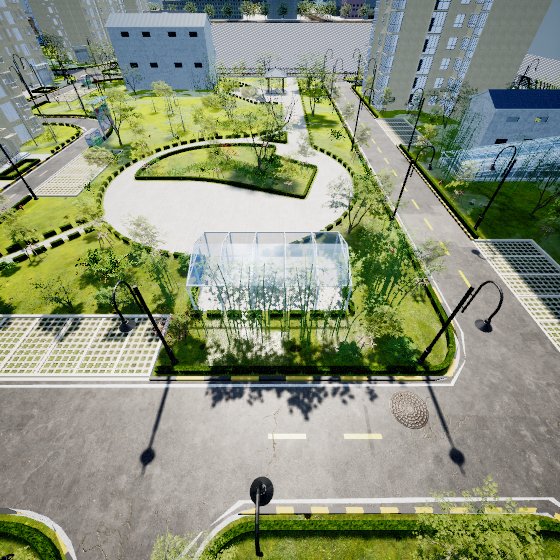}}\par} &
\parbox[t]{12.0cm}{\vspace{0pt}\scriptsize The glass structure I am identifying as the pavilion is now very close and clearly visible. Moving forward one more time will place the drone directly over or very near the pavilion, positioning it perfectly for a subsequent descent.
}
 &
move forth \\

{\centering 13\par} &
{\centering \raisebox{-0.9\height}{\includegraphics[width=1.0cm]{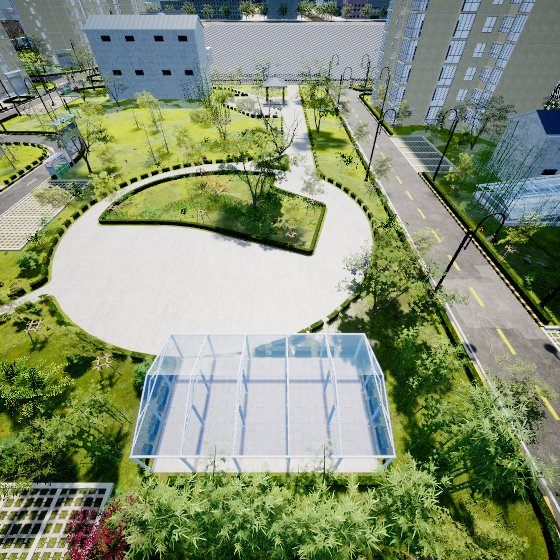}}\par} &
\parbox[t]{12.0cm}{\vspace{0pt}\scriptsize The pavilion is now directly below or very close to being directly below my current position. To reach the pavilion, I now need to lower the drone's altitude.
}
 &
move down \\

{\centering 14\par} &
{\centering \raisebox{-0.9\height}{\includegraphics[width=1.0cm]{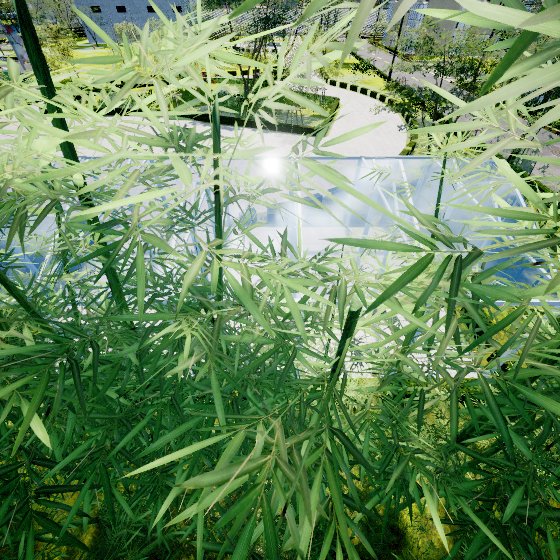}}\par} &
\parbox[t]{12.0cm}{\vspace{0pt}\scriptsize I have descended by 10 meters, and my view is now heavily obstructed by dense green foliage. I am too low and entangled in the vegetation. Moving up by 10 meters will lift the drone above this foliage, allowing me to reassess the pavilion's exact location.
}
 &
move up \\

{\centering 15\par} &
{\centering \raisebox{-0.9\height}{\includegraphics[width=1.0cm]{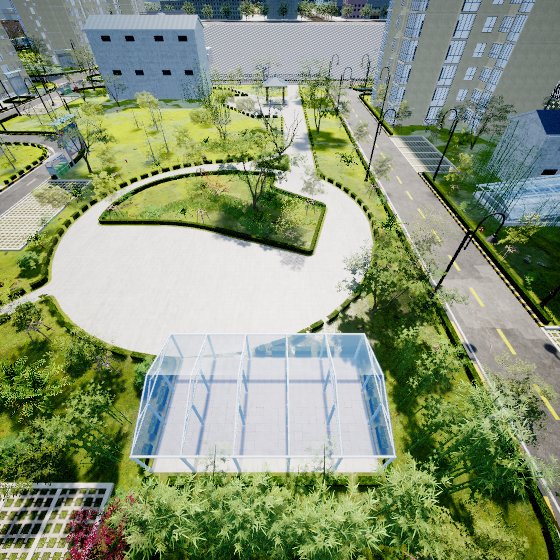}}\par} &
\parbox[t]{12.0cm}{\vspace{0pt}\scriptsize I have moved the drone up after getting too close to the foliage, and now have a clear view of the glass pavilion structure directly below. Since the pavilion is clearly visible and directly beneath the drone, the most logical step is to lower the drone's altitude.
}
 &
move down \\

{\centering 16\par} &
{\centering \raisebox{-0.9\height}{\includegraphics[width=1.0cm]{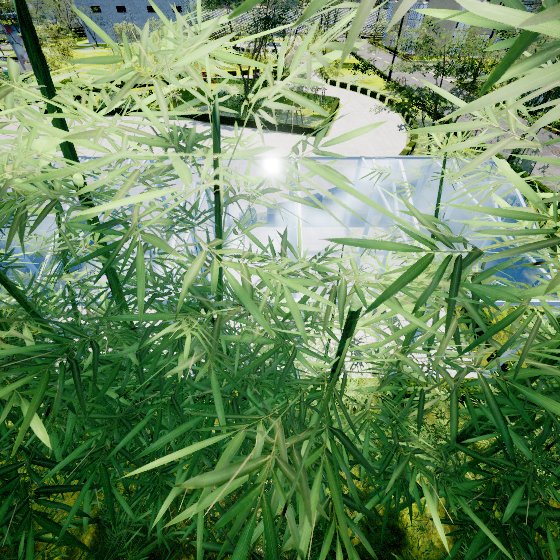}}\par} &
\parbox[t]{12.0cm}{\vspace{0pt}\scriptsize I have descended again and my view is once more obstructed by dense green foliage. I need to clear this obstruction. Moving up will lift the drone above the foliage.
}
 &
move up \\

\bottomrule
\end{tabular}
\end{table*}

\begin{table*}[h]
\centering
\footnotesize
\caption{Action Process of Gemini-2.5-Flash}

\begin{tabular}{p{0.8cm} p{1.6cm} p{12.0cm} p{1.5cm}}
\toprule
Step & Obs & Thinking & Action \\
\midrule

{\centering 17\par} &
{\centering \raisebox{-0.9\height}{\includegraphics[width=1.0cm]{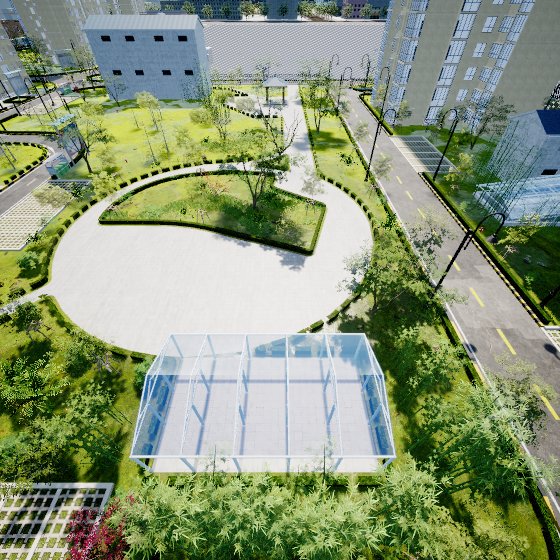}}\par} &
\parbox[t]{12.0cm}{\vspace{0pt}\scriptsize I have successfully moved up, bringing the drone above the obstructing foliage, and now have a clear overhead view of the glass pavilion. Since I am currently directly above it, the next logical step is to descend to its level.
}
 &
move down \\

{\centering 18\par} &
{\centering \raisebox{-0.9\height}{\includegraphics[width=1.0cm]{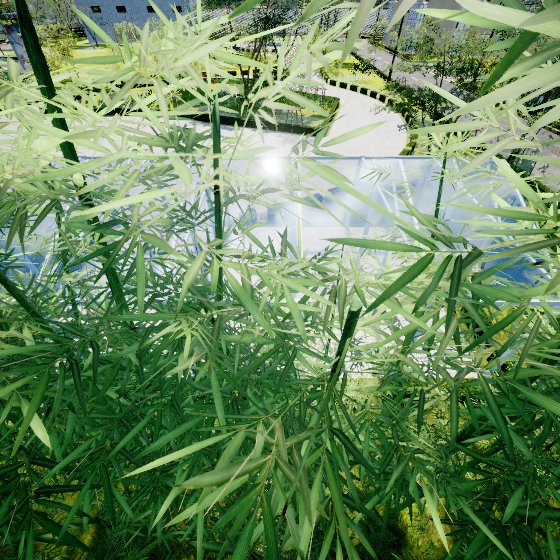}}\par} &
\parbox[t]{12.0cm}{\vspace{0pt}\scriptsize I have descended by 10 meters and my view is now once again obstructed by dense foliage. To get a clear view of the pavilion and its surroundings again, I need to move the drone upwards above the foliage.
}
 &
move up \\

{\centering 19\par} &
{\centering \raisebox{-0.9\height}{\includegraphics[width=1.0cm]{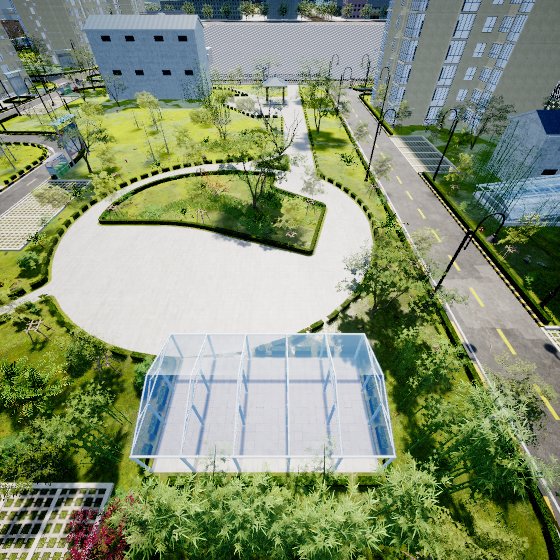}}\par} &
\parbox[t]{12.0cm}{\vspace{0pt}\scriptsize I have successfully moved the drone up again, placing it above the obstructing foliage, and now have a clear overhead view of the glass pavilion. Since the drone is now directly above the target, the most direct action to reach the pavilion is to descend.
}
 &
move down \\

\bottomrule
\end{tabular}
\end{table*}

\begin{table*}[h]
\centering
\footnotesize
\caption{Action Process of Qwen2.5-VL-7B}

\begin{tabular}{p{0.8cm} p{1.6cm} p{12.0cm} p{1.5cm}}
\toprule
Step & Obs & Thinking & Action \\
\midrule

{\centering 0\par} &
{\centering \raisebox{-0.9\height}{\includegraphics[width=1.0cm]{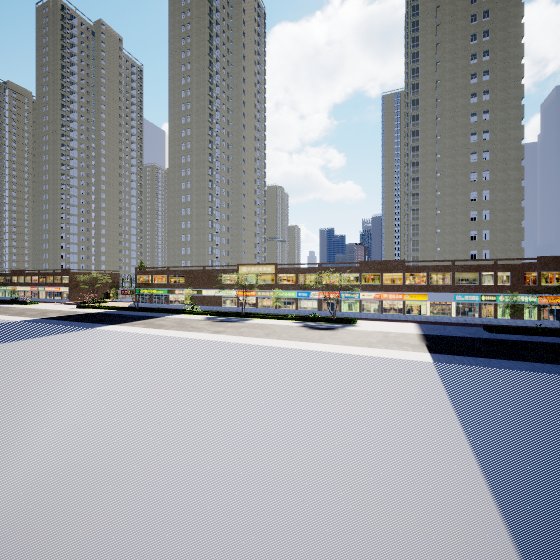}}\par} &
\parbox[t]{12.0cm}{\vspace{0pt}\scriptsize The current scene shows an urban setting with tall buildings on either side. To navigate towards the pavilion located within this residential area, we need to adjust our heading slightly away from these structures.
}
 &
adjust camera gimbal downwards \\

{\centering 1\par} &
{\centering \raisebox{-0.9\height}{\includegraphics[width=1.0cm]{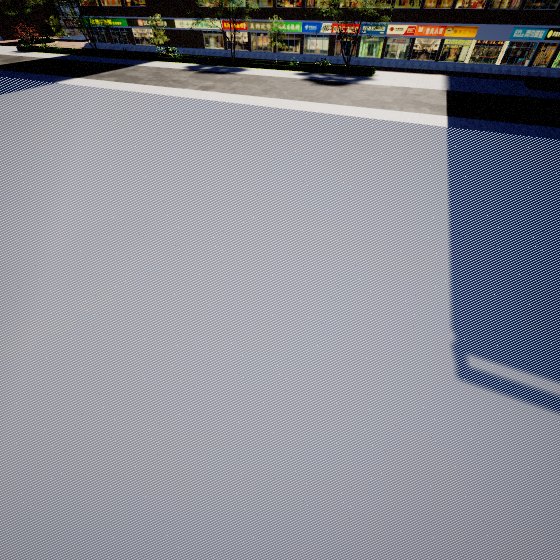}}\par} &
\parbox[t]{12.0cm}{\vspace{0pt}\scriptsize Given the current camera angle at `-45` degrees, which tilts downwards significantly, my next step would involve adjusting both the camera orientation upwards while also preparing to move forward toward potential pathways leading into the residential area. This upward movement helps clear obstructed views before proceeding further.
}
 &
adjust camera gimbal upwards \\

{\centering 2\par} &
{\centering \raisebox{-0.9\height}{\includegraphics[width=1.0cm]{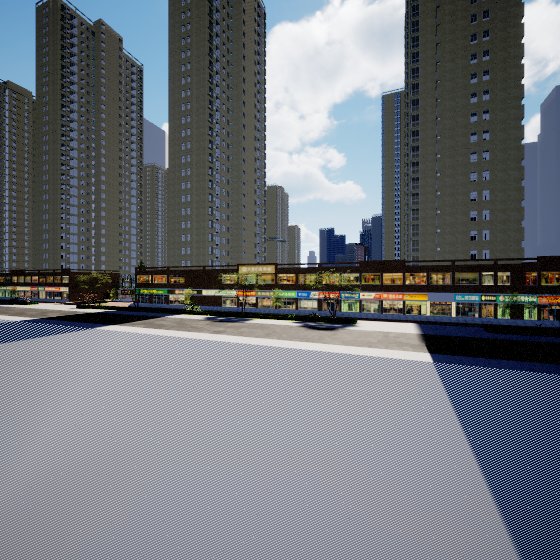}}\par} &
\parbox[t]{12.0cm}{\vspace{0pt}\scriptsize Currently, there seems no immediate obstacle blocking direct progress; however, given the layout visible here—wide open spaces flanked by high-rise buildings—it might make sense initially to move straight ahead if possible without turning right immediately due to potentially narrow paths or tight corners nearby not clearly seen yet.
}
 &
move forth \\

{\centering 3\par} &
{\centering \raisebox{-0.9\height}{\includegraphics[width=1.0cm]{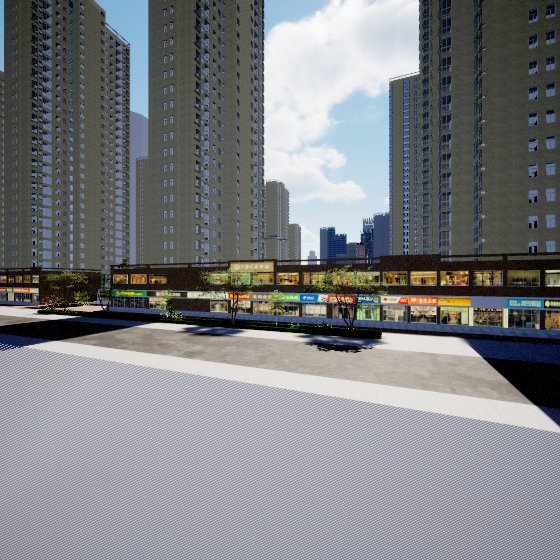}}\par} &
\parbox[t]{12.0cm}{\vspace{0pt}\scriptsize From the image description, it appears like there’s some kind of pathway or road extending directly ahead between two rows of buildings. Given the goal is to reach a pavilion likely situated somewhere along this path, continuing straight could lead us closer to achieving that objective efficiently since other directions seem less promising based solely on visual cues available now.
}
 &
move forth \\

{\centering 4\par} &
{\centering \raisebox{-0.9\height}{\includegraphics[width=1.0cm]{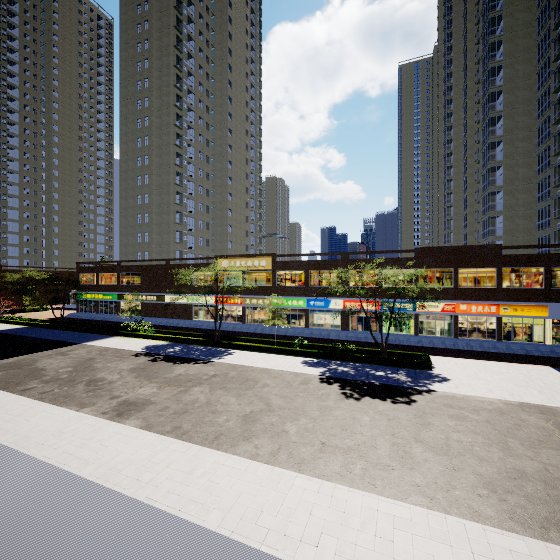}}\par}&
\parbox[t]{12.0cm}{\vspace{0pt}\scriptsize Based on the current perspective shown above, it looks like we're positioned near but still outside a row of shops or commercial establishments. There doesn’t appear to be another obvious route branching off just beyond them unless hidden behind shadows or obscured elements out-of-frame. Since the goal involves reaching something specifically labeled "pavilion," let me consider whether advancing straight may reveal more details about its exact position relative to those buildings.
}
 &
move forth \\

{\centering 5\par} &
{\centering \raisebox{-0.9\height}{\includegraphics[width=1.0cm]{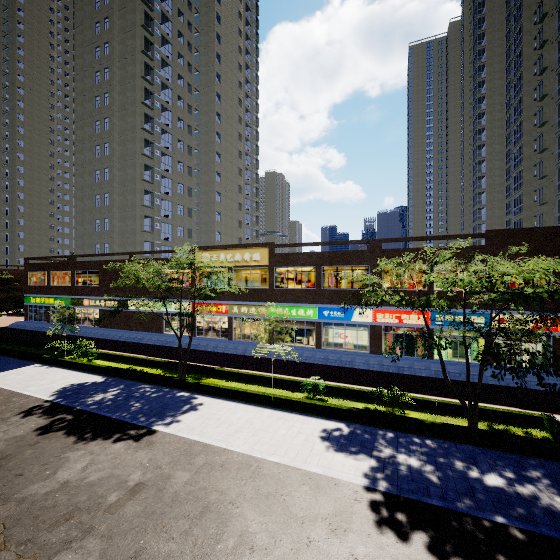}}\par} &
\parbox[t]{12.0cm}{\vspace{0pt}\scriptsize Looking closely at the image again, despite being close enough visually to see part of the building structure resembling a pavilion-like design among others, it isn't entirely certain how far back they extend nor their precise positioning compared against surrounding areas. However, considering proximity suggests continued advancement might help clarify surroundings further.
}
 &
move forth \\

{\centering 6\par} &
{\centering \raisebox{-0.9\height}{\includegraphics[width=1.0cm]{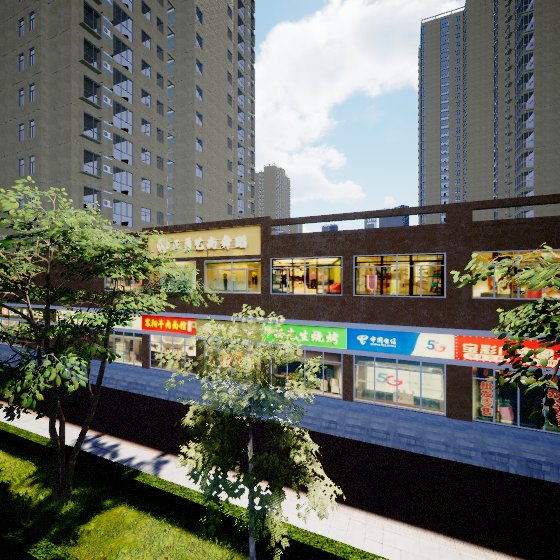}}\par} &
\parbox[t]{12.0cm}{\vspace{0pt}\scriptsize Upon reviewing the latest snapshot, although the presence of trees partially obscures parts of the view, it does provide clearer sightlines than previous images did. It indicates that we’re approaching a specific point where additional features become distinguishable – possibly hinting at the entrance or vicinity of the desired pavilion amidst these storefronts.
}
 &
move forth \\

{\centering 7\par} &
{\centering \raisebox{-0.9\height}{\includegraphics[width=1.0cm]{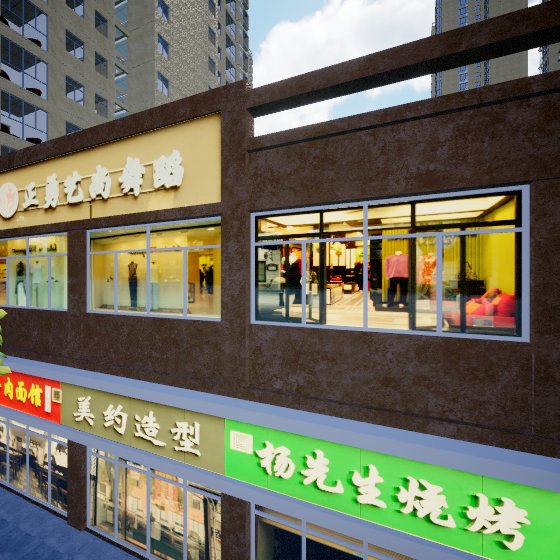}}\par} &
\parbox[t]{12.0cm}{\vspace{0pt}\scriptsize In the most recent frame, though we have moved closer to the stores mentioned earlier, none show signs pointing definitively towards a pavilion. Instead, they suggest typical retail environments rather than recreational or public gathering spots typically associated with pavilions. Therefore, it makes logical sense to re-evaluate our course correction needed prior to making significant directional changes.
}
 &
turn left \\

{\centering 8\par} &
{\centering \raisebox{-0.9\height}{\includegraphics[width=1.0cm]{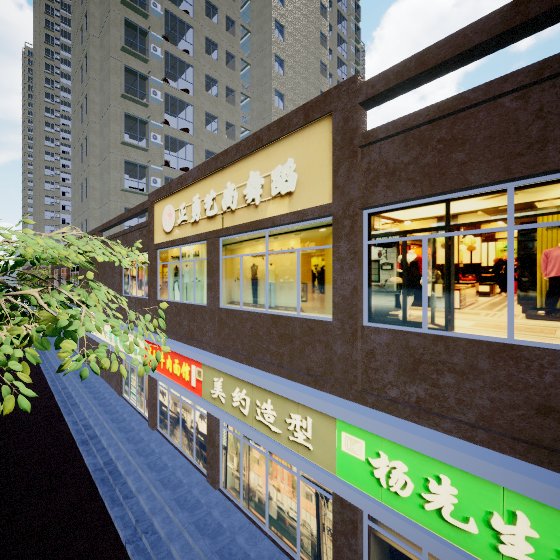}}\par}&
\parbox[t]{12.0cm}{\vspace{0pt}\scriptsize Despite exploring various angles and directions previously, the current view continues showing primarily commercial premises instead of a pavilion. Given limited information regarding the exact nature or location of the pavilion, it remains plausible that we've passed over key landmarks already explored thus far.
}
 &
adjust camera gimbal upwards \\

{\centering 9\par} &
{\centering \raisebox{-0.9\height}{\includegraphics[width=1.0cm]{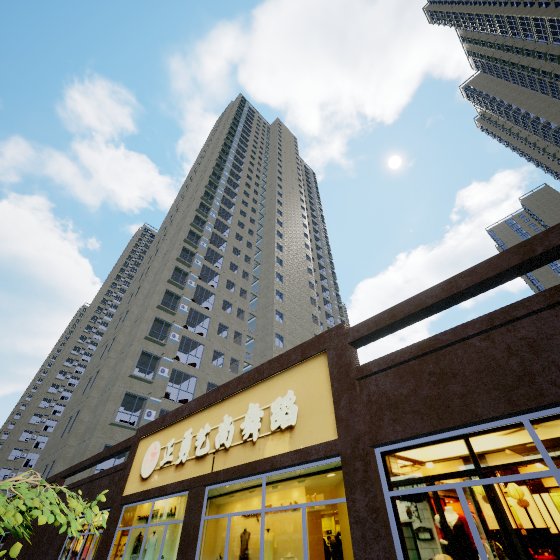}}\par}&
\parbox[t]{12.0cm}{\vspace{0pt}\scriptsize With the camera angled upwards at +45°, we gain a wider field of vision looking vertically below. This change allows observation deeper beneath taller structures, increasing chances of spotting lower-level features including entrances or signage indicative of a pavilion.
}
 &
adjust camera gimbal downwards \\

{\centering 10\par} &
{\centering \raisebox{-0.9\height}{\includegraphics[width=1.0cm]{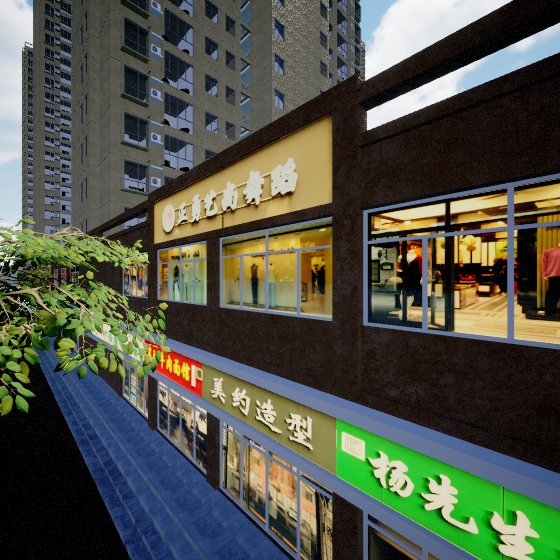}}\par} &
\parbox[t]{12.0cm}{\vspace{0pt}\scriptsize We’ve been observing the same general area repeatedly suggesting we haven’t made much headway finding the pavilion. Letting go of trying to find it precisely via visual exploration alone implies perhaps revisiting initial assumptions or checking alternate locations.
}
 &
move back \\

{\centering 11\par} &
{\centering \raisebox{-0.9\height}{\includegraphics[width=1.0cm]{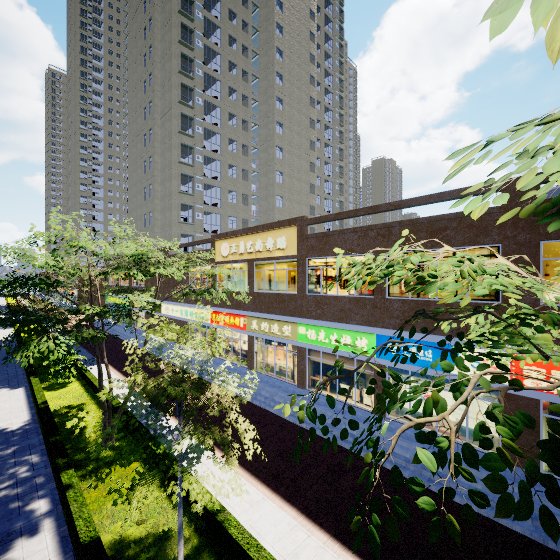}}\par} &
\parbox[t]{12.0cm}{\vspace{0pt}\scriptsize After attempting multiple maneuvers focusing mainly on the current street frontage, it feels like we might benefit from shifting focus outwardly to check neighboring sections or secondary roads that aren't prominently featured in our current view. A strategic retreat combined with a lateral shift could offer novel insights.
}
 &
move back \\

{\centering 12\par} &
{\centering \raisebox{-0.9\height}{\includegraphics[width=1.0cm]{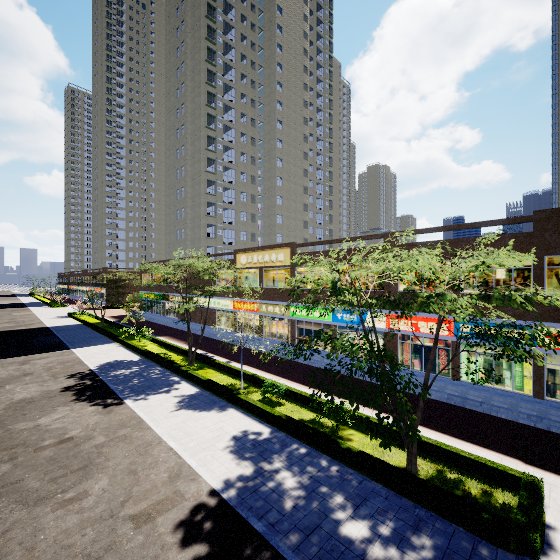}}\par} &
\parbox[t]{12.0cm}{\vspace{0pt}\scriptsize Considering the current view reveals a continuation of similar architecture and greenery, it hints strongly that we remain within the confines of the residential area. We must ensure thorough scanning to locate subtle indicators signaling the existence of a pavilion.
}
 &
adjust camera gimbal upwards \\

{\centering 13\par} &
{\centering \raisebox{-0.9\height}{\includegraphics[width=1.0cm]{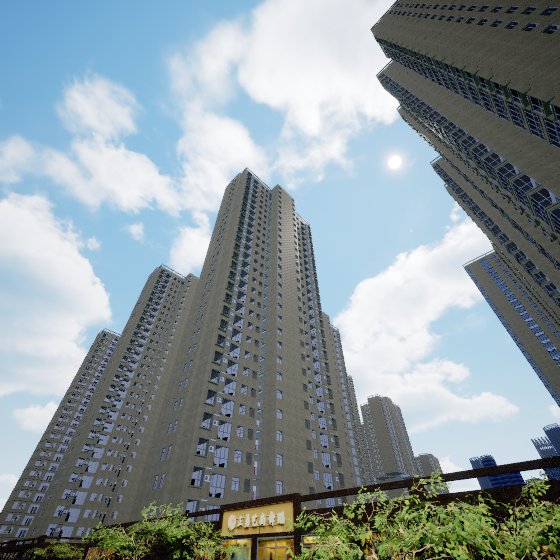}}\par} &
\parbox[t]{12.0cm}{\vspace{0pt}\scriptsize Having elevated the camera angle to 45.0 degrees, we gained a broader overhead view of the residential area. While this gives insight into overall spatial arrangement, pinpointing specifics related to locating a pavilion requires descending back to eye-level perspective.
}
 &
adjust camera gimbal downwards \\

{\centering 14\par} &
{\centering \raisebox{-0.9\height}{\includegraphics[width=1.0cm]{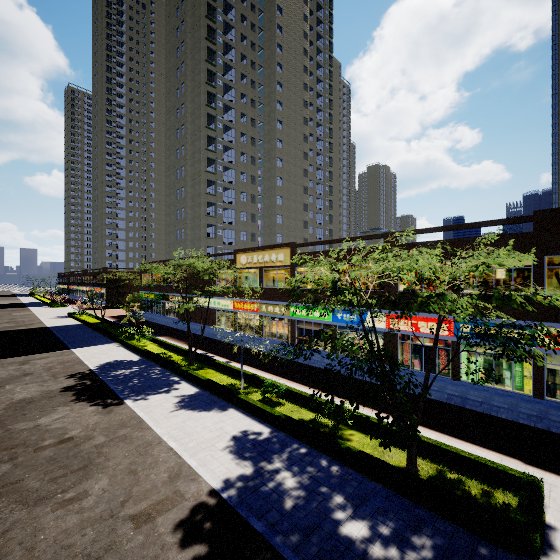}}\par}&
\parbox[t]{12.0cm}{\vspace{0pt}\scriptsize Reviewing the current view, it confirms we are still within the residential area surrounded by multi-story apartment blocks. No distinct markers or unusual shapes stand out that resemble a pavilion under regular circumstances. Thus, it necessitates careful inspection of each shopfront and alleyways for subtle indications.
}
 &
move forth \\

{\centering 15\par} &
{\centering \raisebox{-0.9\height}{\includegraphics[width=1.0cm]{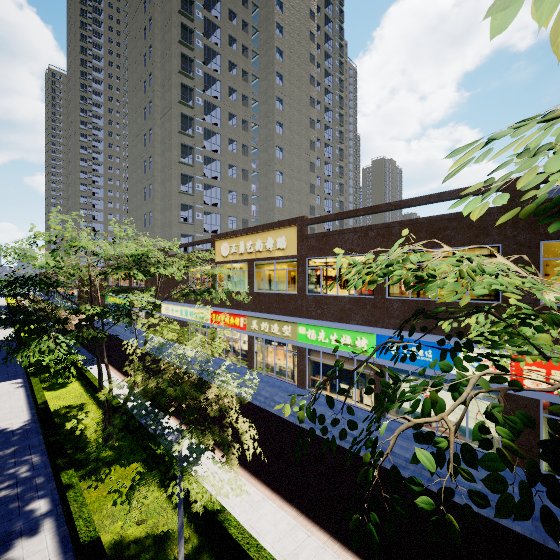}}\par} &
\parbox[t]{12.0cm}{\vspace{0pt}\scriptsize Observation has revealed nothing distinctly marking a transition zone or opening characteristic of a pavilion amongst ordinary storefronts. Further investigation warrants examining the sides and rear aspects of existing buildings for any unexpected openings or covered walkways.
}
 &
turn left \\

{\centering 16\par} &
{\centering \raisebox{-0.9\height}{\includegraphics[width=1.0cm]{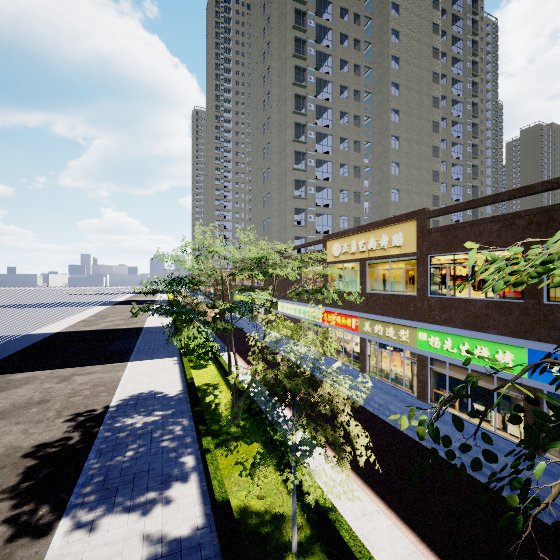}}\par} &
\parbox[t]{12.0cm}{\vspace{0pt}\scriptsize Continuing to look around the area, especially paying attention to the sides and rear of buildings, hasn't yielded evidence of a pavilion yet. The current setup suggests we need to systematically explore all accessible avenues.
}
 &
move forth \\

{\centering 17\par} &
{\centering \raisebox{-0.9\height}{\includegraphics[width=1.0cm]{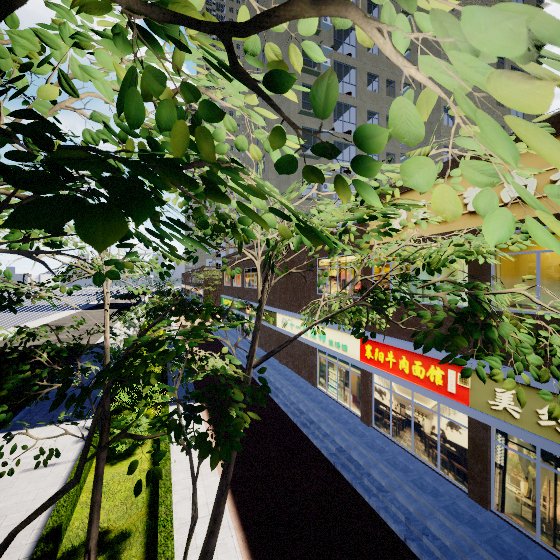}}\par} &
\parbox[t]{12.0cm}{\vspace{0pt}\scriptsize While the current view offers glimpses of tree branches and storefronts, it lacks definitive signs of a pavilion. Continuing straightforward progression might uncover more pertinent details or hidden pathways.
}
 &
move forth \\

\bottomrule
\end{tabular}
\end{table*}

\begin{table*}[h]
\centering
\footnotesize
\caption{Action Process of Qwen2.5-VL-7B}

\begin{tabular}{p{0.8cm} p{1.6cm} p{12.0cm} p{1.5cm}}
\toprule
Step & Obs & Thinking & Action \\
\midrule

{\centering 18\par} &
{\centering \raisebox{-0.9\height}{\includegraphics[width=1.0cm]{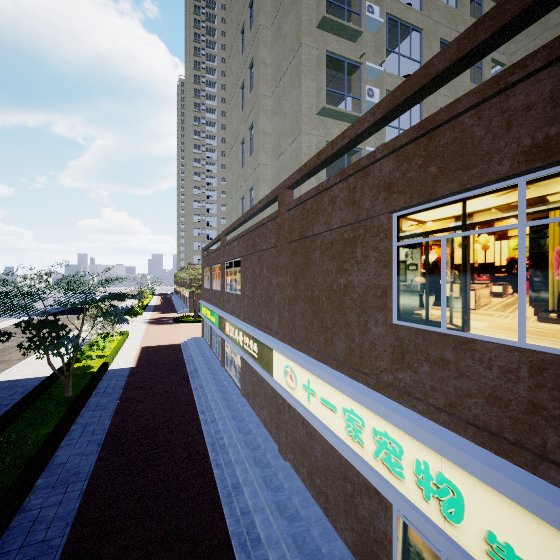}}\par}&
\parbox[t]{12.0cm}{\vspace{0pt}\scriptsize Progression forward didn't yield any noticeable signs of a pavilion. Considering the dense foliage and lack thereof, it might require altering trajectory to explore other possibilities.
}
 &
turn left \\

{\centering 19\par} &
{\centering \raisebox{-0.9\height}{\includegraphics[width=1.0cm]{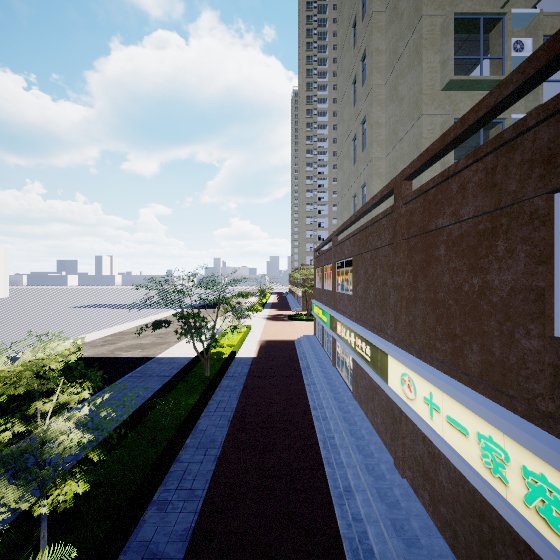}}\par}&
\parbox[t]{12.0cm}{\vspace{0pt}\scriptsize The current view shows a well-maintained pedestrian path alongside buildings, but no evident signs of a pavilion. Turning left might reveal more options or confirm the absence altogether.
}
 &
turn right \\

{\centering 20\par} &
{\centering \raisebox{-0.9\height}{\includegraphics[width=1.0cm]{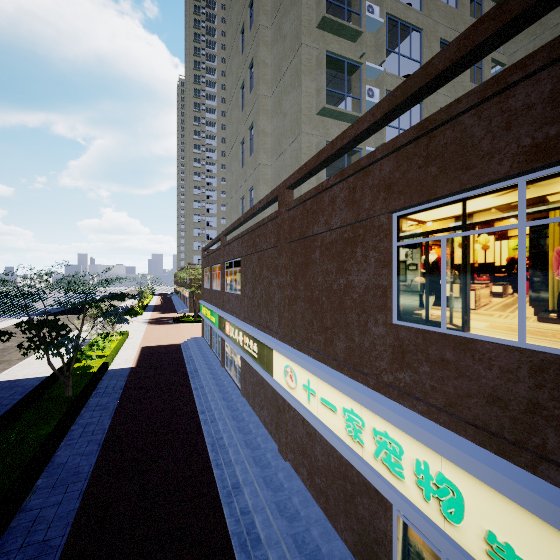}}\par} &
\parbox[t]{12.0cm}{\vspace{0pt}\scriptsize Since the last turn-right maneuver showed no signs of a pavilion, it means we might want to return to the original path or explore further along parallel ways. Given the consistent pattern observed throughout explorations, maintaining steady forward movement aligns best with systematic navigation strategies.
}
 &
move forth \\

{\centering 21\par} &
{\centering \raisebox{-0.9\height}{\includegraphics[width=1.0cm]{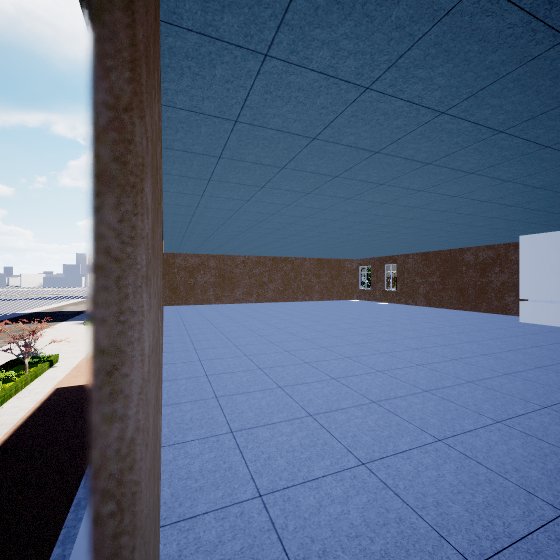}}\par} &
\parbox[t]{12.0cm}{\vspace{0pt}\scriptsize This view presents an empty indoor corridor, which deviates sharply from expectations when seeking a pavilion outdoors. Returning to outdoor navigation ensures adherence strictly to planned goals.
}
 &
move forth \\

{\centering 22\par} &
{\centering \raisebox{-0.9\height}{\includegraphics[width=1.0cm]{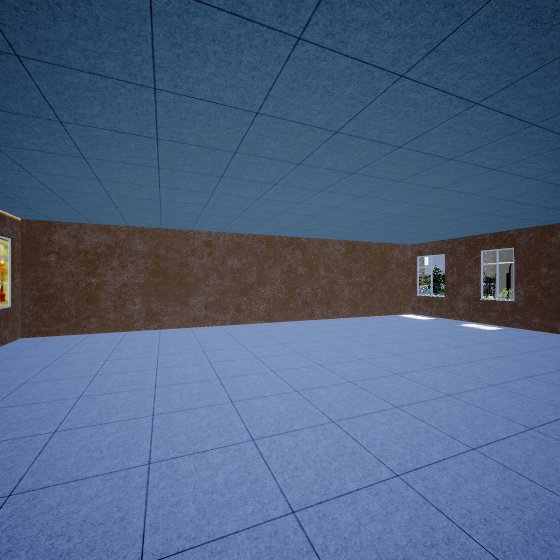}}\par} &
\parbox[t]{12.0cm}{\vspace{0pt}\scriptsize The current view depicts an interior space, which contradicts the goal of navigating to a pavilion within the residential area. It signifies deviation from expected external navigation patterns.
}
 &
move forth \\

{\centering 23\par} &
{\centering \raisebox{-0.9\height}{\includegraphics[width=1.0cm]{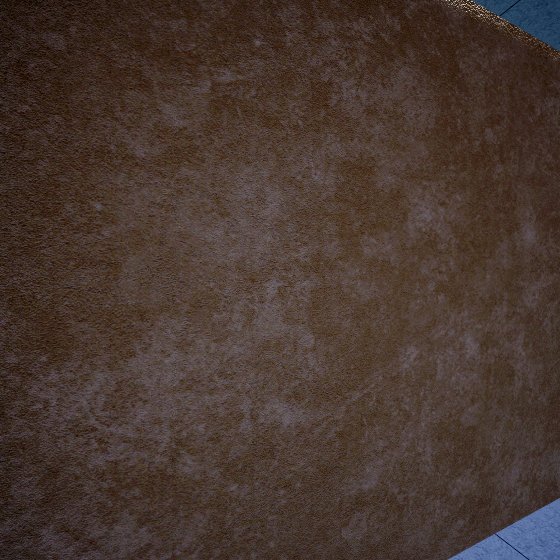}}\par} &
\parbox[t]{12.0cm}{\vspace{0pt}\scriptsize The current view displays a floor texture, indicating that the drone has landed or entered an enclosed space unexpectedly. This situation disrupts ongoing navigation efforts directed towards exiting the residential area and reaching the pavilion.
}
 &
move up \\

{\centering 24\par} &
{\centering \raisebox{-0.9\height}{\includegraphics[width=1.0cm]{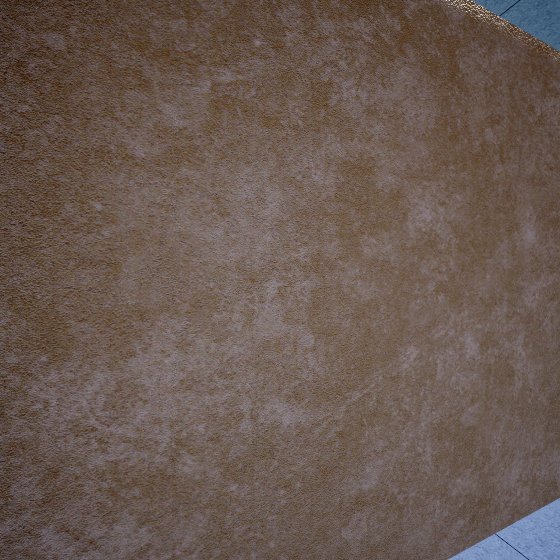}}\par}&
\parbox[t]{12.0cm}{\vspace{0pt}\scriptsize Given the feedback received, it seems important to execute valid movements aligned with predefined commands suitable for drones. Currently, landing or entering an unintended space has halted effective navigation towards the goal.
}
 &
move up \\

{\centering 25\par} &
{\centering \raisebox{-0.9\height}{\includegraphics[width=1.0cm]{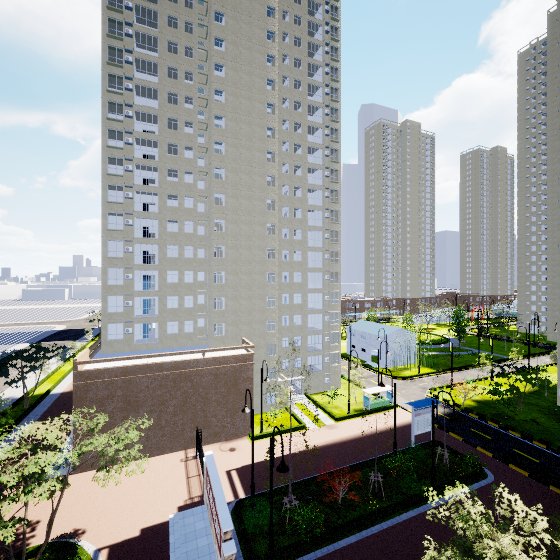}}\par} &
\parbox[t]{12.0cm}{\vspace{0pt}\scriptsize The current aerial view showcases a wide expanse of the residential area, providing ample room to identify potential pathways or structures that might house the pavilion. With the broad scope offered by this vantage point, it's crucial to leverage this opportunity strategically.
}
 &
move up \\

{\centering 26\par} &
{\centering \raisebox{-0.9\height}{\includegraphics[width=1.0cm]{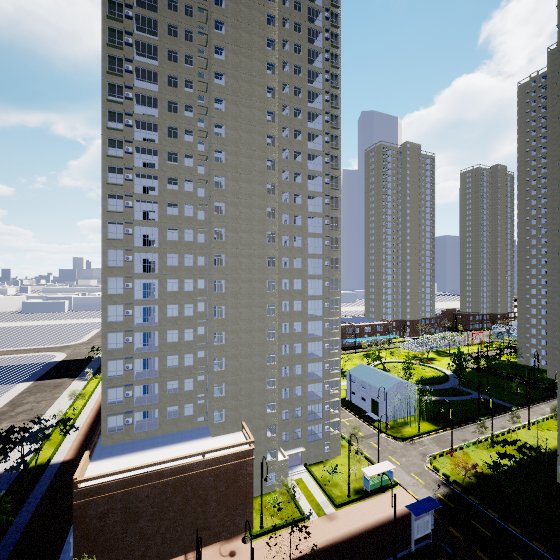}}\par} &
\parbox[t]{12.0cm}{\vspace{0pt}\scriptsize At present, the aerial view captures a large portion of the residential complex featuring numerous high-rise apartments and scattered green patches. Identifying a specific pavilion amid this vast landscape demands precision targeting smaller-scale structures or anomalies that might signify its presence.
}
 &
move down \\

{\centering 27\par} &
{\centering \raisebox{-0.9\height}{\includegraphics[width=1.0cm]{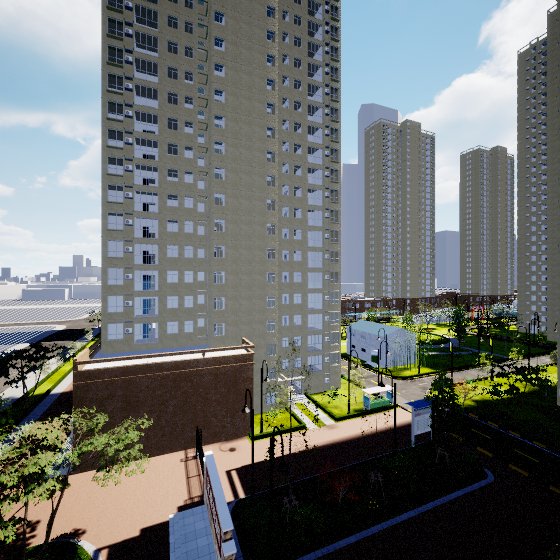}}\par} &
\parbox[t]{12.0cm}{\vspace{0pt}\scriptsize Descending to a lower altitude improves resolution and clarity, helping spot small structures or peculiarities that might denote the pavilion. Maintaining stable elevation aids accurate detection of subtle differences in construction styles often missed otherwise.
}
 &
move down \\

{\centering 28\par} &
{\centering \raisebox{-0.9\height}{\includegraphics[width=1.0cm]{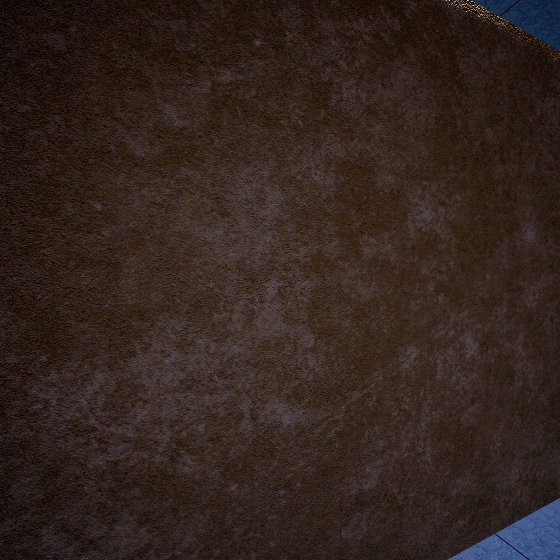}}\par}&
\parbox[t]{12.0cm}{\vspace{0pt}\scriptsize The current view shows a flat surface, indicating that the drone has landed unintentionally. This scenario hinders navigation towards the pavilion because it prevents proper aerial surveillance required for detecting distant targets.
}
 &
move up \\

{\centering 29\par} &
{\centering \raisebox{-0.9\height}{\includegraphics[width=1.0cm]{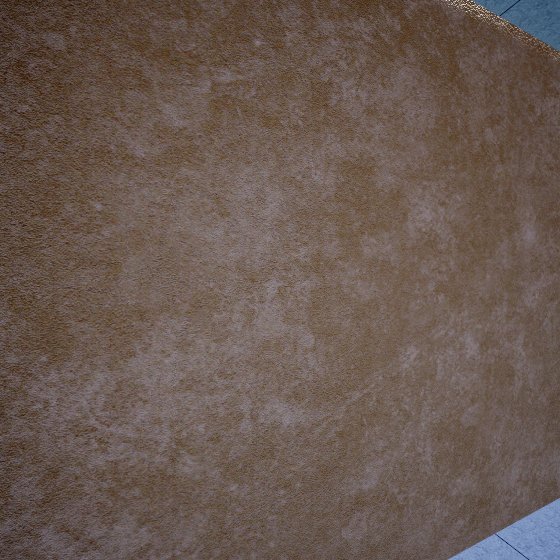}}\par}&
\parbox[t]{12.0cm}{\vspace{0pt}\scriptsize Given the feedback, executing a "takeoff" wasn't acknowledged correctly. As per guidelines, I’ll choose an equivalent action suited for resuming navigation.
}
 &
move up \\

{\centering 30\par} &
{\centering \raisebox{-0.9\height}{\includegraphics[width=1.0cm]{figures/Model/o4_mini/step_026.jpg}}\par}&
\parbox[t]{12.0cm}{\vspace{0pt}\scriptsize I’m much closer now but the pavilion remains off to the lower right and is still partly obscured by the utility building. Shifting the drone to the right will center the pavilion in frame before the final forward approach.
}
 &
move right \\

\bottomrule
\end{tabular}
\end{table*}

\begin{table*}[h]
\centering
\footnotesize
\caption{Action Process of Openfly}

\begin{tabular}{p{0.8cm} p{1.6cm}  p{1.5cm} || p{0.8cm} p{1.6cm}  p{1.5cm}}
\toprule
Step & Obs  & Action & Step & Obs  & Action \\
\midrule

{\centering 0\par} &
{\centering \raisebox{-0.9\height}{\includegraphics[width=1.0cm]{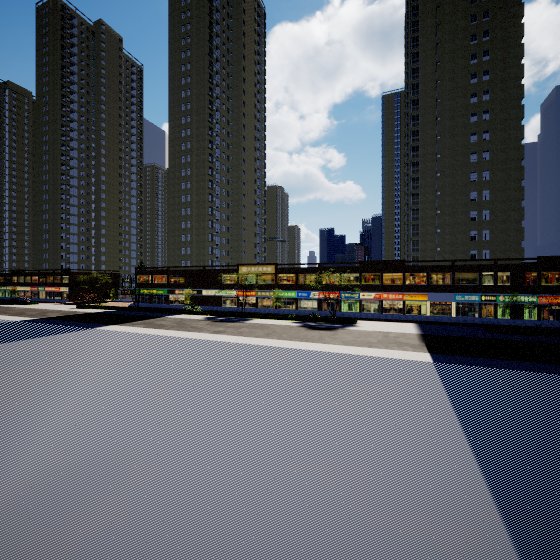}}\par}&
move forth  & {\centering 1\par} &
{\centering \raisebox{-0.9\height}{\includegraphics[width=1.0cm]{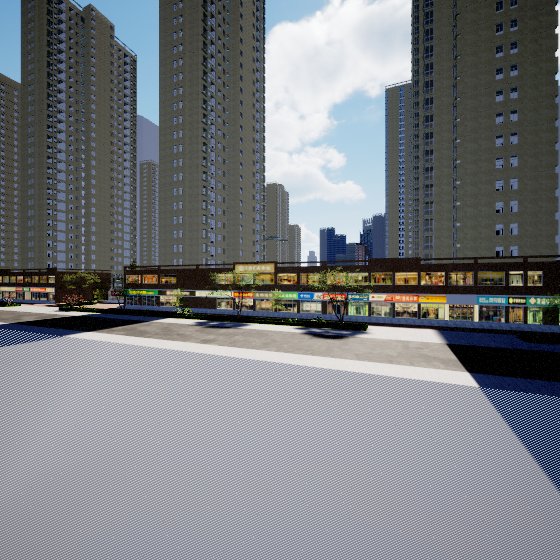}}\par} &

move forth\\

{\centering 2\par} &
{\centering \raisebox{-0.9\height}{\includegraphics[width=1.0cm]{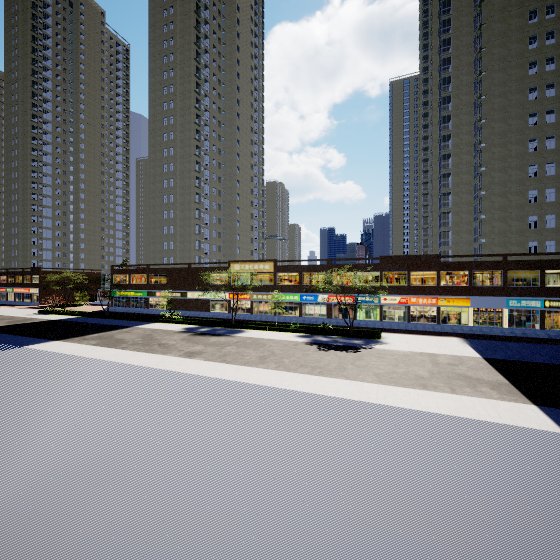}}\par} &

move forth &

{\centering 3\par} &
{\centering \raisebox{-0.9\height}{\includegraphics[width=1.0cm]{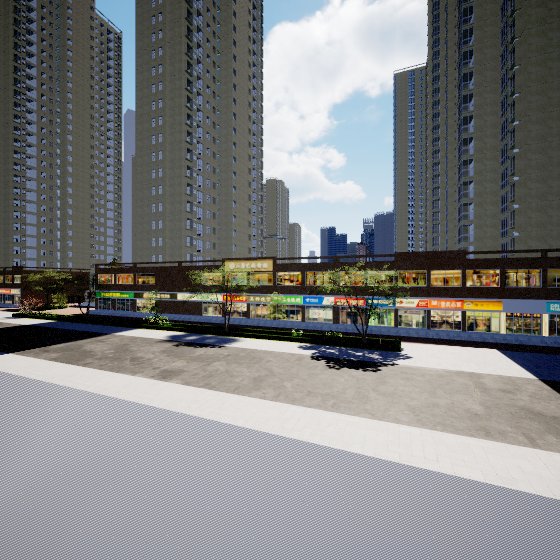}}\par} &
move forth \\

{\centering 4\par} &
{\centering \raisebox{-0.9\height}{\includegraphics[width=1.0cm]{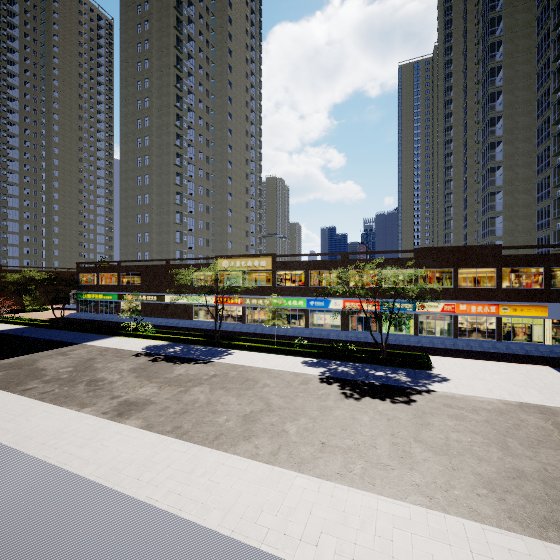}}\par} &

move forth &

{\centering 5\par} &
{\centering \raisebox{-0.9\height}{\includegraphics[width=1.0cm]{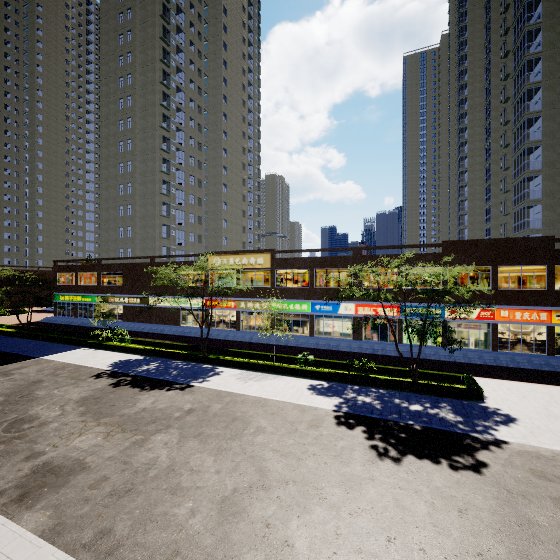}}\par} &
move forth \\

{\centering 6\par} &
{\centering \raisebox{-0.9\height}{\includegraphics[width=1.0cm]{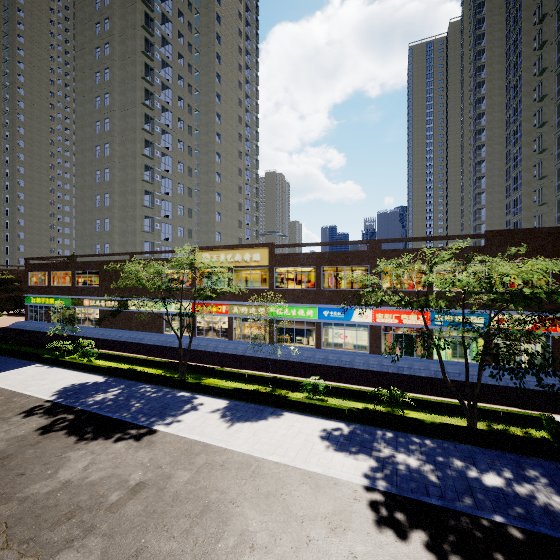}}\par} &
move forth &

{\centering 7\par} &
{\centering \raisebox{-0.9\height}{\includegraphics[width=1.0cm]{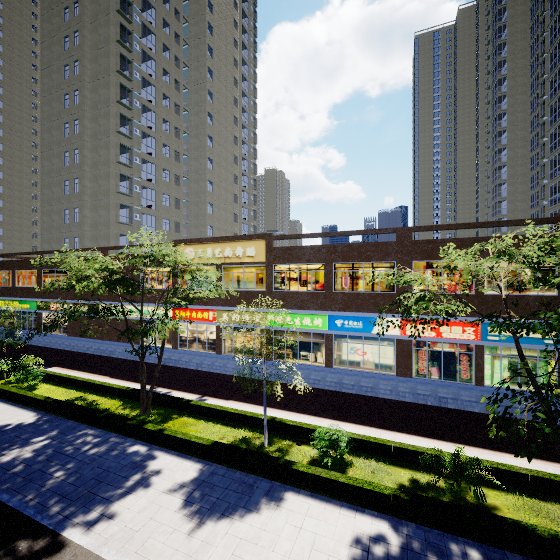}}\par} &

move forth \\

{\centering 8\par} &
{\centering \raisebox{-0.9\height}{\includegraphics[width=1.0cm]{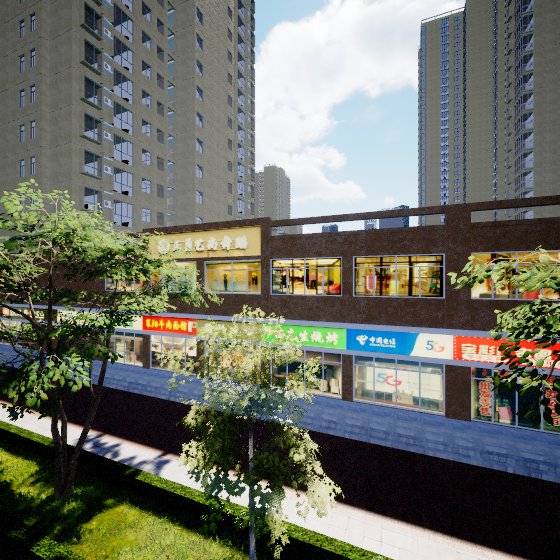}}\par} &
move forth &

{\centering 9\par} &
{\centering \raisebox{-0.9\height}{\includegraphics[width=1.0cm]{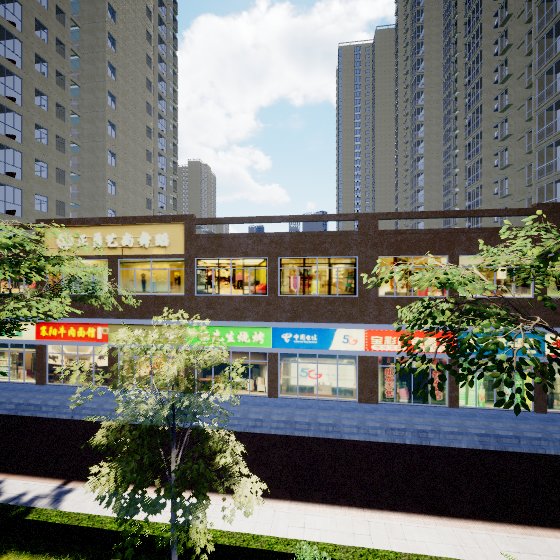}}\par}&
move forth \\

{\centering 10\par} &
{\centering \raisebox{-0.9\height}{\includegraphics[width=1.0cm]{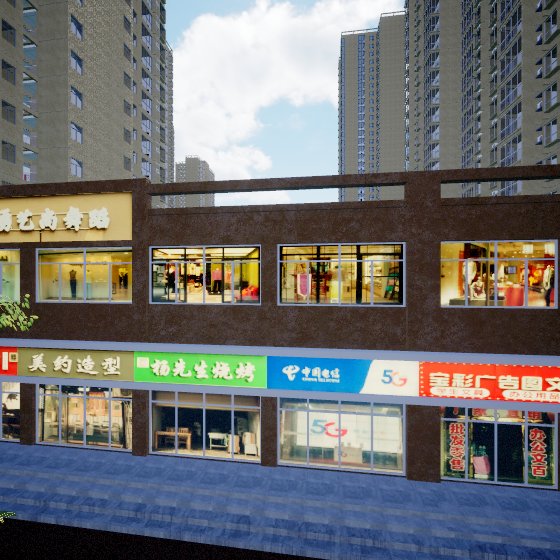}}\par} &
turn right &

{\centering 11\par} &
{\centering \raisebox{-0.9\height}{\includegraphics[width=1.0cm]{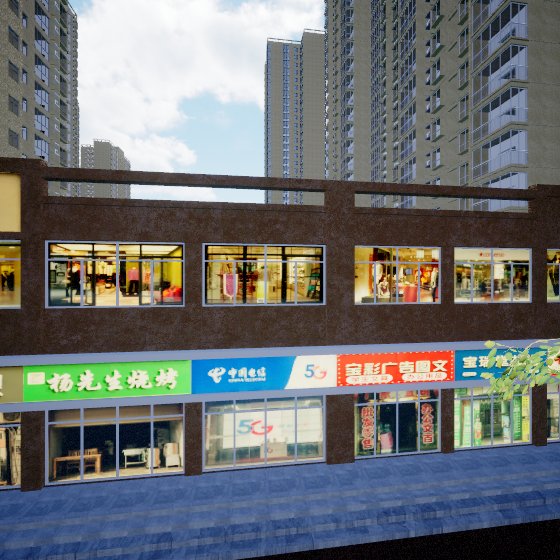}}\par} &
move forth \\

{\centering 12\par} &
{\centering \raisebox{-0.9\height}{\includegraphics[width=1.0cm]{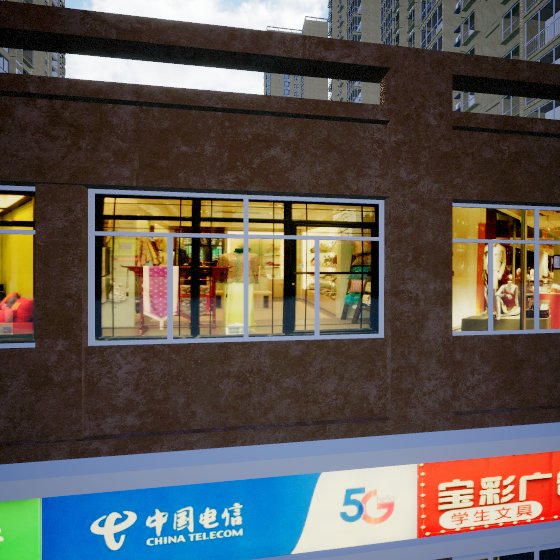}}\par}&
turn right &

{\centering 13\par} &
{\centering \raisebox{-0.9\height}{\includegraphics[width=1.0cm]{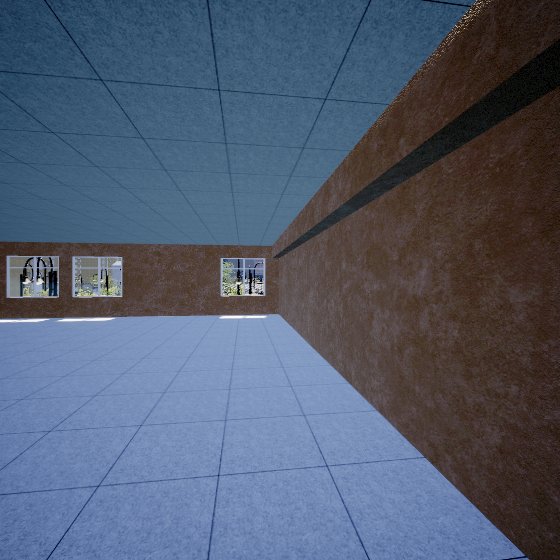}}\par} &
move forth \\

{\centering 14\par} &
{\centering \raisebox{-0.9\height}{\includegraphics[width=1.0cm]{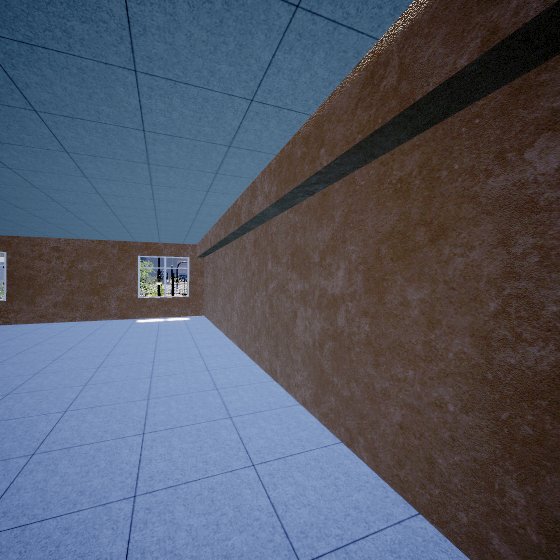}}\par} &
move forth &

{\centering 15\par} &
{\centering \raisebox{-0.9\height}{\includegraphics[width=1.0cm]{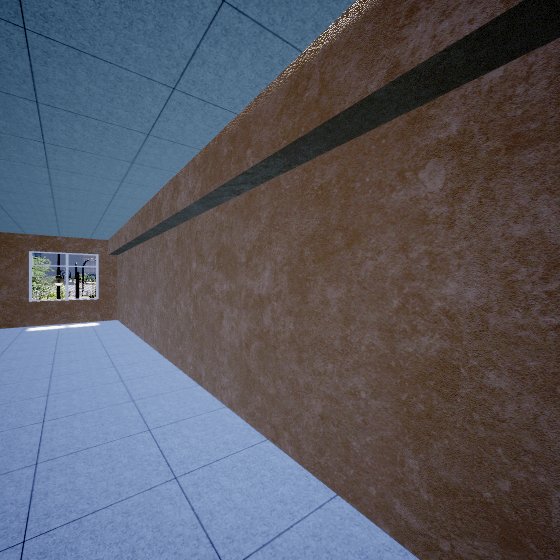}}\par}&
turn right \\

{\centering 16\par} &
{\centering \raisebox{-0.9\height}{\includegraphics[width=1.0cm]{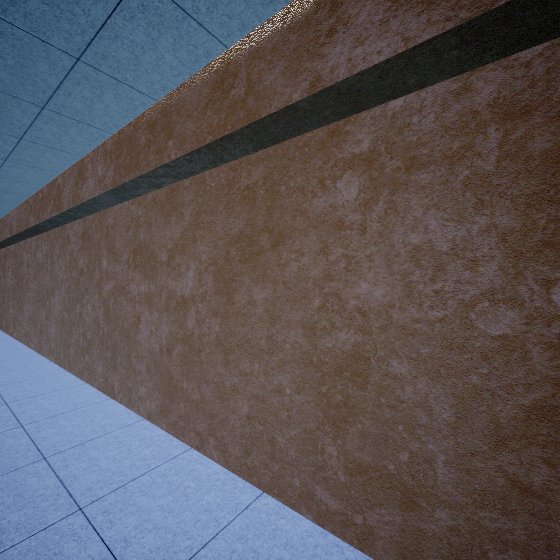}}\par} &
turn right &

{\centering 17\par} &
{\centering \raisebox{-0.9\height}{\includegraphics[width=1.0cm]{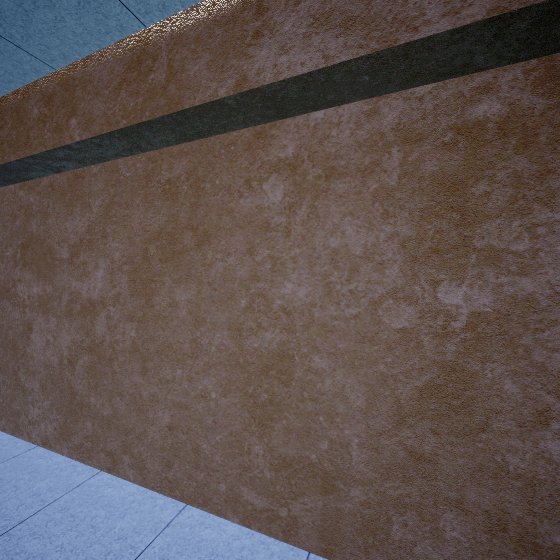}}\par} &
turn right \\

\bottomrule
\end{tabular}
\end{table*}

\end{document}